\documentclass{article}

\usepackage{arxiv}

\usepackage[utf8]{inputenc} 
\usepackage[T1]{fontenc}    
\usepackage{hyperref}       
\usepackage{url}            
\usepackage{booktabs}       
\usepackage{amsfonts}       
\usepackage{nicefrac}       
\usepackage{microtype}      
\usepackage{xcolor}         
\usepackage{tikz}
\usetikzlibrary{calc,decorations.pathreplacing}  

\usepackage{amsmath}
\usepackage{amsthm}

\newcommand\p[1]{\ensuremath{\left( #1 \right)}} 

\ifthenelse{\isundefined{\definition}}{\newtheorem{definition}{Definition}}{}
\ifthenelse{\isundefined{\assumption}}{\newtheorem{assumption}{Assumption}}{}
\ifthenelse{\isundefined{\hypothesis}}{\newtheorem{hypothesis}{Hypothesis}}{}
\ifthenelse{\isundefined{\proposition}}{\newtheorem{proposition}{Proposition}}{}
\ifthenelse{\isundefined{\theorem}}{\newtheorem{theorem}{Theorem}}{}
\ifthenelse{\isundefined{\lemma}}{\newtheorem{lemma}{Lemma}}{}
\ifthenelse{\isundefined{\corollary}}{\newtheorem{corollary}{Corollary}}{}
\ifthenelse{\isundefined{\alg}}{\newtheorem{alg}{Algorithm}}{}
\ifthenelse{\isundefined{\example}}{\newtheorem{example}{Example}}{}

\newcommand{\loss}{\mathcal{L}}
\newcommand{\trainalg}{\mathcal{A}}
\newcommand{\ensalg}{\mathcal{E}}
\newcommand{\ball}{\mathcal{B}}

\DeclareMathOperator*{\argmin}{argmin}
\DeclareMathOperator*{\logitavg}{LogitAvg}

\definecolor{customred}{HTML}{e74c3c}
\definecolor{custompurple}{HTML}{7030A0}
\definecolor{customblue}{HTML}{45BFC6}
\definecolor{customgold}{HTML}{D4AF37}

\usepackage{subcaption}    
\usepackage{enumitem}

\usepackage{multirow}
\usepackage{adjustbox} 
\usepackage{booktabs}
\usepackage{makecell}

\title{Pre-training under infinite compute}

\author{%
  Konwoo Kim\textsuperscript{$\infty$}, Suhas Kotha\textsuperscript{$\infty$}, Percy Liang, Tatsunori Hashimoto \\
  Stanford University \\
}

\begin{document}

\maketitle

\renewcommand{\thefootnote}{$\infty$}
\footnotetext[1]{\href{https://www.youtube.com/watch?v=IOkdMqOVhrM}{Equal contribution.}}
\renewcommand{\thefootnote}{\arabic{footnote}}

\begin{abstract}
\noindent Since compute grows much faster than web text available for language model pre-training, we ask how one should approach pre-training under fixed data and no compute constraints. We first show that existing data-constrained approaches of increasing epoch count and parameter count eventually overfit, and we significantly improve upon such recipes by properly tuning regularization, finding that the optimal weight decay is $30\times$ larger than standard practice. Since our regularized recipe monotonically decreases loss following a simple power law in parameter count, we estimate its best possible performance via the \textbf{asymptote} of its scaling law rather than the performance at a fixed compute budget. We then identify that ensembling independently trained models achieves a significantly lower loss asymptote than the regularized recipe. Our best intervention combining epoching, regularization, parameter scaling, and ensemble scaling achieves an asymptote at 200M tokens using $5.17\times$ less data than our baseline, and our data scaling laws predict that this improvement persists at higher token budgets. We find that our data efficiency gains can be realized at much smaller parameter counts as we can distill an ensemble into a student model that is 8$\times$ smaller and retains $83\%$ of the ensembling benefit. Finally, our interventions designed for validation loss generalize to downstream benchmarks, achieving a $9\%$ improvement for pre-training evals and a $17.5\times$ data efficiency improvement over continued pre-training on math mid-training data. Our results show that simple algorithmic improvements can enable significantly more data-efficient pre-training in a compute-rich future.
\end{abstract}

\section{Introduction}\label{sec:introduction}

Language model pre-training has historically been studied under compute constraints at training \citep{kaplan2020scalinglawsneurallanguage,hoffmann2022trainingcomputeoptimallargelanguage} and inference \citep{sardana2025chinchillaoptimalaccountinginferencelanguage,brown2024largelanguagemonkeysscaling,snell2024scalingllmtesttimecompute} while assuming access to unlimited web text. However, web data increases by $1.03\times$ per year, whereas compute spent on pre-training grows by $4\times$ per year~\citep{villalobos2024rundatalimitsllm,epoch2024trainingcomputeoffrontieraimodelsgrowsby45xperyear}. In anticipation of a regime where compute vastly exceeds data, we ask:

\begin{center}
    \emph{How should one approach pre-training when constrained by data and unconstrained by compute?}
\end{center}

To establish a baseline, we fix a seed training corpus of 200M tokens of web text and evaluate a \emph{standard recipe} following existing data-constrained approaches of repeating data~\citep{muennighoff2023scalingdataconstrainedlanguagemodels} and increasing parameter count~\citep{kaplan2020scalinglawsneurallanguage} (Section~\ref{sec:standard-pt}).
We find that either too many epochs or too many parameters results in the loss eventually increasing due to overfitting. This bounds the performance improvements we can get from tuning this recipe, even if we were willing to spend more compute in exchange for a better model.

\begin{figure}
    \centering
    \includegraphics[width=0.8\textwidth]{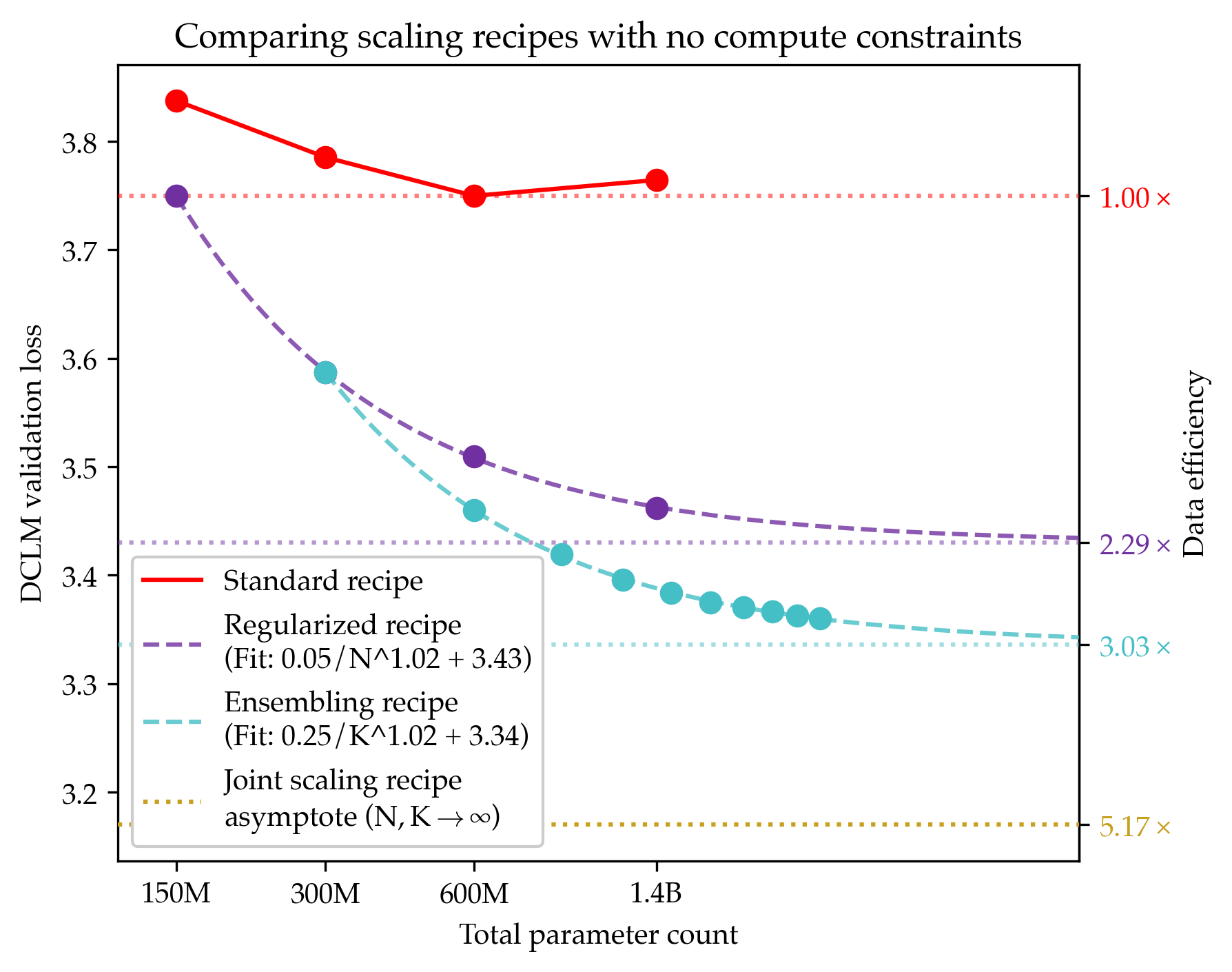}
    \caption{\textbf{Comparing scaling recipes with no compute constraints.} To simulate a data-constrained future, we restrict models to only have access to 200M training tokens. \hyperref[sec:standard-pt]{\textcolor{customred}{(1)}} Standard pre-training recipes overfit with too many epochs or parameters, even if we tune the epoch count at each parameter count $N$ \hyperref[sec:fix-standard-pt]{\textcolor{custompurple}{(2)}} By correctly tuning regularization for each parameter count, loss monotonically decreases following a power law in parameter count $N$. We predict the best possible loss of the regularized recipe by the \textbf{asymptote} of its power law \hyperref[sec:ensembling]{\textcolor{customblue}{(3)}} Instead of scaling $N$, we achieve a lower asymptote by ensembling $K$ models of a fixed parameter count as $K$ approaches infinity. \hyperref[sec:scaling-seed-token]{\textcolor{customgold}{(4)}} Composing parameter scaling and ensemble scaling further improves the asymptote, and we estimate that the baseline would need $5.17\times$ more data to match its loss, even with infinite compute. We additionally find that these data efficiency improvements hold for larger token counts, distilled models, downstream benchmarks, and continued pre-training (Sections~\ref{sec:scaling-seed-token},~\ref{sec:distillation},~\ref{sec:downstream}).}
    \label{fig:sketch-figure-1}
\end{figure}

We instead get predictable monotone scaling in parameter count by considering a \emph{regularized recipe} (Section~\ref{sec:fix-standard-pt}). Currently, regularization used in pre-training is often adopted from existing recipes, for example defaulting to a weight decay of $0.1$ from~\cite{brown2020languagemodelsfewshotlearners}. We find this amount to be inadequate for preventing overfitting under data constraints and that the optimal weight decay is $30\times$ larger than standard practice for our most over-parameterized models. After jointly tuning weight decay, learning rate, and epoch count at each parameter count $N$, loss closely follows a power law in $N$ for parameter-to-token ratios $140\times$ larger than Chinchilla, as shown in Figure~\ref{fig:sketch-figure-1}. 

Normally, researchers would compare two recipes such as regularized and standard by evaluating performance at different train or inference compute budgets~\citep{kaplan2020scalinglawsneurallanguage,hoffmann2022trainingcomputeoptimallargelanguage,brown2024largelanguagemonkeysscaling,snell2024scalingllmtesttimecompute,sardana2025chinchillaoptimalaccountinginferencelanguage}. However, this does not reflect our interest in the best possible performance when constrained by data and unconstrained by compute. Since the loss of the regularized recipe continues to decrease as parameter count increases, we are interested in the limit of the loss as parameter count goes to infinity. More generally, we propose evaluating monotone scaling recipes by the \textbf{asymptote} of their scaling law (for example, $3.43$ for the regularized recipe as seen in Figure~\ref{fig:sketch-figure-1}). By preferring recipes with lower loss asymptotes, we can train better models at sufficiently high compute budgets.

Though taking the parameter count to infinity is one possible limit under infinite compute, we ask if we can design recipes with even lower asymptotes. We consider an alternative \emph{ensembling recipe} where instead of training a larger model, we average the logits of $K$ independently trained models, each of a fixed size \citep{10.5555/648054.743935} (Section~\ref{sec:ensembling}). As shown in Figure~\ref{fig:sketch-figure-1}, the ensembling recipe achieves a lower loss asymptote as ensemble member count approaches infinity compared to the regularized recipe as model parameter count approaches infinity. This implies that at sufficiently high parameter counts, it is advantageous to train multiple smaller models (e.g. two 300M models) instead of a single larger model (e.g. one 600M model). Furthermore, we find that ensembling and parameter scaling compose, achieving a lower asymptote when following the \emph{joint scaling recipe} of taking both the number of members and the parameters of each member to infinity. 

Since our previous experiments were on the scale of $200$M tokens, we study how our recipes scale across higher seed token counts and find that the asymptotes themselves follow a scaling law (Section~\ref{sec:scaling-seed-token}). Our estimates indicate that the 
joint scaling recipe achieves its $200$M asymptote with $5.17\times$ less data than the standard recipe. Importantly, extrapolation of our data scaling laws indicate that the data efficiency improvements will persist at higher token counts.

Though the asymptotes of our recipes benefit the most from large parameter counts, we find that distillation~\citep{hinton2015distillingknowledgeneuralnetwork,kim2016sequencelevelknowledgedistillation} allows us to retain most of the loss improvements without increasing parameter count at inference. Specifically, we find that distilling an 8-ensemble into a single 300M model retains $83\%$ of the ensembling loss improvement over the best regularized 300M model and even outperforms the asymptote of the regularized recipe. We also find that self-distilling a 300M teacher into a student of the same architecture surprisingly reduces loss, improving data efficiency without ever explicitly training a model of higher parameter count.

Finally, we confirm that our validation loss improvements translate to improvements on downstream benchmarks (Section~\ref{sec:downstream}). Ensembles with better validation loss perform better on downstream benchmarks, with our best ensemble outperforming our best unregularized model by $9\%$ on average over PIQA, SciQ, and ARC Easy (standard pre-training benchmarks for models at our scale \citep{thrush2025improvingpretrainingdatausing}). We also test the immediate applicability of our interventions for continued pre-training (CPT) on the \texttt{MegaMath-Web-Pro} dataset from~\cite{wang2025octothinkermidtrainingincentivizesreinforcement}. We find that with only 4B tokens, the ensembling recipe (i.e. an ensemble of epoched models) outperforms default CPT on the full 73B tokens following their training hyperparameters, resulting in a $17.5\times$ data efficiency improvement. 

We open-source all of our \href{https://wandb.ai/stanford-mercury/suhas-data-efficiency/reports/Pre-training-under-infinite-compute--VmlldzoxNDM5NzUzMQ}{runs} on WandB and our \href{https://github.com/marin-community/marin/tree/suhas/data-efficiency}{code} on Github.

\section{Standard pre-training}\label{sec:standard-pt}

Historically, pre-training has focused on training the best possible models subject to compute or parameter constraints. 
Under train compute constraints, scaling recipes like Chinchilla recommend jointly increasing data and model size with $20\times$ more tokens than parameters~\citep{kaplan2020scalinglawsneurallanguage,hoffmann2022trainingcomputeoptimallargelanguage}. Under parameter constraints for cheaper inference and fine-tuning, current practice opts for over-training language models relative to Chinchilla with token counts $2000\times$ larger than the parameter count or distilling from preexisting larger models~\citep{gadre2024languagemodelsscalereliably,grattafiori2024llama3herdmodels,sardana2025chinchillaoptimalaccountinginferencelanguage,busbridge2025distillationscalinglaws}. 

Many prior works implicitly assume no constraint over the number of tokens and always train on fresh data. In this paper, we are instead interested in studying how algorithms fare under data constraints, preventing us from continuing to jointly scale token and parameter count. To study data-constrained pre-training in its purest form, we return to the classical statistical formulation of learning, lifting all other constraints such as train compute and parameter count. To formulate our objective, we formalize standard pre-training as a training routine $\trainalg$ that accepts arguments such as token count $D$, parameter count $N$, epoch count $E$ to produce a model $M$ with loss $\loss(M)$. Any arguments that are not explicitly specified are passed through hyperparameter tuple $H$. Our objective for pre-training with $D$ tokens of data unconstrained by compute becomes
$$\loss^*_{D} = \min_{\substack{H}} \loss\p{\trainalg\p{D, H}}$$

To measure the performance of an algorithm, we construct a controlled pre-training environment with a limited amount of web data from DCLM~\citep{li2025datacomplmsearchgenerationtraining}. Since we are interested in algorithms that spend orders of magnitude more compute than Chinchilla scaling at a fixed data budget, we default to a smaller token count of 200M tokens and test whether our findings hold across higher token counts in Section~\ref{sec:scaling-seed-token}. For evaluation, we defer to loss on a held-out i.i.d. validation set since this is shown to correlate with downstream pre-trained capabilities in our analysis in Section~\ref{sec:downstream} and prior work~\citep{chen2025scalinglawspredictingdownstream,thrush2025improvingpretrainingdatausing, gadre2024languagemodelsscalereliably}.To best represent standard practice, we follow a standard auto-regressive recipe, using Llama-style architecture, AdamW optimizer with cosine learning rate schedule, context length 4096, etc. (full details in Appendix~\ref{app:problem-setting})

\subsection{Evaluating existing data-constrained recipes}\label{sec:chinchilla-fails}

\begin{figure}[b!]
    \centering
    \begin{minipage}{0.47\textwidth}
        \centering
        \includegraphics[width=1.0\linewidth]{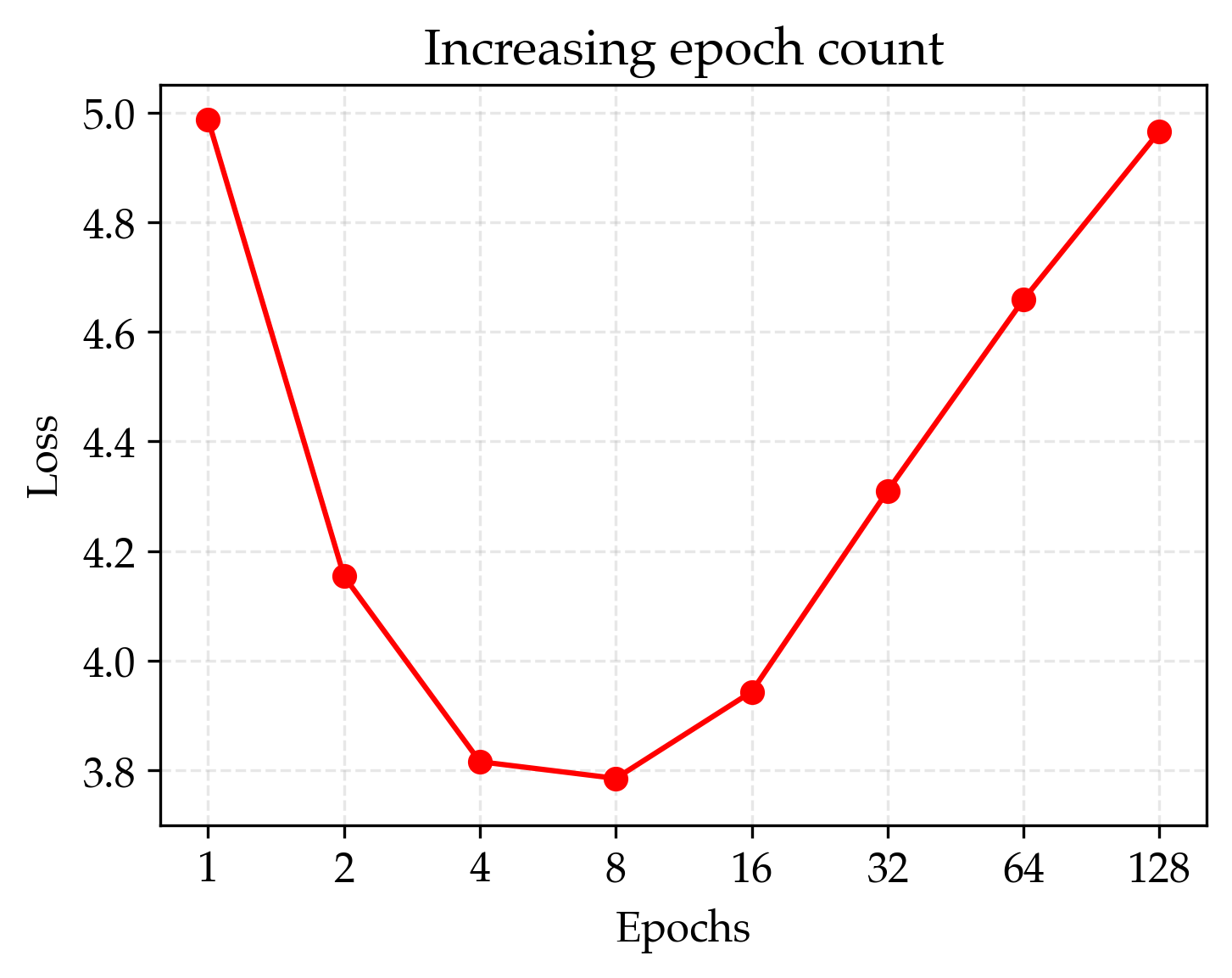}
        \small
        \begin{tabular}{lcccc}
            \hline
            \textbf{Tuned $H$} & \textbf{1} & \textbf{8} & \textbf{128} \\
            \hline
            Learning rate &  1e-3 & 1e-3 & 3e-3 \\
            \hline
            \phantom{Epoch count} & \phantom{16} & \phantom{16} & \phantom{8} & \phantom{8} \\
        \end{tabular}
    \end{minipage}\hfill
    \begin{minipage}{0.47\textwidth}
        \centering
        \includegraphics[width=1.0\linewidth]{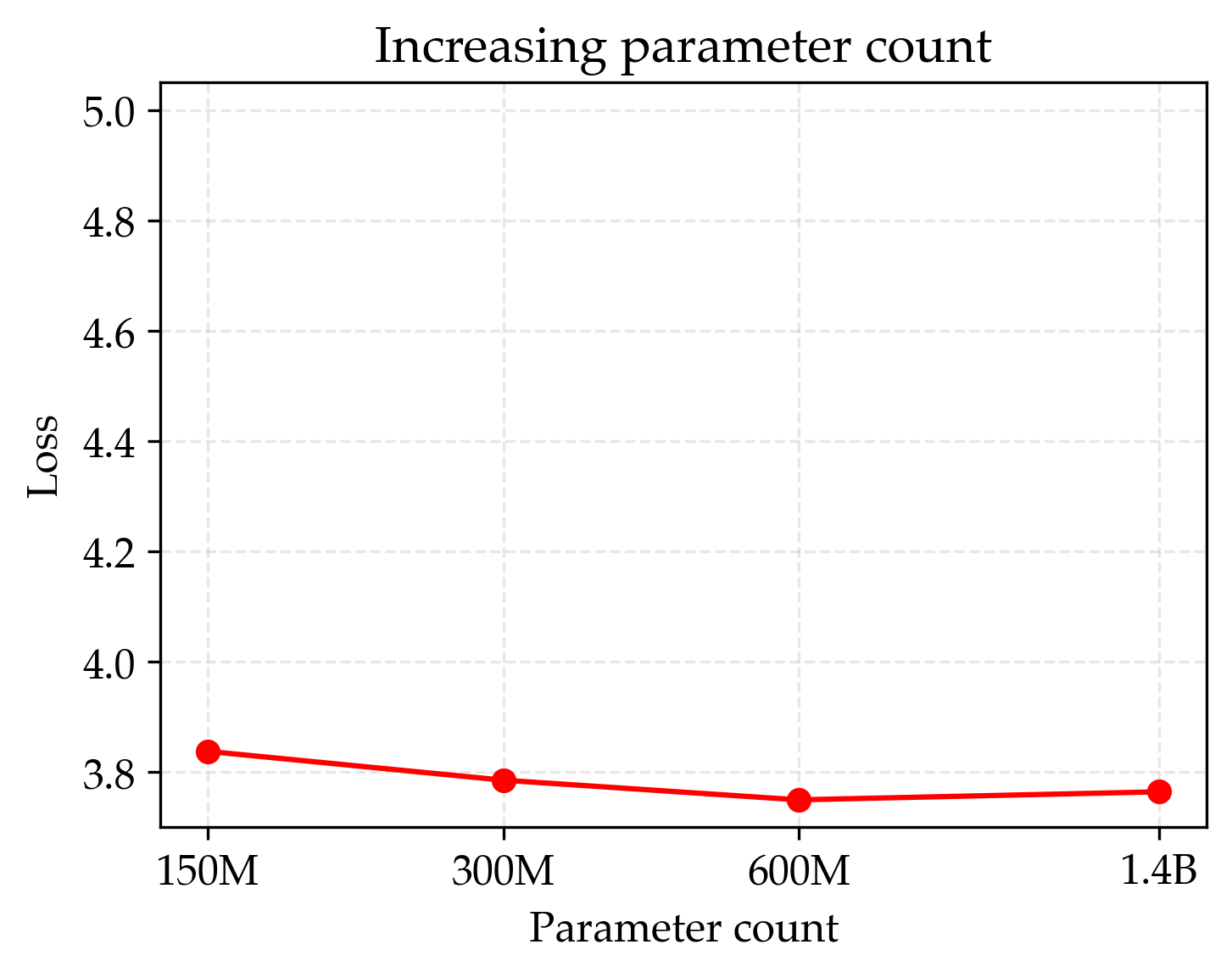}
        \setlength{\tabcolsep}{6pt}
        \small
        \renewcommand{\arraystretch}{1.15}
        \begin{tabular}{lcccc}
            \hline
            \textbf{Tuned $H$} & \textbf{150M} & \textbf{300M} & \textbf{600M} & \textbf{1.4B}  \\
            \hline
            Learning rate &  3e-3  & 1e-3 & 1e-3 & 3e-4 \\
            Epoch count & 8 & 8 & 4 & 4 \\
            \hline
        \end{tabular}
    \end{minipage}
    \caption{\textbf{Evaluating standard recipe of epoching and parameter scaling for 200M tokens.} Left: Though repeating the data lowers the loss, too many repetitions results in overfitting for 300M models. Right: We try increasing parameter count, tuning the epoch count at each parameter count. We similarly find that loss starts increasing. Moreover, increasing the parameter count $10\times$ improves the loss by less than $0.1$.}
    \label{fig:sketch-chinchilla-fails}
\end{figure}

Since the amount of fresh data is limited, we build a standard recipe of increasing repetition count~\citep{muennighoff2023scalingdataconstrainedlanguagemodels} and parameter count~\citep{kaplan2020scalinglawsneurallanguage}. Since there is unlimited compute, we depart from compute-efficient practice by training models that are much larger relative to the token count, defaulting to 300M parameter models for a 200M seed token constraint.

We first consider increasing the epoch count $E$ at a fixed parameter count, taking $E\times$ more training compute. We find that for sufficiently high epoch counts, the models start overfitting and loss starts increasing (Figure~\ref{fig:sketch-chinchilla-fails}, left), indicating that the data can not be epoched forever. These experiments disagree with the functional form of the decay-based scaling law in~\cite{muennighoff2023scalingdataconstrainedlanguagemodels}, which posits that loss monotonically decreases in epoch count. Their work acknowledges this discrepancy and removes most runs that overfit when fitting their scaling law, discussed in their Appendix D.

Since epoch count cannot be increased arbitrarily without increasing loss, we turn to increasing parameter count. To establish a competitive scaling recipe, we jointly tune epoch count and learning rate for each parameter count (described in Appendix~\ref{app:convex-certificates}). Even after tuning these hyperparameters, increasing parameter count does not significantly decrease loss, and the 1.4B model performs worse than the 600M model. This is consistent with the single-pass findings in~\cite{kaplan2020scalinglawsneurallanguage}, Figure 9 which show that increasing parameter count eventually starts increasing loss for fixed data. It is likely that both increasing repetition count and parameter count result in overfitting the train set, detailed in Appendix~\ref{app:overfitting}.


\section{Regularized parameter scaling}
\label{sec:fix-standard-pt}

\begin{figure}[b]
    \centering
    \begin{minipage}{0.52\textwidth}
        \centering
        \includegraphics[width=\linewidth]{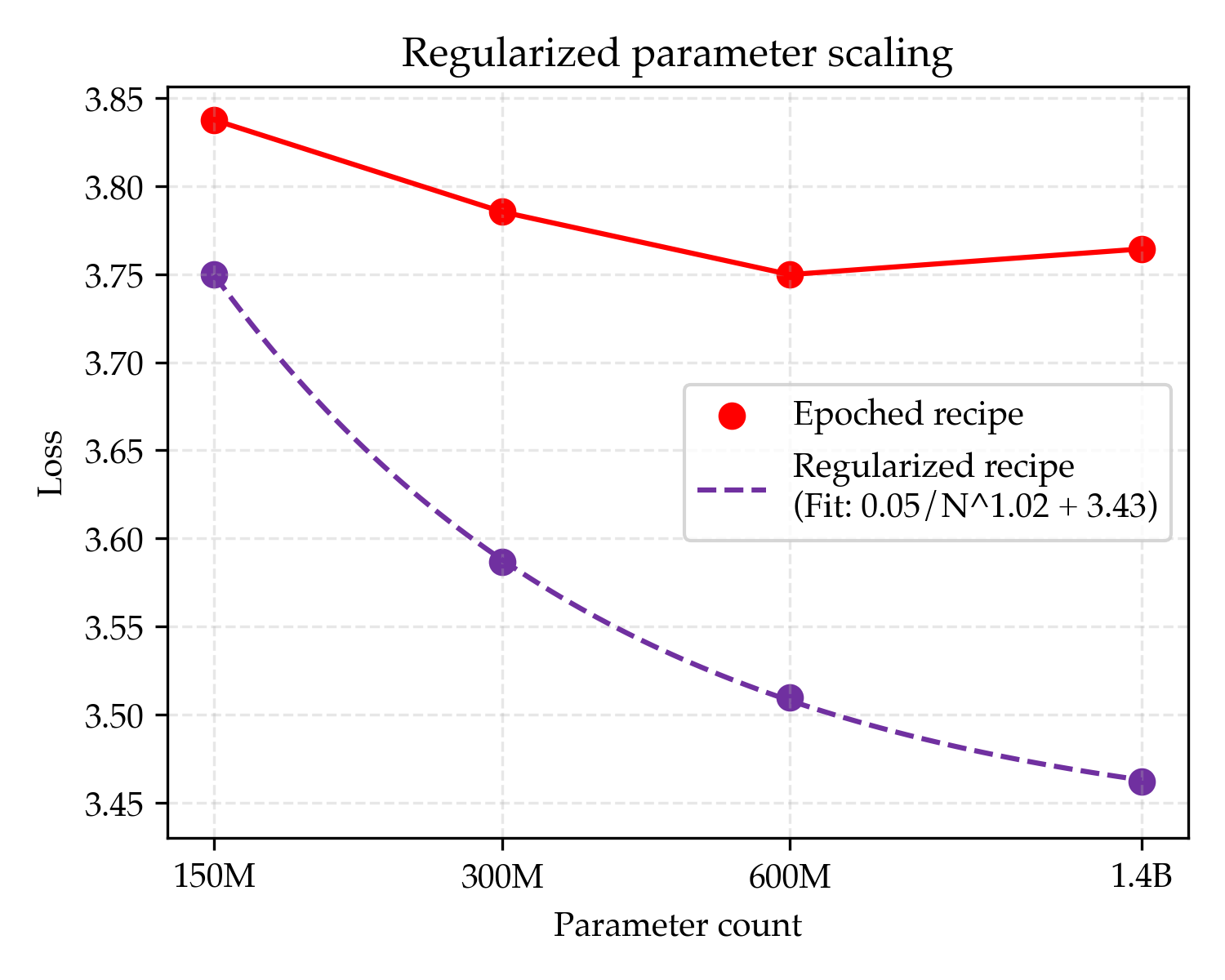}
    \end{minipage}\hfill
    \begin{minipage}{0.04\textwidth}
        \phantom{.}
    \end{minipage}\hfill{}
    \begin{minipage}{0.42\textwidth}
        \centering
        \setlength{\tabcolsep}{6pt}
        \renewcommand{\arraystretch}{1.15}
        \small
        \begin{tabular}{lcccc}
            \hline
            \textbf{Tuned $H$} & \textbf{150M} & \textbf{300M} & \textbf{600M} & \textbf{1.4B} \\
            \hline
            Learning rate &  3e-3  & 3e-3 & 1e-3 & 1e-3 \\
            Epoch count & 16 & 16 & 8 & 8 \\
            Weight decay & 0.8 & 1.6 & 3.2 & 3.2 \\
            \hline
        \end{tabular}
        
        \phantom{.}

        \phantom{.}

        \caption{\textbf{Power law scaling from jointly tuning regularization.} We compare the standard recipe in Figure~\ref{fig:sketch-chinchilla-fails} (red line) to our regularized recipe that jointly tunes learning rate, epoch count, and weight decay (purple line). After tuning with regularization, the loss decreases proportional to $\approx \frac{1}{N}$. The power law predicts that as parameter count goes to infinity, the best model achieves $3.43$ loss.}\label{fig:chinchilla-comparison}
    \end{minipage}
\end{figure}

We show that to get the best performance from these over-parameterized, epoched models, it is critical to regularize pre-training with much higher weight decay than standard practice.  To jointly tune weight decay, learning rate, and epoch count, we perform an extensive search for ``locally optimal'' hyperparameters where increasing or decreasing any hyperparameter does not result in a better model. To find these hyperparameters, we use a coordinate descent algorithm inspired by~\cite{wen2025fantasticpretrainingoptimizers} (more details in Appendix~\ref{app:convex-certificates}). We find that over-parametrized models need much higher weight decay, over $30\times$ larger than the standard practice of $0.1$ (Figure~\ref{fig:chinchilla-comparison}, right table).

With this tuning, loss follows monotone scaling in parameter count for models up to $140\times$ larger than Chinchilla as shown in Figure~\ref{fig:chinchilla-comparison}. This agrees with theory for over-parameterized regression that predicts that even when loss does not monotonically decrease due to double descent, the loss will monotonically decrease when regularization is optimally tuned~\citep{PhysRevX.6.031034,nakkiran2021optimalregularizationmitigatedouble,Canatar_2021,simon2024bettermodernmachinelearning}. In Appendix~\ref{app:ablate-convex-tuning}, we show how our locally-optimal tuning procedure is critical to achieve this monotone scaling.

To capture how increasing parameter count improves loss, we fit a power law with an asymptote as
$$\hat{\loss}_{D,N} \coloneq \frac{A_D}{N^{\alpha_D}} + E_D$$ where we fit free variables $A_D, \alpha_D, E_D$. Fitting this law to the runs across four parameter counts results in $\hat{\loss}_{200\text{M}, N} = \frac{0.05}{N^{1.02}} + 3.43$\footnote{For this power law and all following ones, the units of parameters and tokens will be in billions for cleaner visualization and comparison. This only affects the numerator of the scaling law, not its asymptote or exponent.}. This exponent of $1.02$ for parameter scaling is quite high considering that Chinchilla finds a parameter scaling exponent of $0.34$. This suggests that when we better leverage the data, there is faster improvement from using larger models.

However, if we truly had infinite compute, the monotone scaling law suggests that we should increase the parameter count $N$ as much as we can, independent of the scaling exponent. This is different from the train compute-constrained regime where increasing $N$ comes at the cost of training on less data and can hurt performance. To characterize the best possible performance when unconstrained by compute, we are interested in the loss under the limit as parameter count goes to infinity. Assuming the power law fit, we can estimate the limit as $\lim_{N\to\infty} \hat{\loss}_{D,N}$, equivalent to the asymptote $E_D$. The asymptote for the regularized recipe law predicts that the best possible model it can train achieves loss $3.43$. Because of run-to-run variance of the points forming the power law, we share a sensitivity analysis in Appendix~\ref{app:sensitivity} and find that the estimated asymptotes vary by at most $0.02$ loss across 3 seeds.

\section{Ensemble scaling}\label{sec:ensembling}

The regularized recipe offers a straightforward way to improve performance by taking $N\to\infty$. Is this the best possible loss under infinite compute, or can different training algorithms better leverage the data? In this section, we consider simple ensembling \citep{10.5555/648054.743935}: independently train $K$ models and average their logits for generation, formalized in Section~\ref{sec:formalizing-ensembles}. In Section~\ref{sec:ens-vs-parameter}, we show how ensembling can outperform parameter scaling at fixed parameter counts and under the limit as total parameter count approaches infinity. In Section~\ref{sec:infinite-ensembles}, we construct our best recipe composing regularized parameter scaling and ensemble scaling by taking the limit as both $N, K\to\infty$.

\subsection{Formalizing ensembles}\label{sec:formalizing-ensembles}


The ensembling pre-training algorithm $\ensalg$ accepts a standard pre-training algorithm $\trainalg$ and trains $K$ members that are identical except for random seed $Z_i$ controlling the data order and model initialization. The output of the ensembling algorithm is a model that averages the logits of the $K$ models, computed by querying all $K$ models. More formally, we define the ensembling algorithm as
$$
\ensalg_\trainalg(D, N, K, H) = \logitavg\p{\left\{\trainalg\p{D, N, Z_i, H}\right\}_{i\in[K]}}
$$

for randomness $Z_i$, where $\logitavg$ produces a model with likelihood for a sequence $x$ given by
$$
\logitavg\p{M_{i\in[K]}}\p{x} \propto \exp\p{\frac{1}{K} \sum_{i\in[K]} \log \p{M(x)}}
$$

The number of FLOPs needed to generate from or evaluate an ensemble is simply the sum of the costs for all members. Since the number of FLOPs in a forward pass is approximately linear in parameter count~\citep{kaplan2020scalinglawsneurallanguage,hoffmann2022trainingcomputeoptimallargelanguage}, we will consider an ensemble's total parameter count as $NK$ when comparing it to standard pre-training. 

Scaling ensembles raises two significant questions. In Section~\ref{sec:ens-vs-parameter}, we ask whether the asymptote under $K\to\infty$ for fixed $N$ outperforms the asymptote under $N\to\infty$ for $K=1$. In Section~\ref{sec:infinite-ensembles}, we ask whether we can compose both recipes by measuring the asymptote under $N, K\to\infty$.

\begin{figure}
    \centering
    \includegraphics[width=0.6\textwidth]{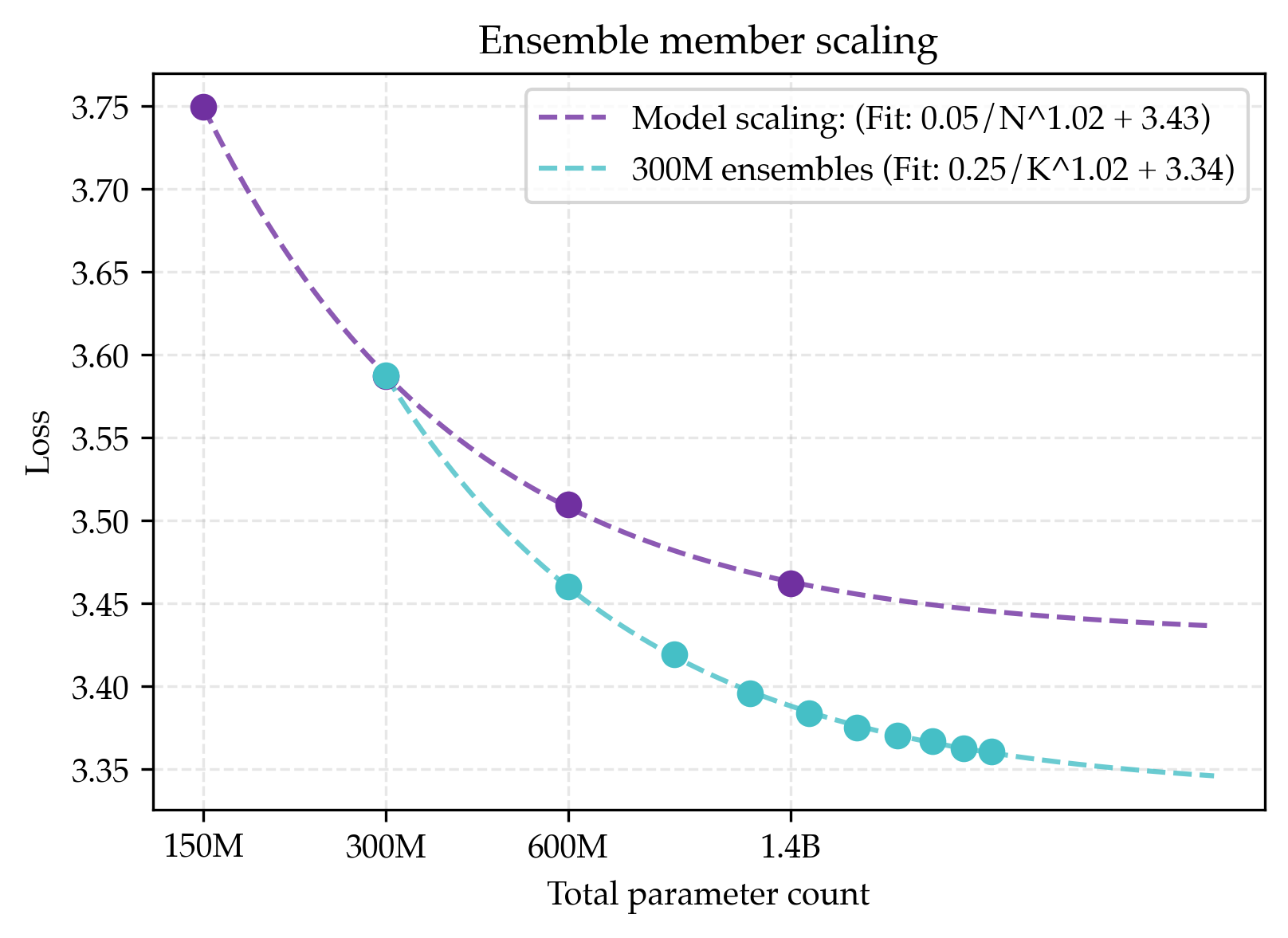}
    \caption{\textbf{Comparing scaling parameter count vs scaling ensemble member count.} Instead of scaling the parameter count of a single model, we can train an ensemble of smaller models and scale the number of ensemble members (resulting in $NK$ total parameters for $K$ ensemble members). Scaling up member count $K$ can similarly be fit by a power law with exponent approximately $1$. Importantly, this law achieves a better asymptote than scaling up parameter count.}
    \label{fig:sketch-ens-vs-model-scale}
\end{figure}

\subsection{Scaling member count instead of parameter count}\label{sec:ens-vs-parameter}

We compare the regularized and ensembling recipes under the best regularized hyperparameters from Section~\ref{sec:fix-standard-pt}. In Figure~\ref{fig:sketch-ens-vs-model-scale}, we find that the excess loss for the ensembling recipe decreases close to a rate of $\frac{1}{K}$, similar to how the excess loss for the regularized recipe decreases at a rate close to $\frac{1}{N}$ in Figure~\ref{fig:chinchilla-comparison}. However, under the limit of infinite compute, the asymptote of ensembling ($N=300\text{M, } K\to\infty$) is $3.34$, which is lower than the asymptote of the regularized recipe ($N\to\infty, K=1$), which is $3.43$. This implies that for sufficiently large parameter counts, it is advantageous to train multiple small models (e.g. two 300M models) instead of a single large model (e.g. one 600M model). In fact, even the $K=3$ ensemble outperforms the asymptote of the regularized recipe.

Why does ensembling improve over standard parameter scaling?~\cite{allenzhu2023understandingensembleknowledgedistillation} shows that ensembling helps when the data can be well-classified with one of many unique features but is best classified when using all such features. Under this ``multi-view'' structure, they find that training a single model is biased towards only learning one feature, whereas each member of an ensemble happens to learn different features when independently trained. 

\begin{figure}
    \centering
    \includegraphics[height=0.36\textwidth]{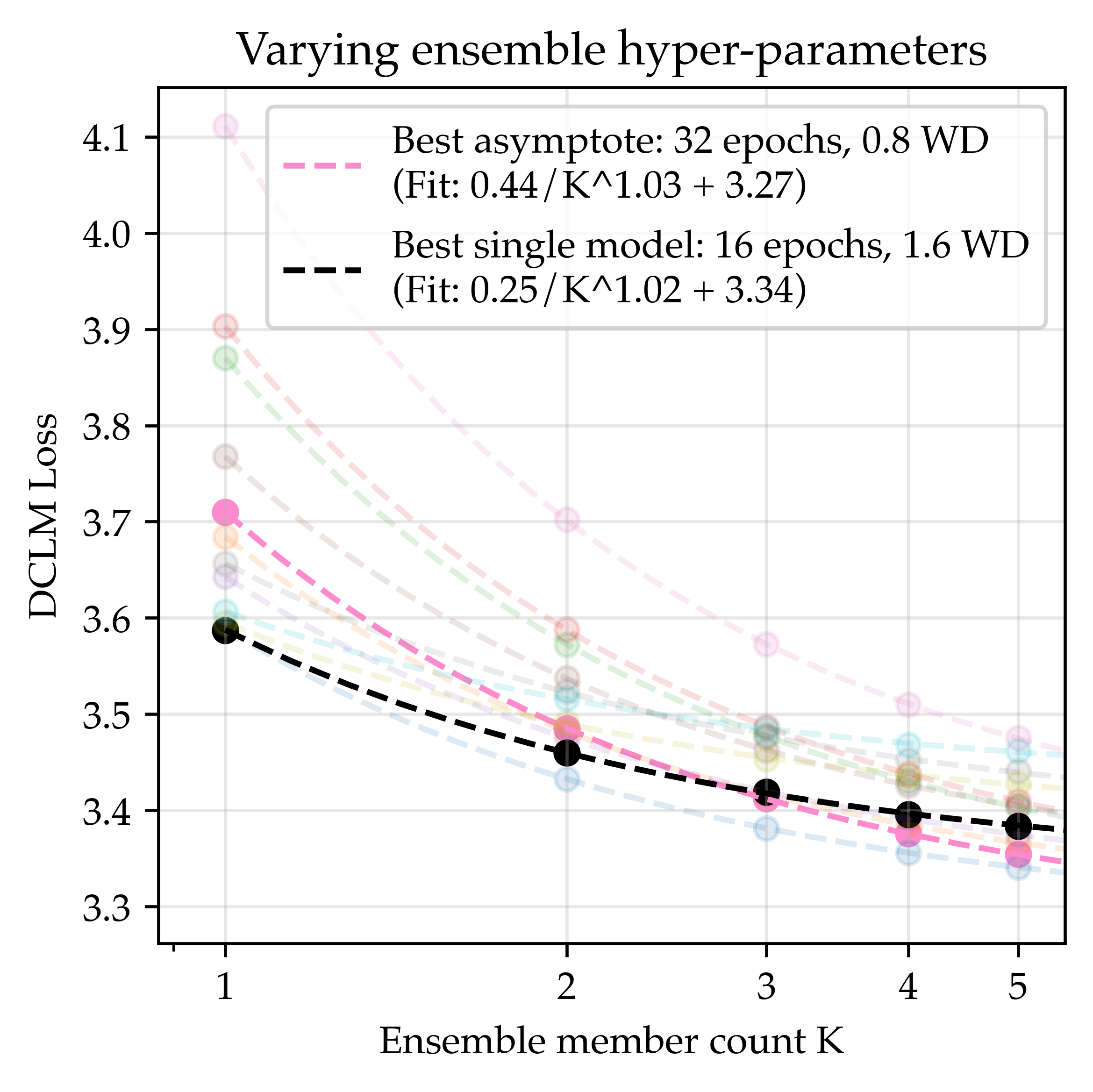}
    \includegraphics[height=0.36\textwidth]{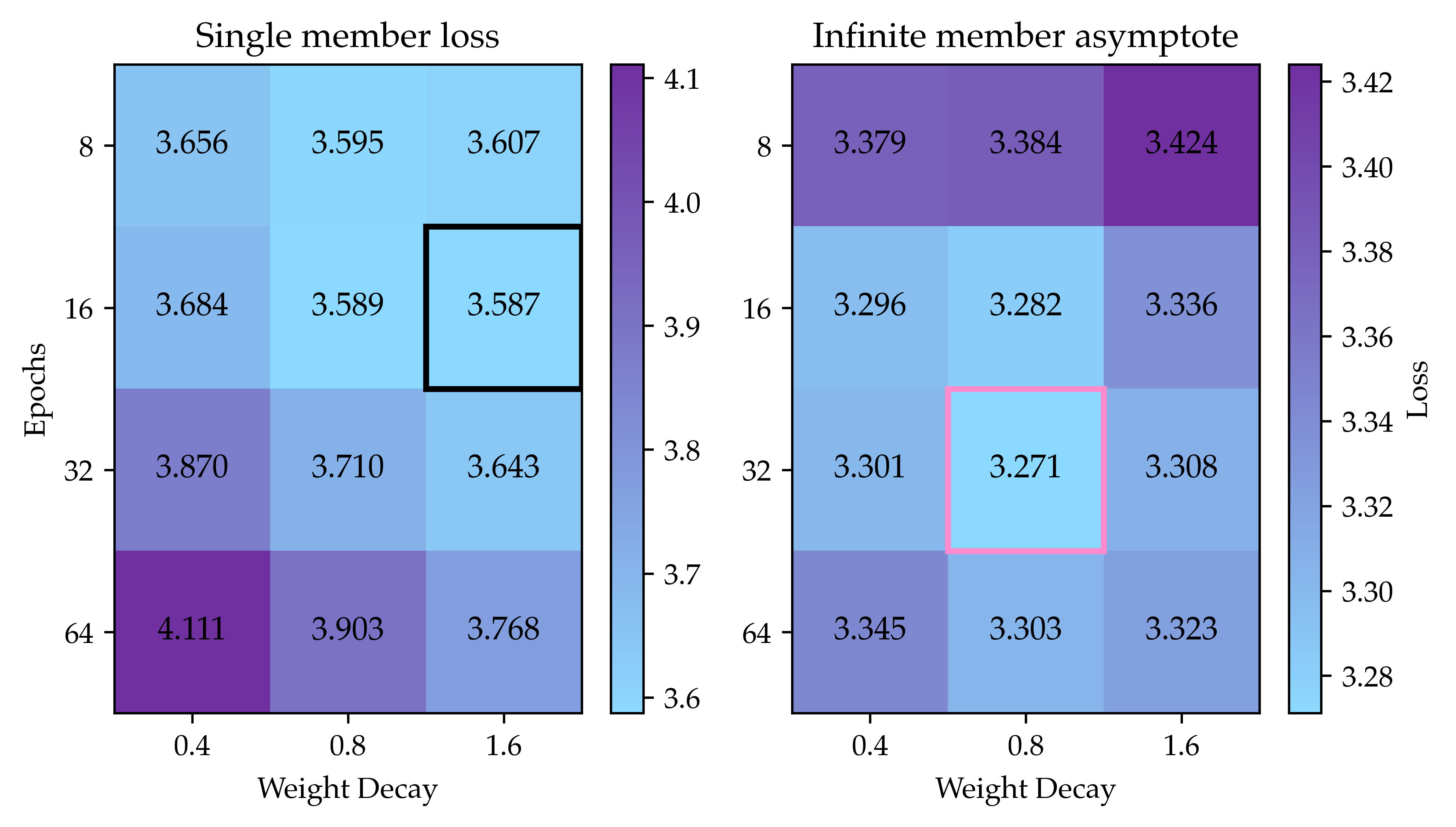}
    \caption{\textbf{Tuning hyperparameters of ensemble members for lowest asymptote under $K\to\infty$.} We construct ensembles for different $K$ when varying epoch count and weight decay. We find that the ranking between hyperparameters changes across $K$ (left) and that the infinite member asymptote benefiting from more epochs and less weight decay per member.}
    \label{fig:sketch-ensemble-hparams}
\end{figure}

Given the success of ensembles, we study how to tune their hyperparameters. In Figure~\ref{fig:sketch-ens-vs-model-scale}, the ensemble members were trained using the best hyperparameters for a single model, but this may not be optimal for large $K$ ensembles. More formally,
$$
\underbrace{\argmin_{H} \loss\p{\trainalg(D, N, H)} = \argmin_{H} \loss\p{\ensalg_{\trainalg}(D, N, K = 1, H)}}_{\text{best for single model}} \neq \underbrace{\argmin_{H} \lim_{K\to\infty}  \loss\p{\ensalg_{\trainalg}(D, N, K, H)}}_{\text{best for ensemble member asymptote}}
$$

In Figure~\ref{fig:sketch-ensemble-hparams}, we train ensembles for $K \in [5]$ as we vary weight decay and epoch count\footnote{We find that optimal learning rate is generally consistent across $K$, Appendix~\ref{app:heuristic-hparams}}. To estimate the best hyperparameters as $K\to\infty$, we fit a power law and refer to the asymptote. We find that the ranking between hyperparameters changes depending on $K$ (Figure~\ref{fig:sketch-ensemble-hparams}, left). Importantly, the optimal hyperparameters for $K=1$ (black) are not the best asymptote under $K\to\infty$ (pink), which benefits from more epochs and less weight decay. This intuitively corresponds to each member being more overfit to the data, which might be helpful to learn different views of the data. Selecting the hyperparameters when considering $K\to\infty$ improves the ensembling asymptote from $3.34$ to $3.27$ (Figure~\ref{fig:sketch-ensemble-hparams}, right).

Another design choice for our ensembles is the source of randomness $Z_i$. Though we vary both the data order and model initialization, we find that either one by itself obtains most of the benefits of ensembling and reserve a full analysis to Appendix~\ref{app:seed-science}. 

\subsection{Joint scaling recipe composing parameter and ensemble scaling}\label{sec:infinite-ensembles}

\begin{figure}
    \centering
    \includegraphics[height=0.4\textwidth]{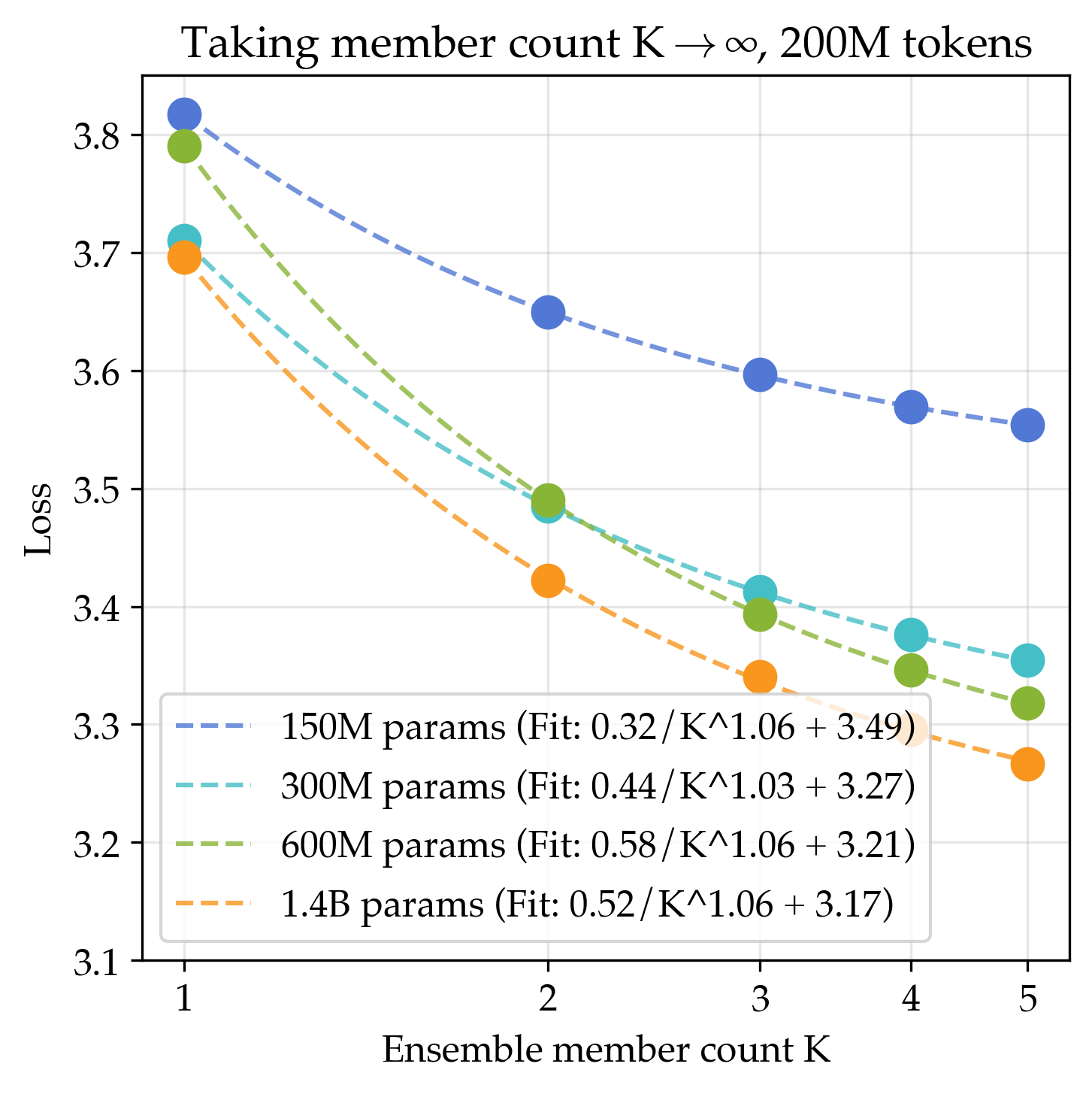}
    \includegraphics[height=0.4\textwidth]{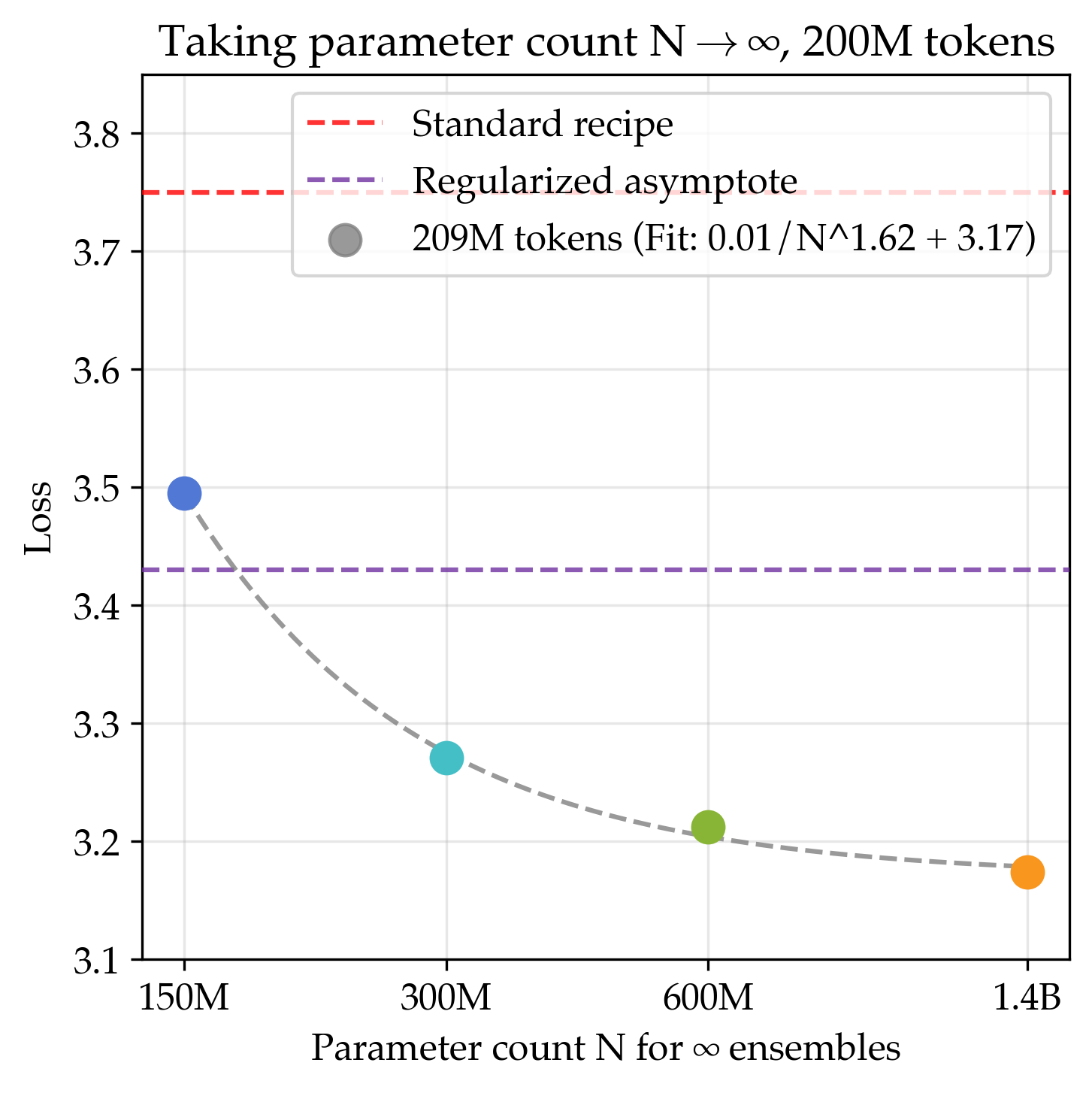}
    \caption{\textbf{Composing the regularized recipe and ensembling recipe under the double limit.} Left: For each parameter count, we fit a power law on the loss as $K$ increases. We select hyperparameters that result in low asymptotes instead of loss at small $K$. Right: We take the asymptotes from the left plot and fit a power law to capture how the asymptote changes for ensembles of larger models. This asymptote estimates the best possible loss under the joint scaling recipe.}
    \label{fig:ensemble-double-limit}
\end{figure}

Although the ensembling recipe outperforms parameter scaling, we can compose both by taking the number of members and the size of each member to infinity ($N,K\to\infty$). To estimate the best possible loss of a joint scaling recipe, we take two limits:
$$
\hat{\loss}_D = \lim_{N\to\infty} \lim_{K\to\infty} \min_{\substack{H}} \loss\p{\ensalg_{\trainalg}\p{D, N, K, H}}
$$
As long as $\min_{\substack{H}} \loss\p{\ensalg_{\trainalg}\p{D, N, K, H}}$ monotonically decreases in $N$ and $K$ when the other variable is fixed, the value does not depend on the order we take the limits. We choose this particular order since we found it results in the most convenient hyperparameter tuning (discussed in Appendix~\ref{app:order-of-limits}). Due to pragmatic experimental constraints, we cannot search for locally optimal hyperparameters in the same way we did for parameter scaling. We instead estimate the best ensemble hyperparameters via the heuristic of the locally optimal regularized hyperparameters with $2\times$ epochs and $0.5\times$ weight decay, since we found this works best across almost all of our scales (Appendix~\ref{app:heuristic-hparams}).

In Figure~\ref{fig:ensemble-double-limit}, we show how we take this double limit. On the left, we vary $K$ for different fixed parameter counts $N$, fitting power laws that estimate loss as $K$ increases. We then take the asymptotes of these four power laws and plot them on the right with respect to $N$. Though the asymptote of 150M ensembles is actually worse than the regularized asymptote, the ensemble asymptote decreases as $N$ increases, beating the regularized asymptote for sufficiently large $N$. We then fit a second power law, allowing us to extrapolate how the estimated asymptote for $K\to\infty$ changes as we take $N\to\infty$. Our final estimate for the best possible loss is $3.17$, which is a significant improvement over the loss of $3.43$ for the regularized pre-training recipe and $3.75$ for the unregularized recipe.

\section{Scaling the seed token count under infinite compute}
\label{sec:scaling-seed-token}

Our previous experiments show large improvements in loss 
for 200M tokens and we ask whether these findings generalize to larger token counts. To test this, we measure the best possible loss of the standard recipe, regularized recipe, and joint scaling recipe at higher token counts (Sections~\ref{subsec:unregularized-law},~\ref{subsec:param-count},~\ref{subsec:ensemble}). To contextualize the loss improvements, we measure the \textbf{data efficiency} of a recipe by interpolating how much data the standard recipe would need to match its asymptote. At 200M seed tokens, we estimate that the regularized recipe and joint scaling recipe are $2.29\times$ and $5.17\times$ more data efficient than the standard recipe. We then fit data-scaling laws to our estimates of the best possible loss to extrapolate how recipes would perform for higher seed token counts. Although our tiered data scaling laws are bound to be noisy, they share similar exponents and asymptotes, which would imply that the data efficiency improvement is constant across all data scales (Section~\ref{subsec:scaling-data}).

\subsection{Tuning the standard recipe}\label{subsec:unregularized-law}

We first detail how we construct the data scaling law for the standard recipe. For a given seed token count $D$, we estimate the best possible loss by tuning learning rate, epoch count, and parameter count. As we found in Section~\ref{sec:standard-pt}, increasing parameter count eventually results in validation loss increasing so we cannot build monotonic scaling laws. Therefore, we search for the best parameter count and hyperparameters for 200M, 400M, 800M, and 1.6B tokens (detailed in Appendix~\ref{app:epoch-data-scaling}). These four losses form the red points in Figure~\ref{fig:seed-token-scaling-standard}, right. Given these four estimates of the best possible loss at each token count, we can fit the data scaling power law shown as the red line using 
$$\hat{\loss}_{D} \coloneq \frac{A}{D^{\alpha}} + E$$

\subsection{Scaling parameter count}\label{subsec:param-count}

\begin{figure}
    \centering
    \includegraphics[height=0.4\textwidth]{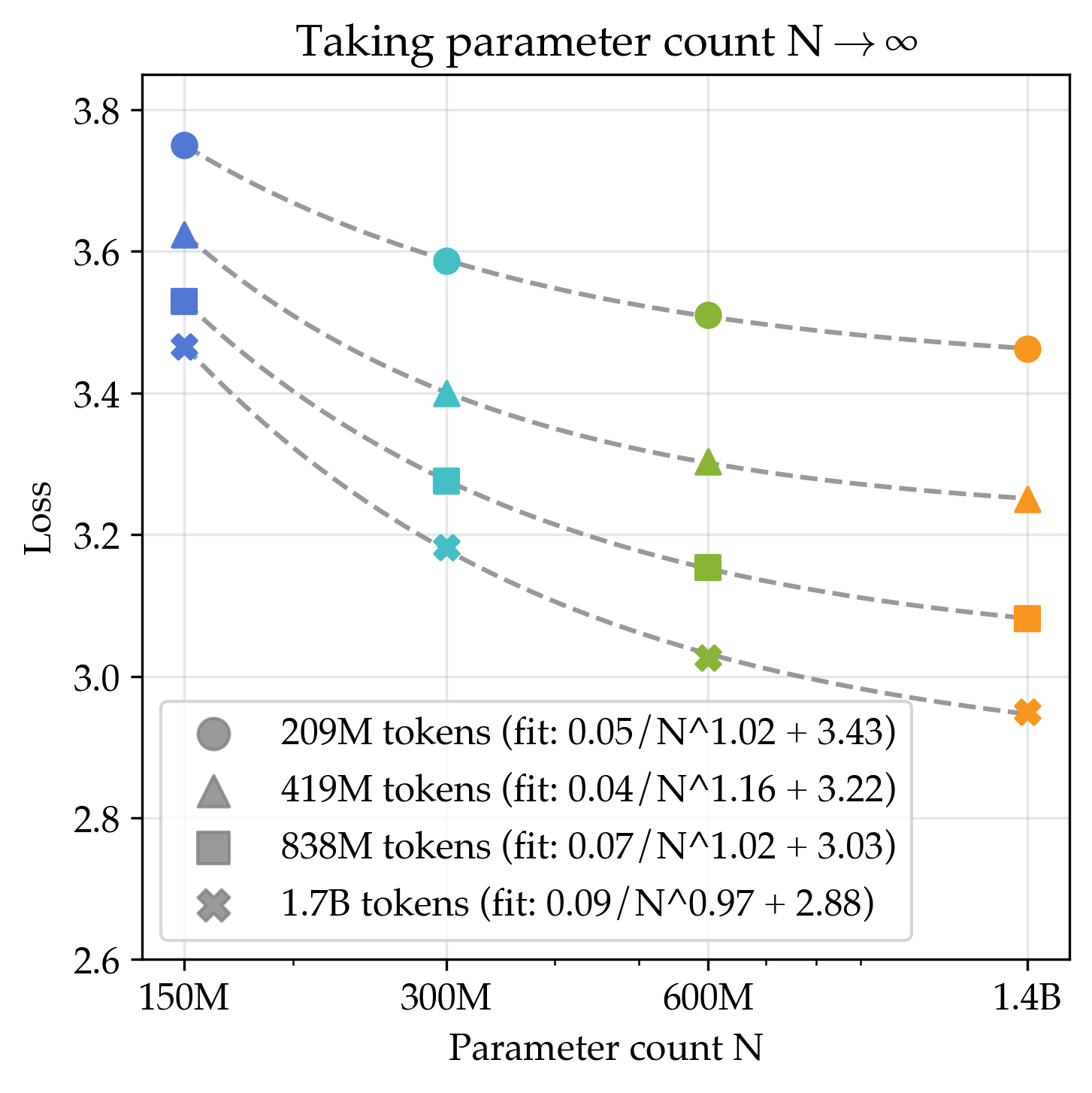}
    \includegraphics[height=0.4\textwidth]{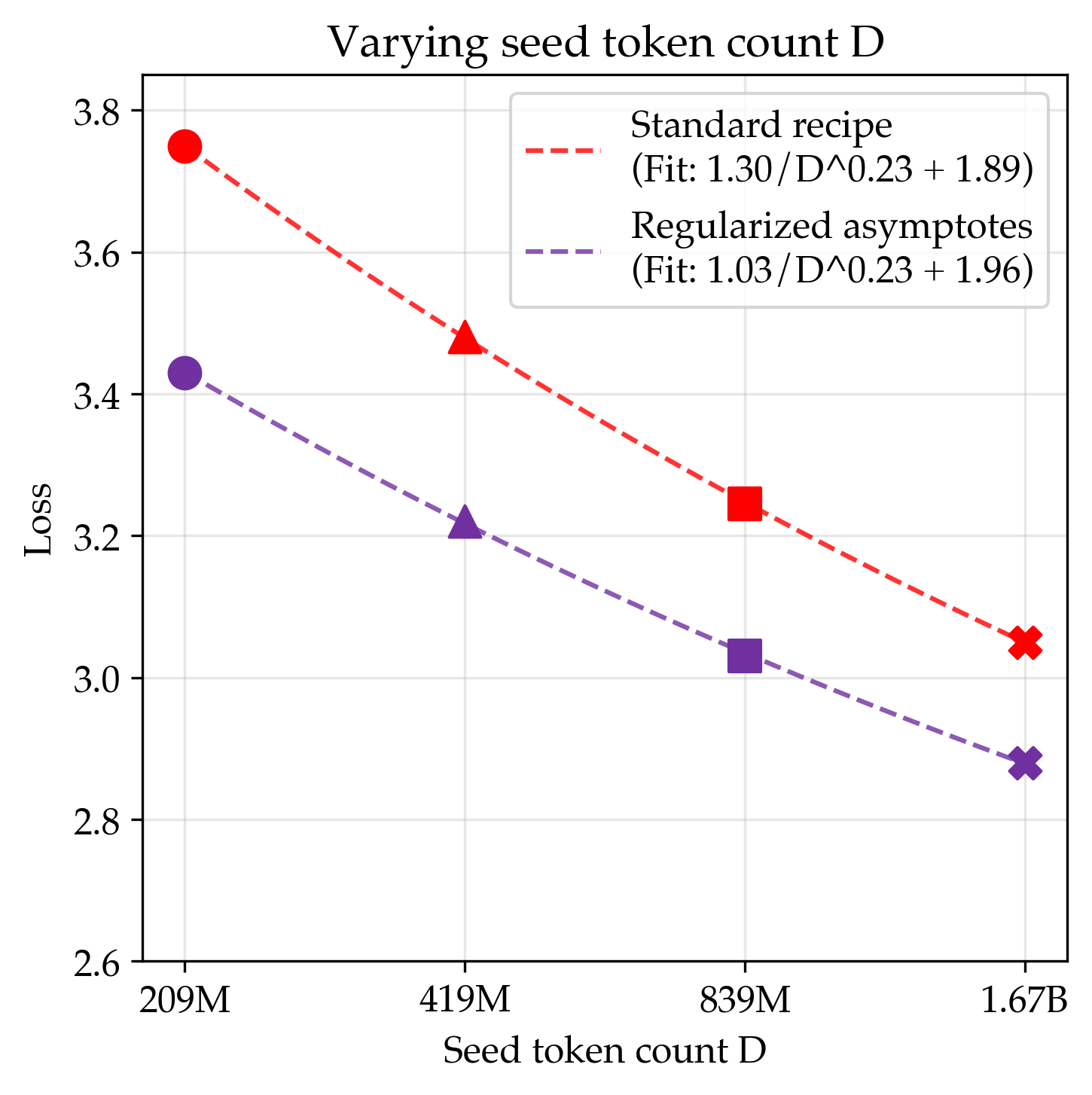}
    \caption{\textbf{Scaling the seed token count for single model pre-training.} We first consider the best possible loss of the standard recipe tuning epochs and parameters (red, right). We then consider the best possible loss of the regularized recipe by fitting parameter scaling laws across four token counts, shown on the left. The asymptotes of these laws form the purple points on the right. We further fit data-scaling laws to extrapolate performance as seed token count increases.}
    \label{fig:seed-token-scaling-standard}
\end{figure}

We now characterize the best possible loss of the regularized recipe by estimating $\lim_{N\to\infty} \min_{\substack{H}} \loss\p{\trainalg\p{D, N, H}}$ via the asymptote of the power law as described in Section~\ref{sec:fix-standard-pt}. Since we now have to compute asymptotes to build the points for the data scaling law, we follow a two step procedure as visualized in Figure~\ref{fig:seed-token-scaling-standard}.
\begin{enumerate}
    \item \textbf{Varying parameter count (left plot).} We first follow the locally-optimal hyperparameter search detailed in Section~\ref{sec:fix-standard-pt} to find the best possible models for four parameter counts across four seed token counts. These are the 16 points in Figure~\ref{fig:seed-token-scaling-standard}, left. 
    \item \textbf{Varying token count (right plot).} For each seed token count, we can fit a power law and use the asymptote $E_D$ as an estimate of $\loss^*_{D}$. These four $(D, E_D)$ tuples form the purple points in Figure~\ref{fig:seed-token-scaling-standard}, right. We then fit a data scaling power law over these asymptotes, shown as the purple line.
\end{enumerate}

\paragraph{Measuring data efficiency.} Since we are data-constrained, we care about the loss improvement at a fixed value of $D$, unlike our compute scaling laws where we care about the asymptote. Visually inspecting the plot suggests that the standard recipe would need approximately twice as much data to match the performance of regularized scaling across all of our token scales. To formalize the data efficiency improvement of algorithm $\trainalg_2$ over algorithm $\trainalg_1$ at a token count $D$, we compute the effective data $D'$ that $\trainalg_1$ would need to match $\trainalg_2$. After interpolating $D'$ via the data scaling law of $\trainalg_1$, we report the data efficiency improvement as $\frac{D'}{D}$. This metric characterizes the asymptote of the regularized recipe as $2.29\times$ more data efficient than the standard recipe at 200M tokens. Since our tiered asymptote estimation can be unreliable, we note that even without any extrapolation, the best 1.4B model at 200M tokens is $2.09\times$ more data efficient than our baseline.

\subsection{Scaling member and parameter count}\label{subsec:ensemble}

\begin{figure}[b!]
    \centering
    \includegraphics[height=0.33\textwidth]{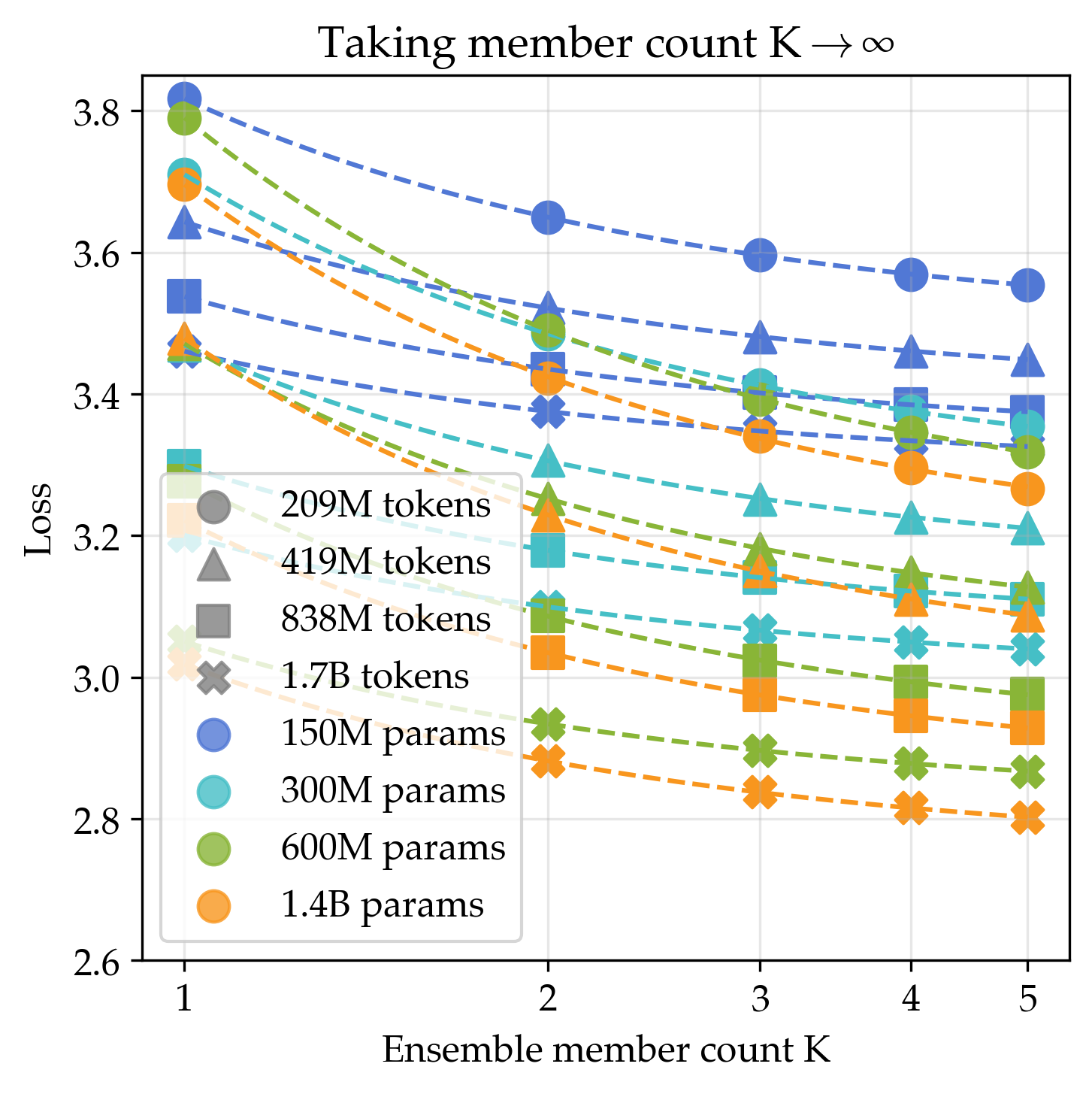}
    \includegraphics[height=0.33\textwidth]{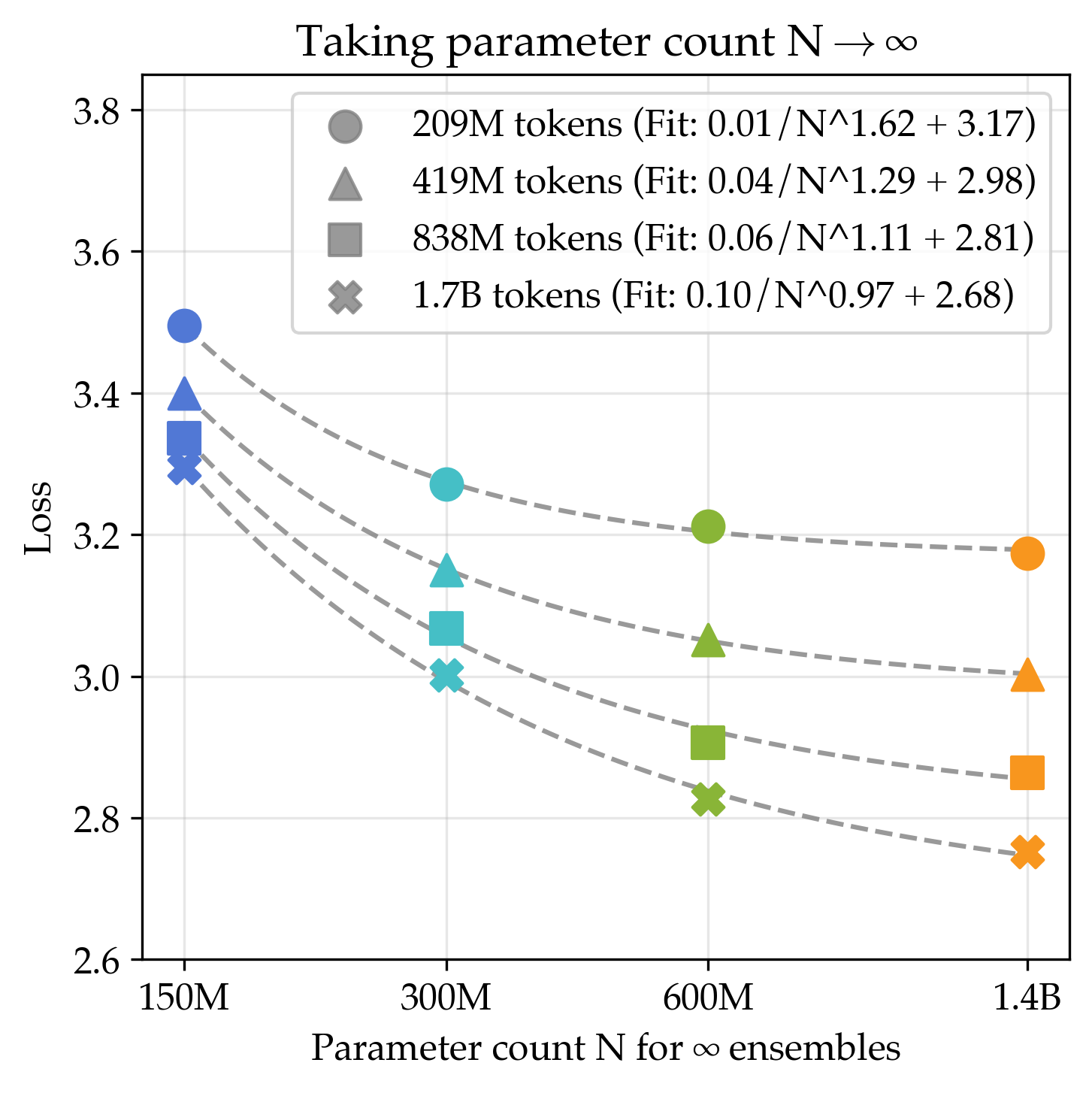}
    \includegraphics[height=0.33\textwidth]{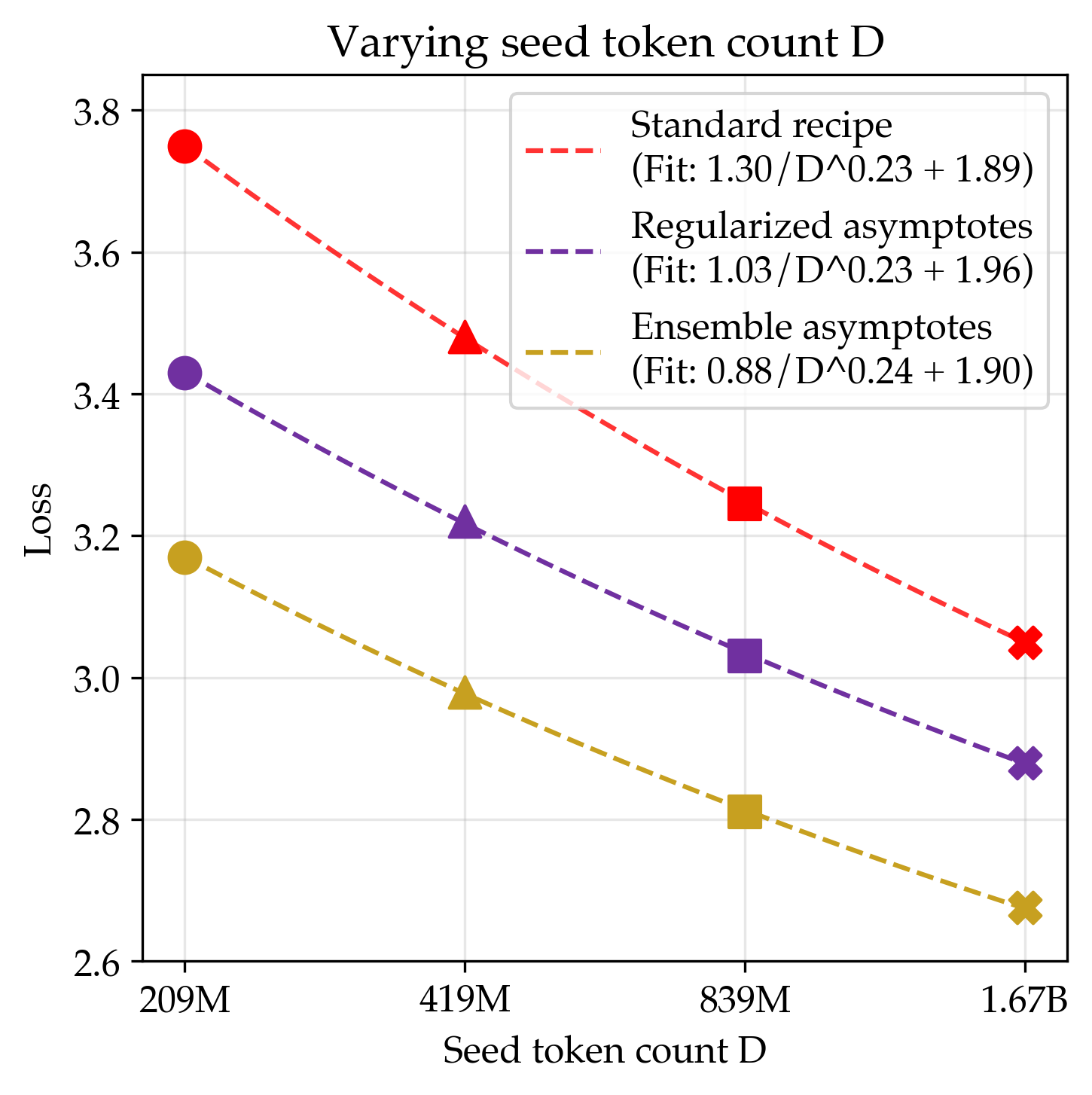}
    \caption{\textbf{Scaling the seed token count for ensembles.} Left: For a fixed parameter count and token count, we fit a power law in $K$, with hyperparameters optimized for the asymptote. Middle: We take the asymptote of the 16 laws on the left and fit a power law to measure how the asymptote changes in $N$. Right: We take the asymptote of the 4 laws in the middle and fit a power law to measure how the asymptote of asymptotes changes in $D$. We find over $2\times$ data efficiency wins over the regularized recipe and $5\times$ data efficiency wins over the standard recipe at all tested token counts.}
    \label{fig:seed-token-scaling-ensembling}
\end{figure}

We repeat the above procedure for ensembles by taking $\lim_{N\to\infty} \lim_{K\to\infty} \min_{\substack{H}} \loss\p{\ensalg_{\trainalg}\p{D, N, K, H}}$ for each seed token count $D$ following Section~\ref{sec:infinite-ensembles}. This results in the following three step procedure, visualized in Figure~\ref{fig:seed-token-scaling-ensembling}.

\begin{enumerate}
    \item \textbf{Varying parameter count (left plot).} In Figure~\ref{fig:seed-token-scaling-ensembling}, left, we picture the losses of ensembles across our 4 token counts, 4 parameter counts, and 5 member counts with hyperparameters selected for better asymptotes. For each of the 16 pairs of $D$ and $N$, we use a power law to extrapolate the performance as we take member count $K\to\infty$.
    \item \textbf{Varying ensemble member count (middle plot).} The asymptotes of the above 16 limits are visualized in Figure~\ref{fig:seed-token-scaling-ensembling}, middle. Given these asymptotes, we can fit a second set of 4 power laws that extrapolate the loss as we take parameter count $N\to\infty$.
    \item \textbf{Varying token count (right plot).} The asymptotes of these 4 limits are visualized by the gold points on Figure~\ref{fig:seed-token-scaling-ensembling}, right. We then fit a data scaling law, depicted by the gold line.
\end{enumerate}

At 200M tokens, the asymptote of the joint scaling recipe is $5.17\times$ more data efficient than the standard recipe. Without taking asymptotes, our best ensemble of five 1.4B models is itself $3.75\times$ more data efficient.

\subsection{Scaling data}\label{subsec:scaling-data}

We compare the data scaling laws across our three recipes to measure whether the data efficiency improvements persist across token count. Though the data scaling laws are expected to be noisy, these laws predict that all three scaling recipes decay at a similar rate with exponents between $0.23$ and $0.24$ and asymptotes between $1.89$ and $1.96$. Asymptotic statistics suggests that the asymptotes are equal if the algorithms achieve Bayes-optimal error under infinite data and compute, in which case their loss would be the entropy of text~\citep{shannon1951prediction,van2000asymptotic}. In the case where the asymptote $E$ and exponent $\alpha$ of the laws are the same for algorithms $\trainalg_1,\trainalg_2$, there is a constant data efficiency improvement at all token counts determined by the numerators $A_1, A_2$, equal to $\p{{A_2} / {A_1}}^{\frac{1}{\alpha}}$. Our preliminary analysis suggests that the our data efficiency improvements will not disappear across all data scales even if they perform similarly under infinite data.


\section{Data efficiency under parameter constraints}
\label{sec:distillation}

The asymptotes of the regularized and ensembling recipes rely on arbitrarily high parameter models. For example, the smallest ensemble that achieves a loss of 3.37 needs 1.2B total parameters. In this section, we investigate whether high parameter counts are necessary for data efficiency, either for the final model or for training. In Section~\ref{sec:reducing-parameters}, we show that we can distill an ensemble of 2.4B total parameters into a 300M student, preserving $83\%$ of the loss improvement without increasing the parameter count of the final model. In Section~\ref{sec:self-distillation}, we show that by self-distilling a 300M teacher model into a student model of the same parameter count and architecture, we can train a student that outperforms the teacher, removing the need for large parameter counts at training.



\begin{figure}
    \centering
    \includegraphics[width=0.6\textwidth]{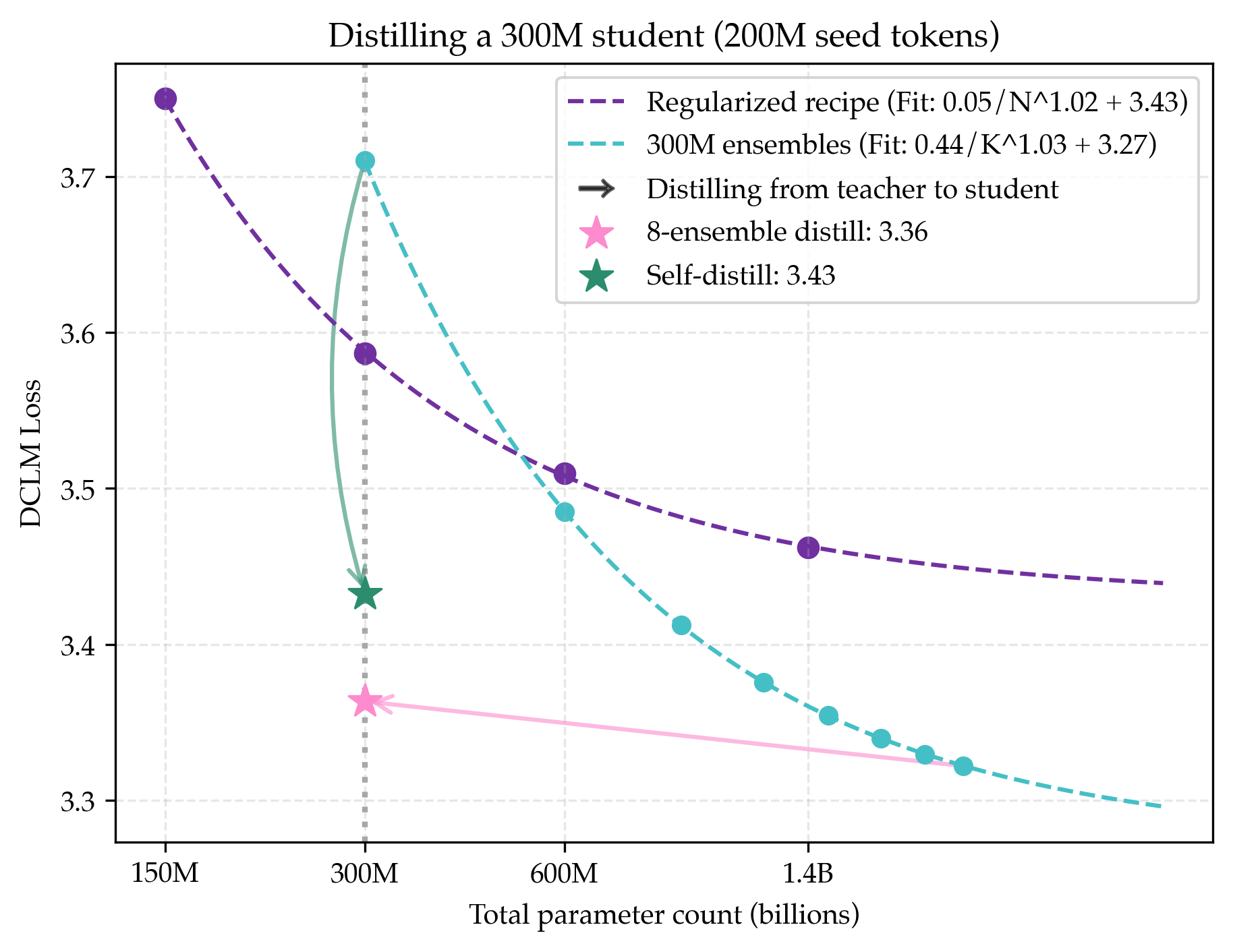}
    \caption{\textbf{Ensemble distillation and self-distillation.} We can compress our data efficiency gains into smaller models through distillation. Distilling an 8-ensemble teacher into a 300M student retains most of the loss improvement (pink star) and outperforms the asymptote of model scaling. We also find that self-distillation with a 300M teacher and 300M student (green star) is surprisingly effective, matching the asymptote of the regularized recipe without increasing parameter count at training.}
    \label{fig:sketch-distill}
\end{figure}

\subsection{Reducing final parameter count via ensemble distillation}\label{sec:reducing-parameters}

So far, the benefit of ensembling is only realized at a high enough parameter count. Therefore, even if our best scaling recipe helps in the limit as $N,K\to\infty$, it is not immediately clear whether the scaling recipe helps train models that are small relative to $D$.
However, it is known that better large models can improve the performance of smaller models through knowledge distillation~\citep{hinton2015distillingknowledgeneuralnetwork}. In fact, many smaller 
models today are pre-trained through distilling large models~\citep{yang2025qwen3technicalreport, gemmateam2025gemma3technicalreport,grattafiori2024llama3herdmodels, goyal2025distilledpretrainingmodernlens}. Since we are not bound by train compute, we can first pre-train a data-efficient teacher $M'$ using our existing recipes and then distill $M'$ to train a student $M$ via sequence-level knowledge distillation~\citep{kim2016sequencelevelknowledgedistillation}:
\begin{enumerate}
    \item Train a teacher model $M'$ on $D$ tokens.
    \item Sample from $M'$ unconditionally (i.e. with no prompt) to generate a dataset of $D'$ tokens.
    \item Train a student model $M$ from scratch on the mixture of $D$ and $D'$.
\end{enumerate}
In Figure~\ref{fig:sketch-distill}, we show the student model (pink star) obtained from using an 8-ensemble of 300M models (right-most blue point) with loss $3.32$. Despite the $8\times$ smaller inference compute budget, our distillation procedure attains a loss of $3.36$, preserving $83\%$ of the ensemble improvement over the regularized 300M model loss of $3.57$ (purple point). Our student outperforms the regularized recipe asymptote and even matches the loss of a 4-ensemble of 300M models (details in Appendix~\ref{app:distill}). 

\subsection{Reducing train parameter count via self-distillation}\label{sec:self-distillation}

The success of ensemble distillation shows how the final model does not need to have a high parameter count and pushes the loss-vs-parameter tradeoff to the left. Is it possible to train a small model (i.e. 300M) better than our recipes without high parameter count at train time as well? We consider this question in the context of self-distillation where the teacher and student are of the same size and architecture. A priori, it seems impossible for a student to outperform its teacher following arguments such as the data processing inequality. In fact, recent papers discuss how training a new student model on model generations can result in model degradation via model collapse~\citep{shumailov2024ai,gerstgrasser2024modelcollapseinevitablebreaking,dohmatob2024strongmodelcollapse,taori2022datafeedbackloopsmodeldriven}.

On the contrary, we find that correctly distilling a 300M teacher into a fresh student of the same architecture can \emph{improve} loss. Crucially, by mixing together the $D$ real tokens and $D'$ synthetic tokens, we find that we avoid collapse and train a student that vastly outperforms its teacher. In Figure~\ref{fig:sketch-distill}, we show how using a single 300M model as a teacher (blue point) results in a 300M student model (green star) that outperforms the best 300M model from standard pre-training (purple point). 

Why does self-distillation help?\cite{allenzhu2023understandingensembleknowledgedistillation} provide theory interpreting self-distillation as implicitly ensembling the teacher and the freshly initialized student. The connection between ensembling and self-distillation is further reflected in their strong empirical performance. Beyond just our experiments, recent work suggests that synthetic data augmented from the original pre-training data can provide data efficiency wins~\citep{maini2024rephrasingwebrecipecompute,allenzhu2024physicslanguagemodels31,su2025nemotroncctransformingcommoncrawl,kimiteam2025kimik2openagentic,yang2024syntheticcontinuedpretraining,ruan2025reasoninglearnlatentthoughts}. We view self-distillation as a different type of synthetic data that does not require human priors such as hand-crafted invariances, reward signals, or prompting strategies, making it appealing for scalable data-constrained pre-training.



\section{Downstream tasks}\label{sec:downstream}


\begin{figure}
    \centering
    \includegraphics[height=0.38\textwidth]{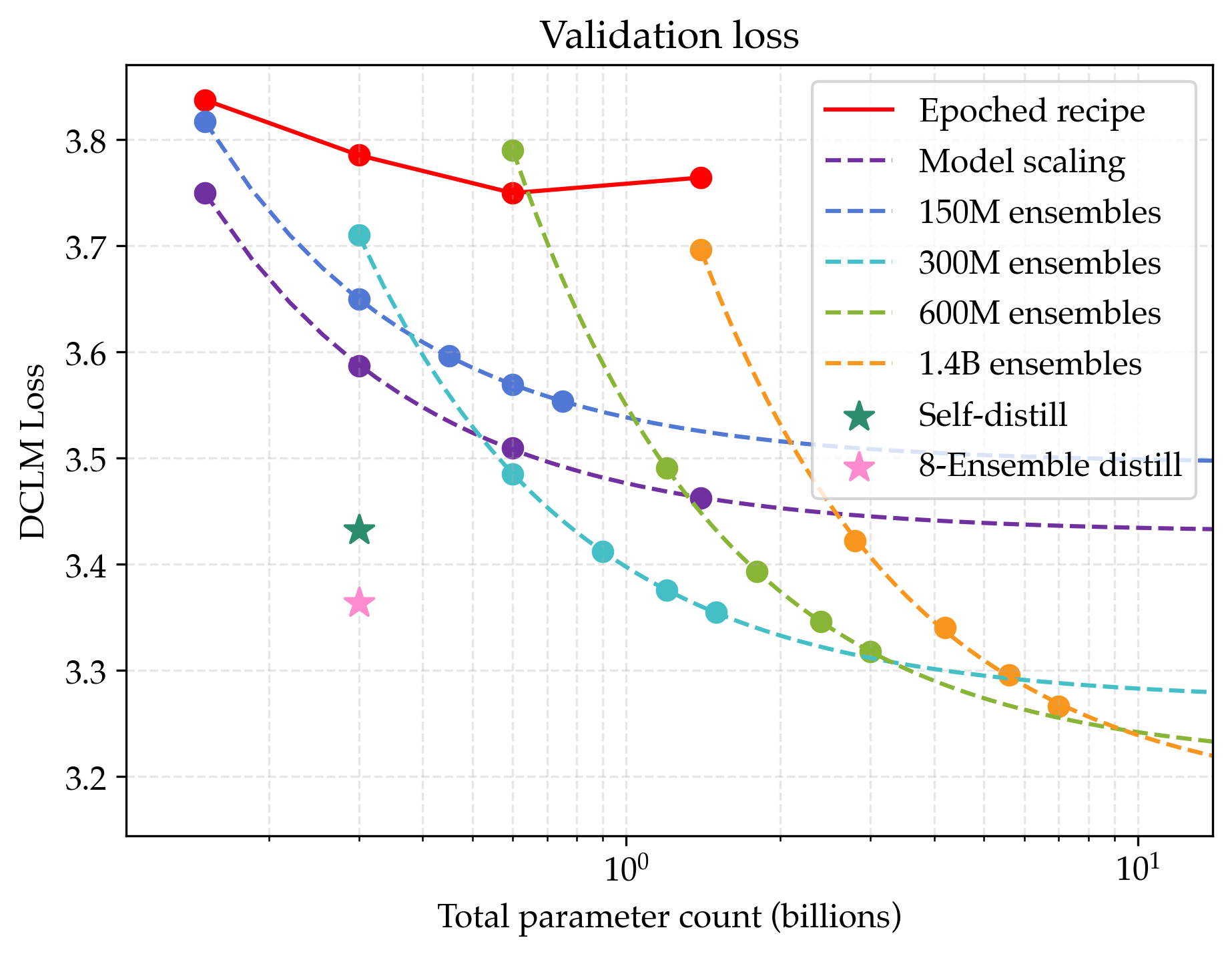}
    \includegraphics[height=0.38\textwidth]{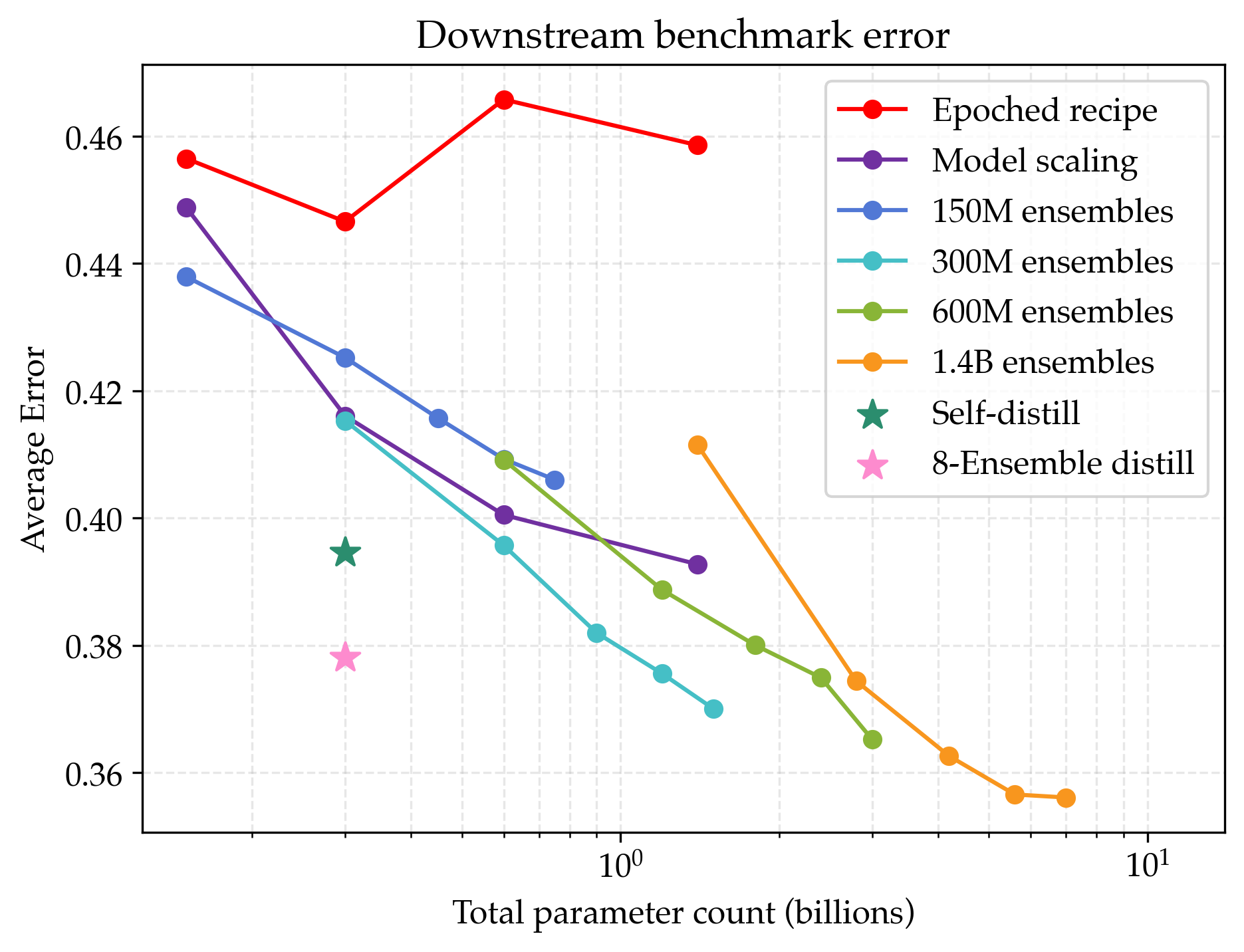}
    \caption{\textbf{Performance of pre-trained models on downstream tasks.} We have thus far been using validation loss (left) to seperate whether models are better pre-trained models or not. We evaluate the same models and ensembles on downstream benchmarks (right). Models with lower validation loss have lower average error across downstream benchmarks.}
    \label{fig:sketch-benchmarks}
\end{figure}

\subsection{Downstream benchmarks}\label{subsec:downstream-benchmarks}
So far, we have only analyzed pre-training validation loss as a proxy for model quality. Although loss is known to correlate with capabilities of interest \citep{chen2025scalinglawspredictingdownstream,thrush2025improvingpretrainingdatausing, gadre2024languagemodelsscalereliably}, it may not be perfectly reflective. Therefore, we test our model's general capabilities using downstream benchmarks. Since we need evaluations that are informative for models at our scale, we take all the accuracy-based benchmarks from~\cite{thrush2025improvingpretrainingdatausing}, namely PIQA~\citep{bisk2019piqareasoningphysicalcommonsense}, SciQ~\citep{welbl2017crowdsourcingmultiplechoicescience}, and ARC Easy~\citep{clark2018thinksolvedquestionanswering}. Notably, we did not evaluate on \emph{any} benchmarks until the end of the project after we selected the best recipes following validation loss, making these benchmarks a strong test of generalization.

In Figure~\ref{fig:sketch-benchmarks}, we show both the validation loss (left) and downstream benchmark error (right) of our pre-trained models, ensembles, and distilled models for 200M tokens. Without our intervention of regularization, the standard recipe does not strongly benefit from parameter scaling. Adding in regularization (purple points) makes downstream accuracy scale smoothly with diminishing returns, similar to validation loss. For ensembles, we observe similar trends as validation loss with increasing $N$ and $K$ improving performance. Furthermore, our distilled models improve upon all other 300M parameter models. Overall, our best ensemble outperforms our best unregularized model by over $9\%$ on average and our best distilled model outperforms the unregularized 300M model by around $7\%$, marking a large improvement over standard pre-training. A full breakdown of results is available in Appendix~\ref{app:downstream}. 

\subsection{Continued pre-training}\label{subsec:cpt}

We demonstrate the immediate applicability of our findings in settings outside of pre-training from scratch by improving the data efficiency of existing CPT recipes. We adopt the setup from~\cite{wang2025octothinkermidtrainingincentivizesreinforcement} of performing continued pre-training on Llama 3.2 3B Base~\citep{grattafiori2024llama3herdmodels} with the \texttt{MegaMath-Web-Pro} mid-training dataset. To simulate a data-constrained setting, we restrict ourselves to only 4B seed tokens of the full 73B tokens. We evaluate math performance by via accuracy on a subset of representative benchmarks from~\citet{wang2025octothinkermidtrainingincentivizesreinforcement}: GSM8K~\citep{cobbe2021trainingverifierssolvemath}, MATH~\citep{hendrycks2021measuringmathematicalproblemsolving}, and 
MathQA~\citep{amini2019mathqainterpretablemathword}. 

In Table~\ref{tab:cpt-evals}, we show that many of our results for pre-training from scratch transfer to continued pre-training setting. 
We start with a CPT baseline (Default) from the reference hyperparameters of~\citet{wang2025octothinkermidtrainingincentivizesreinforcement} which gives a $6.34\%$ improvement over the base model. We then apply our interventions from Appendix~\ref{app:hyper-ablations} and Section~\ref{sec:fix-standard-pt} by decreasing batch size and epoching, and we find that weight decay was not helpful for CPT. Our final single model baseline provides an additional $5.23\%$ lift in average accuracy on top of the reference hyperparameters. See Appendix~\ref{app:cpt} for details. 

We further show that ensembling eight epoched models provides an additional $4.76\%$ gain in average accuracy over just a single epoched CPT model ($K = 1$). We observe that average accuracy scales with increasing ensemble member count, and our best ensembles exceed the performance of a baseline continually pre-trained on the full $73$B tokens, providing a $17.5\times$ data efficiency win.
 \renewcommand{\arraystretch}{1.2}
 \begin{table}[h!]
 \centering
 \caption{\textbf{Data efficiency improvements on OctoThinker.} We take reasoning mid-training data from~\cite{wang2025octothinkermidtrainingincentivizesreinforcement} and apply our data efficiency interventions of reducing batch size (Appendix~\ref{app:hyper-ablations}), epoching data, and ensembling multiple models. Our best ensemble utilizing 4B tokens outperforms vanilla CPT on 73B tokens following the original paper's training hyperparameters, resulting in a 17.5$\times$ data efficiency improvement.}
 \vspace{1.5mm}
 \begin{adjustbox}{max width=\linewidth}
 \begin{tabular}{l|c|ccc|ccc|c}
 \hline
\multirow{2}{*}{\textbf{Benchmarks}}
& \multirow{2}{*}{\textbf{Llama 3B}}
  & \multicolumn{3}{c|}{\textbf{CPT (4B tokens)}}
  & \multicolumn{3}{c|}{\textbf{$K$-ensembles}}
  & \multirow{2}{*}{\textbf{CPT (73B tokens)}} \\
\cline{3-5}\cline{6-8}
&
& Default
& Lower BS
& Epoching ($K=1$)
& $K=2$ & $K=4$ & $K=8$ & \\
\hline
$\text{GSM8K}_{\text{(8-shot)}}$  & 28.23 & 38.44 & 44.50 & 44.05 &  49.28 & 51.80 & \textbf{52.99} & 49.51 \\
$\text{MATH}_{\text{(4-shot)}}$    & 6.90 & 14.38 & 17.64 & 19.74 & 21.84 & 23.04 & \textbf{23.50} & 23.40 \\
$\text{MATHQA}_{\text{(8-shot)}}$  &  35.07 & 38.96  & 41.31 &42.58 & 45.12 & \textbf{46.06} & 45.26 & 44.79 \\
\hline
Average                 & 24.25 & 30.59 & 34.48 & 35.82 &  38.79 &  40.35  & \textbf{40.58} & 39.23 \\
\hline
\end{tabular}
\end{adjustbox}
\label{tab:cpt-evals}
\end{table}
\renewcommand{\arraystretch}{1.0}

\section{Related Work}\label{sec:related-work}

\paragraph{Scaling laws.}
Much of the success of modern language model pre-training was built upon scaling laws which accurately predict performance at a given resource budget~\citep{hestness2017deeplearningscalingpredictable,hestness2019humanlevelaccuracycomputationalchallenges,rosenfeld2019constructivepredictiongeneralizationerror,henighan2020scalinglawsautoregressivegenerative,kaplan2020scalinglawsneurallanguage,sorscher2023neuralscalinglawsbeating,hoffmann2022trainingcomputeoptimallargelanguage,ruan2024observationalscalinglawspredictability,NIPS1993_1aa48fc4}. Past work has studied the scaling properties of machine learning under various constraints such as data and compute~\citep{muennighoff2023scalingdataconstrainedlanguagemodels,goyal2024scalinglawsdatafiltering}, hardware precision~\citep{kumar2024scalinglawsprecision}, parameter count~\citep{sardana2025chinchillaoptimalaccountinginferencelanguage,springer2025overtrainedlanguagemodelsharder,gadre2024languagemodelsscalereliably}, and test-time compute~\citep{brown2024largelanguagemonkeysscaling,snell2024scalingllmtesttimecompute}. We first show that scaling laws in past work~\citep{muennighoff2023scalingdataconstrainedlanguagemodels} do not account for over-fitting of standard recipes with too many epochs and rectify this via regularization. We also show that even though early work in double descent suggests that over-parameterized deep learning does not have clean scaling laws due to double descent~\citep{Belkin_2019,hastie2020surpriseshighdimensionalridgelesssquares,nakkiran2019deepdoubledescentbigger}, we can get clean scaling via tuning regularization, agreeing with theory in over-parameterized regression~\citep{PhysRevX.6.031034,nakkiran2021optimalregularizationmitigatedouble,Canatar_2021,simon2024bettermodernmachinelearning}. Finally, we propose asymptote estimation of scaling laws as a new metric to evaluate the performance of pretrained models under infinite compute. 

\paragraph{Ensembling.}
Ensembling is a well-established algorithm that boosts performance across many settings~\citep{10.5555/648054.743935}. Past work has established the success of ensembling deep networks for uncertainty estimation 
\citep{lakshminarayanan2017simplescalablepredictiveuncertainty}, 
image classification~\citep{huang2017snapshotensemblestrain1,garipov2018losssurfacesmodeconnectivity}, and reinforcement learning~\citep{vanhasselt2015deepreinforcementlearningdouble}. Ensembling has been shown to have surprising connections to other successful techniques including distillation~\citep{allenzhu2023understandingensembleknowledgedistillation} and residual networks~\citep{veit2016residualnetworksbehavelike} and are shown to follow power laws~\citep{lobacheva2021powerlawsdeepensembles}. On the other hand, ensembling is sometimes not believed to outperform parameter scaling in certain theoretical models of scaling~\citep{vyas2023featurelearningnetworksconsistentwidths,ruben2024freelunchrandomfeature}. We show how the simplest form of ensembling can be adopted for pre-training and then build scaling laws to characterize their loss. See Appendix~\ref{app:alternatives} for further discussion on related alternatives including Mixture of Experts and weight-averaging. 


\paragraph{Distillation.} Distillation spends compute to produce strong models with lower inference and fine-tuning costs~\citep{hinton2015distillingknowledgeneuralnetwork}. Though we use sequence knowledge distillation~\citep{kim2016sequencelevelknowledgedistillation} as a preliminary demonstration, there are many better distillation algorithms such as using more supervision via logits~\citep{sanh2020distilbertdistilledversionbert} and minimizing different divergences between the teacher and student~\citep{agarwal2024onpolicydistillationlanguagemodels,gu2024minillmknowledgedistillationlarge}. Past work has further quantified the scaling properties of distillation~\citep{busbridge2025distillationscalinglaws}, and distillation is increasingly used to pre-train the most performant small language models 
\citep{goyal2025distilledpretrainingmodernlens, yang2025qwen3technicalreport, gemmateam2025gemma3technicalreport}. For self-distillation, there is recent work showing how training on self-generated inputs can be harmful~\citep{shumailov2024ai,dohmatob2024strongmodelcollapse,taori2022datafeedbackloopsmodeldriven}.~\cite{gerstgrasser2024modelcollapseinevitablebreaking} is the most optimistic work, suggesting that training on self-generated data can be helpful in limited scenarios, though their comparisons are neither compute-matched nor data-matched. We find that self-distillation can actually be helpful over optimal regularized pre-training if done correctly, aligned with prior evidence from data-constrained deep learning~\citep{mobahi2020selfdistillationamplifiesregularizationhilbert,zhang2019teacherimproveperformanceconvolutional}. Notably,~\cite{allenzhu2023understandingensembleknowledgedistillation} draws a connection between how self-distillation can be viewed as implicitly performing ensembling and distillation. We can interpret distillation as a form of synthetic data, and unlike recent work on synthetic data to improve data efficiency~\citep{maini2024rephrasingwebrecipecompute,allenzhu2024physicslanguagemodels31,su2025nemotroncctransformingcommoncrawl,kimiteam2025kimik2openagentic,yang2024syntheticcontinuedpretraining,ruan2025reasoninglearnlatentthoughts}, distillation requires minimal human priors such as a trusted reward function or known augmentation invariance. 

\paragraph{Classical data-constrained deep learning.}
Historically, many machine learning benchmarks before the era of large language models were data-constrained~\citep{marcus-etal-1993-building,warstadt2023papersbabylmchallenge, 5206848,726791}, resulting in many effective algorithms. For example, the best models on Penn Tree Bank utilized dynamic evaluation~\citep{inproceedings,krause2017dynamicevaluationneuralsequence}, ensembling and model averaging~\citep{takase2018directoutputconnectionhighrank,zaremba2015recurrentneuralnetworkregularization}, regularization (drop-out, weight decay, weight tying, lower batch size)~\citep{merity2017regularizingoptimizinglstmlanguage,zaremba2015recurrentneuralnetworkregularization,gal2016theoreticallygroundedapplicationdropout}, data augmentation~\citep{shi2021substructuresubstitutionstructureddata,xie2017datanoisingsmoothingneural}, and novel architectures~\citep{zilly2017recurrenthighwaynetworks,grave2016improvingneurallanguagemodels,yang2018breakingsoftmaxbottleneckhighrank}. We revisit a few of these algorithms and advocate for future work to explore all others.

\paragraph{Modern data-constrained pre-training.} There are several more recent works which study interventions for data-efficient pre-training. These works have shown the benefit of epoching~\citep{muennighoff2023scalingdataconstrainedlanguagemodels}, rephrased synthetic data~\citep{ maini2024rephrasingwebrecipecompute, yang2024syntheticcontinuedpretraining, datologyai2025beyondweblessonsscalingsynthetic,ruan2024observationalscalinglawspredictability}, diffusion language models~\citep{prabhudesai2025diffusionbeatsautoregressivedataconstrained, ni2025difflm}, and energy-based models~\citep{gladstone2025energybasedtransformersscalablelearners}. Some of these methods such as rephrased synthetic data have been adopted to train the strongest open-source models~\citep{kimiteam2025kimik2openagentic, yang2025qwen3technicalreport}. This collection of recent work does not fully leverage additional compute: most of the papers do not optimally epoch the models and none of the papers tune regularization, build scaling laws to estimate the infinite parameter limits, or build scaling laws as data increases to estimate data efficiency. We also hope that the performance gains we demonstrate through ensembling and self-distillation are orthogonal to those of existing work and can compose for better performance.
\section{Discussion}

Notably, the algorithms we have considered (parameter scaling, regularization, ensembling, distillation) mirror classical techniques from when deep learning was data-constrained by limited sentences~\citep{marcus-etal-1993-building,warstadt2023papersbabylmchallenge}, images~\citep{5206848,726791}, etc. The success of such simple methods suggests that there is free lunch on the table for data-efficient pre-training, encouraging us to revisit basic decisions such as objective~\citep{ni2025difflm,prabhudesai2025diffusionbeatsautoregressivedataconstrained}, architecture~\citep{gladstone2025energybasedtransformersscalablelearners}, and data augmentation~\citep{maini2024rephrasingwebrecipecompute}. Since available compute grows far more quickly than available data, we are excited by algorithms that can better leverage additional compute for better performance, in line with The Bitter Lesson~\citep{sutton2019bitterlesson}. We hope that by carefully evaluating scaling recipes via their asymptotes instead of their compute-constrained performance, we can design algorithms that are better prepared for a data-constrained future. 

\section{Acknowledgements}

We thank Steven Cao, Sam Park, Jacob Mitchell Springer, Kaiyue Wen, Yu Sun, Nathan Hu, Meena Jagadeesan, Luke Bailey, Neil Band, Sally Zhu, Ben Spector, and Audrey Xie for their helpful discussions or feedback on the paper draft.

This work is a part of the \href{marin.community}{Marin Project} and the compute is supported by the Google TPU Research Cloud (TRC). TH was supported by a grant by HAI, DSO labs, gifts from Open Philanthropy, Amazon, Schmidt Sciences, the Tianqiao and Chrissy Chen Foundation and a grant under the NSF CAREER IIS-2338866, ONR N00014-24-1-2609, and DARPA Cooperative Agreement HR00112520013. PL was supported by DARPA Cooperative Agreement HR00112520013. This work does not necessarily reflect the position or policy of the government and no official endorsement should be inferred.

\newpage
\bibliographystyle{abbrvnat}
\bibliography{references}

\begin{thebibliography}{109}
\providecommand{\natexlab}[1]{#1}
\providecommand{\url}[1]{\texttt{#1}}
\expandafter\ifx\csname urlstyle\endcsname\relax
  \providecommand{\doi}[1]{doi: #1}\else
  \providecommand{\doi}{doi: \begingroup \urlstyle{rm}\Url}\fi

\bibitem[Advani and Ganguli(2016)]{PhysRevX.6.031034}
M.~Advani and S.~Ganguli.
\newblock Statistical mechanics of optimal convex inference in high dimensions.
\newblock \emph{Phys. Rev. X}, 6:\penalty0 031034, Aug 2016.
\newblock \doi{10.1103/PhysRevX.6.031034}.
\newblock URL \url{https://link.aps.org/doi/10.1103/PhysRevX.6.031034}.

\bibitem[Agarwal et~al.(2024)Agarwal, Vieillard, Zhou, Stanczyk, Ramos, Geist, and Bachem]{agarwal2024onpolicydistillationlanguagemodels}
R.~Agarwal, N.~Vieillard, Y.~Zhou, P.~Stanczyk, S.~Ramos, M.~Geist, and O.~Bachem.
\newblock On-policy distillation of language models: Learning from self-generated mistakes, 2024.
\newblock URL \url{https://arxiv.org/abs/2306.13649}.

\bibitem[Ainsworth et~al.(2023)Ainsworth, Hayase, and Srinivasa]{ainsworth2023gitrebasinmergingmodels}
S.~K. Ainsworth, J.~Hayase, and S.~Srinivasa.
\newblock Git re-basin: Merging models modulo permutation symmetries, 2023.
\newblock URL \url{https://arxiv.org/abs/2209.04836}.

\bibitem[Allen-Zhu and Li(2023)]{allenzhu2023understandingensembleknowledgedistillation}
Z.~Allen-Zhu and Y.~Li.
\newblock Towards understanding ensemble, knowledge distillation and self-distillation in deep learning, 2023.
\newblock URL \url{https://arxiv.org/abs/2012.09816}.

\bibitem[Allen-Zhu and Li(2024)]{allenzhu2024physicslanguagemodels31}
Z.~Allen-Zhu and Y.~Li.
\newblock Physics of language models: Part 3.1, knowledge storage and extraction, 2024.
\newblock URL \url{https://arxiv.org/abs/2309.14316}.

\bibitem[Amini et~al.(2019)Amini, Gabriel, Lin, Koncel-Kedziorski, Choi, and Hajishirzi]{amini2019mathqainterpretablemathword}
A.~Amini, S.~Gabriel, P.~Lin, R.~Koncel-Kedziorski, Y.~Choi, and H.~Hajishirzi.
\newblock Mathqa: Towards interpretable math word problem solving with operation-based formalisms, 2019.
\newblock URL \url{https://arxiv.org/abs/1905.13319}.

\bibitem[Belkin et~al.(2019)Belkin, Hsu, Ma, and Mandal]{Belkin_2019}
M.~Belkin, D.~Hsu, S.~Ma, and S.~Mandal.
\newblock Reconciling modern machine-learning practice and the classical bias–variance trade-off.
\newblock \emph{Proceedings of the National Academy of Sciences}, 116\penalty0 (32):\penalty0 15849–15854, July 2019.
\newblock ISSN 1091-6490.
\newblock \doi{10.1073/pnas.1903070116}.
\newblock URL \url{http://dx.doi.org/10.1073/pnas.1903070116}.

\bibitem[Besiroglu et~al.(2024)Besiroglu, Erdil, Barnett, and You]{besiroglu2024chinchillascalingreplicationattempt}
T.~Besiroglu, E.~Erdil, M.~Barnett, and J.~You.
\newblock Chinchilla scaling: A replication attempt, 2024.
\newblock URL \url{https://arxiv.org/abs/2404.10102}.

\bibitem[Bisk et~al.(2019)Bisk, Zellers, Bras, Gao, and Choi]{bisk2019piqareasoningphysicalcommonsense}
Y.~Bisk, R.~Zellers, R.~L. Bras, J.~Gao, and Y.~Choi.
\newblock Piqa: Reasoning about physical commonsense in natural language, 2019.
\newblock URL \url{https://arxiv.org/abs/1911.11641}.

\bibitem[Brown et~al.(2024)Brown, Juravsky, Ehrlich, Clark, Le, Ré, and Mirhoseini]{brown2024largelanguagemonkeysscaling}
B.~Brown, J.~Juravsky, R.~Ehrlich, R.~Clark, Q.~V. Le, C.~Ré, and A.~Mirhoseini.
\newblock Large language monkeys: Scaling inference compute with repeated sampling, 2024.
\newblock URL \url{https://arxiv.org/abs/2407.21787}.

\bibitem[Brown et~al.(2020)Brown, Mann, Ryder, Subbiah, Kaplan, Dhariwal, Neelakantan, Shyam, Sastry, Askell, Agarwal, Herbert-Voss, Krueger, Henighan, Child, Ramesh, Ziegler, Wu, Winter, Hesse, Chen, Sigler, Litwin, Gray, Chess, Clark, Berner, McCandlish, Radford, Sutskever, and Amodei]{brown2020languagemodelsfewshotlearners}
T.~B. Brown, B.~Mann, N.~Ryder, M.~Subbiah, J.~Kaplan, P.~Dhariwal, A.~Neelakantan, P.~Shyam, G.~Sastry, A.~Askell, S.~Agarwal, A.~Herbert-Voss, G.~Krueger, T.~Henighan, R.~Child, A.~Ramesh, D.~M. Ziegler, J.~Wu, C.~Winter, C.~Hesse, M.~Chen, E.~Sigler, M.~Litwin, S.~Gray, B.~Chess, J.~Clark, C.~Berner, S.~McCandlish, A.~Radford, I.~Sutskever, and D.~Amodei.
\newblock Language models are few-shot learners, 2020.
\newblock URL \url{https://arxiv.org/abs/2005.14165}.

\bibitem[Busbridge et~al.(2025)Busbridge, Shidani, Weers, Ramapuram, Littwin, and Webb]{busbridge2025distillationscalinglaws}
D.~Busbridge, A.~Shidani, F.~Weers, J.~Ramapuram, E.~Littwin, and R.~Webb.
\newblock Distillation scaling laws, 2025.
\newblock URL \url{https://arxiv.org/abs/2502.08606}.

\bibitem[Canatar et~al.(2021)Canatar, Bordelon, and Pehlevan]{Canatar_2021}
A.~Canatar, B.~Bordelon, and C.~Pehlevan.
\newblock Spectral bias and task-model alignment explain generalization in kernel regression and infinitely wide neural networks.
\newblock \emph{Nature Communications}, 12\penalty0 (1), May 2021.
\newblock ISSN 2041-1723.
\newblock \doi{10.1038/s41467-021-23103-1}.
\newblock URL \url{http://dx.doi.org/10.1038/s41467-021-23103-1}.

\bibitem[Chen et~al.(2025)Chen, Huang, Gao, Wang, Yang, and Ji]{chen2025scalinglawspredictingdownstream}
Y.~Chen, B.~Huang, Y.~Gao, Z.~Wang, J.~Yang, and H.~Ji.
\newblock Scaling laws for predicting downstream performance in llms, 2025.
\newblock URL \url{https://arxiv.org/abs/2410.08527}.

\bibitem[Clark et~al.(2018)Clark, Cowhey, Etzioni, Khot, Sabharwal, Schoenick, and Tafjord]{clark2018thinksolvedquestionanswering}
P.~Clark, I.~Cowhey, O.~Etzioni, T.~Khot, A.~Sabharwal, C.~Schoenick, and O.~Tafjord.
\newblock Think you have solved question answering? try arc, the ai2 reasoning challenge, 2018.
\newblock URL \url{https://arxiv.org/abs/1803.05457}.

\bibitem[Cobbe et~al.(2021)Cobbe, Kosaraju, Bavarian, Chen, Jun, Kaiser, Plappert, Tworek, Hilton, Nakano, Hesse, and Schulman]{cobbe2021trainingverifierssolvemath}
K.~Cobbe, V.~Kosaraju, M.~Bavarian, M.~Chen, H.~Jun, L.~Kaiser, M.~Plappert, J.~Tworek, J.~Hilton, R.~Nakano, C.~Hesse, and J.~Schulman.
\newblock Training verifiers to solve math word problems, 2021.
\newblock URL \url{https://arxiv.org/abs/2110.14168}.

\bibitem[Cortes et~al.(1993)Cortes, Jackel, Solla, Vapnik, and Denker]{NIPS1993_1aa48fc4}
C.~Cortes, L.~D. Jackel, S.~Solla, V.~Vapnik, and J.~Denker.
\newblock Learning curves: Asymptotic values and rate of convergence.
\newblock In J.~Cowan, G.~Tesauro, and J.~Alspector, editors, \emph{Advances in Neural Information Processing Systems}, volume~6. Morgan-Kaufmann, 1993.
\newblock URL \url{https://proceedings.neurips.cc/paper_files/paper/1993/file/1aa48fc4880bb0c9b8a3bf979d3b917e-Paper.pdf}.

\bibitem[D'Angelo et~al.(2024)D'Angelo, Andriushchenko, Varre, and Flammarion]{dangelo2024needweightdecaymodern}
F.~D'Angelo, M.~Andriushchenko, A.~Varre, and N.~Flammarion.
\newblock Why do we need weight decay in modern deep learning?, 2024.
\newblock URL \url{https://arxiv.org/abs/2310.04415}.

\bibitem[DatologyAI et~al.(2025)DatologyAI, :, Maini, Dorna, Doshi, Carranza, Pan, Urbanek, Burstein, Fang, Deng, Abbas, Larsen, Blakeney, Bannur, Baek, Teh, Schwab, Mongstad, Yin, Wills, Mentzer, Merrick, Monti, Adiga, Joshi, Das, Wang, Gaza, Morcos, and Leavitt]{datologyai2025beyondweblessonsscalingsynthetic}
DatologyAI, :, P.~Maini, V.~Dorna, P.~Doshi, A.~Carranza, F.~Pan, J.~Urbanek, P.~Burstein, A.~Fang, A.~Deng, A.~Abbas, B.~Larsen, C.~Blakeney, C.~Bannur, C.~Baek, D.~Teh, D.~Schwab, H.~Mongstad, H.~Yin, J.~Wills, K.~Mentzer, L.~Merrick, R.~Monti, R.~Adiga, S.~Joshi, S.~Das, Z.~Wang, B.~Gaza, A.~Morcos, and M.~Leavitt.
\newblock Beyondweb: Lessons from scaling synthetic data for trillion-scale pretraining, 2025.
\newblock URL \url{https://arxiv.org/abs/2508.10975}.

\bibitem[Deng et~al.(2009)Deng, Dong, Socher, Li, Li, and Fei-Fei]{5206848}
J.~Deng, W.~Dong, R.~Socher, L.-J. Li, K.~Li, and L.~Fei-Fei.
\newblock Imagenet: A large-scale hierarchical image database.
\newblock In \emph{2009 IEEE Conference on Computer Vision and Pattern Recognition}, pages 248--255, 2009.
\newblock \doi{10.1109/CVPR.2009.5206848}.

\bibitem[Dietterich(2000)]{10.5555/648054.743935}
T.~G. Dietterich.
\newblock Ensemble methods in machine learning.
\newblock In \emph{Proceedings of the First International Workshop on Multiple Classifier Systems}, MCS '00, page 1–15, Berlin, Heidelberg, 2000. Springer-Verlag.
\newblock ISBN 3540677046.

\bibitem[Dohmatob et~al.(2024)Dohmatob, Feng, Subramonian, and Kempe]{dohmatob2024strongmodelcollapse}
E.~Dohmatob, Y.~Feng, A.~Subramonian, and J.~Kempe.
\newblock Strong model collapse, 2024.
\newblock URL \url{https://arxiv.org/abs/2410.04840}.

\bibitem[Everett et~al.(2024)Everett, Xiao, Wortsman, Alemi, Novak, Liu, Gur, Sohl-Dickstein, Kaelbling, Lee, and Pennington]{everett2024scalingexponentsparameterizationsoptimizers}
K.~Everett, L.~Xiao, M.~Wortsman, A.~A. Alemi, R.~Novak, P.~J. Liu, I.~Gur, J.~Sohl-Dickstein, L.~P. Kaelbling, J.~Lee, and J.~Pennington.
\newblock Scaling exponents across parameterizations and optimizers, 2024.
\newblock URL \url{https://arxiv.org/abs/2407.05872}.

\bibitem[Gadre et~al.(2024)Gadre, Smyrnis, Shankar, Gururangan, Wortsman, Shao, Mercat, Fang, Li, Keh, Xin, Nezhurina, Vasiljevic, Jitsev, Soldaini, Dimakis, Ilharco, Koh, Song, Kollar, Carmon, Dave, Heckel, Muennighoff, and Schmidt]{gadre2024languagemodelsscalereliably}
S.~Y. Gadre, G.~Smyrnis, V.~Shankar, S.~Gururangan, M.~Wortsman, R.~Shao, J.~Mercat, A.~Fang, J.~Li, S.~Keh, R.~Xin, M.~Nezhurina, I.~Vasiljevic, J.~Jitsev, L.~Soldaini, A.~G. Dimakis, G.~Ilharco, P.~W. Koh, S.~Song, T.~Kollar, Y.~Carmon, A.~Dave, R.~Heckel, N.~Muennighoff, and L.~Schmidt.
\newblock Language models scale reliably with over-training and on downstream tasks, 2024.
\newblock URL \url{https://arxiv.org/abs/2403.08540}.

\bibitem[Gal and Ghahramani(2016)]{gal2016theoreticallygroundedapplicationdropout}
Y.~Gal and Z.~Ghahramani.
\newblock A theoretically grounded application of dropout in recurrent neural networks, 2016.
\newblock URL \url{https://arxiv.org/abs/1512.05287}.

\bibitem[Gao et~al.(2024)Gao, Tow, Abbasi, Biderman, Black, DiPofi, Foster, Golding, Hsu, Le~Noac'h, Li, McDonell, Muennighoff, Ociepa, Phang, Reynolds, Schoelkopf, Skowron, Sutawika, Tang, Thite, Wang, Wang, and Zou]{eval-harness}
L.~Gao, J.~Tow, B.~Abbasi, S.~Biderman, S.~Black, A.~DiPofi, C.~Foster, L.~Golding, J.~Hsu, A.~Le~Noac'h, H.~Li, K.~McDonell, N.~Muennighoff, C.~Ociepa, J.~Phang, L.~Reynolds, H.~Schoelkopf, A.~Skowron, L.~Sutawika, E.~Tang, A.~Thite, B.~Wang, K.~Wang, and A.~Zou.
\newblock The language model evaluation harness, 07 2024.
\newblock URL \url{https://zenodo.org/records/12608602}.

\bibitem[Garipov et~al.(2018)Garipov, Izmailov, Podoprikhin, Vetrov, and Wilson]{garipov2018losssurfacesmodeconnectivity}
T.~Garipov, P.~Izmailov, D.~Podoprikhin, D.~Vetrov, and A.~G. Wilson.
\newblock Loss surfaces, mode connectivity, and fast ensembling of dnns, 2018.
\newblock URL \url{https://arxiv.org/abs/1802.10026}.

\bibitem[Gerstgrasser et~al.(2024)Gerstgrasser, Schaeffer, Dey, Rafailov, Sleight, Hughes, Korbak, Agrawal, Pai, Gromov, Roberts, Yang, Donoho, and Koyejo]{gerstgrasser2024modelcollapseinevitablebreaking}
M.~Gerstgrasser, R.~Schaeffer, A.~Dey, R.~Rafailov, H.~Sleight, J.~Hughes, T.~Korbak, R.~Agrawal, D.~Pai, A.~Gromov, D.~A. Roberts, D.~Yang, D.~L. Donoho, and S.~Koyejo.
\newblock Is model collapse inevitable? breaking the curse of recursion by accumulating real and synthetic data, 2024.
\newblock URL \url{https://arxiv.org/abs/2404.01413}.

\bibitem[Gladstone et~al.(2025)Gladstone, Nanduru, Islam, Han, Ha, Chadha, Du, Ji, Li, and Iqbal]{gladstone2025energybasedtransformersscalablelearners}
A.~Gladstone, G.~Nanduru, M.~M. Islam, P.~Han, H.~Ha, A.~Chadha, Y.~Du, H.~Ji, J.~Li, and T.~Iqbal.
\newblock Energy-based transformers are scalable learners and thinkers, 2025.
\newblock URL \url{https://arxiv.org/abs/2507.02092}.

\bibitem[Goyal et~al.(2024)Goyal, Maini, Lipton, Raghunathan, and Kolter]{goyal2024scalinglawsdatafiltering}
S.~Goyal, P.~Maini, Z.~C. Lipton, A.~Raghunathan, and J.~Z. Kolter.
\newblock Scaling laws for data filtering -- data curation cannot be compute agnostic, 2024.
\newblock URL \url{https://arxiv.org/abs/2404.07177}.

\bibitem[Goyal et~al.(2025)Goyal, Lopez-Paz, and Ahuja]{goyal2025distilledpretrainingmodernlens}
S.~Goyal, D.~Lopez-Paz, and K.~Ahuja.
\newblock Distilled pretraining: A modern lens of data, in-context learning and test-time scaling, 2025.
\newblock URL \url{https://arxiv.org/abs/2509.01649}.

\bibitem[Grattafiori et~al.(2024)Grattafiori, Dubey, Jauhri, Pandey, Kadian, Al-Dahle, Letman, Mathur, Schelten, Vaughan, Yang, Fan, Goyal, Hartshorn, Yang, Mitra, Sravankumar, Korenev, Hinsvark, Rao, Zhang, Rodriguez, Gregerson, Spataru, Roziere, Biron, Tang, Chern, Caucheteux, Nayak, Bi, Marra, McConnell, Keller, Touret, Wu, Wong, Ferrer, Nikolaidis, Allonsius, Song, Pintz, Livshits, Wyatt, Esiobu, Choudhary, Mahajan, Garcia-Olano, Perino, Hupkes, Lakomkin, AlBadawy, Lobanova, Dinan, Smith, Radenovic, Guzmán, Zhang, Synnaeve, Lee, Anderson, Thattai, Nail, Mialon, Pang, Cucurell, Nguyen, Korevaar, Xu, Touvron, Zarov, Ibarra, Kloumann, Misra, Evtimov, Zhang, Copet, Lee, Geffert, Vranes, Park, Mahadeokar, Shah, van~der Linde, Billock, Hong, Lee, Fu, Chi, Huang, Liu, Wang, Yu, Bitton, Spisak, Park, Rocca, Johnstun, Saxe, Jia, Alwala, Prasad, Upasani, Plawiak, Li, Heafield, Stone, El-Arini, Iyer, Malik, Chiu, Bhalla, Lakhotia, Rantala-Yeary, van~der Maaten, Chen, Tan, Jenkins, Martin, Madaan, Malo, Blecher,
  Landzaat, de~Oliveira, Muzzi, Pasupuleti, Singh, Paluri, Kardas, Tsimpoukelli, Oldham, Rita, Pavlova, Kambadur, Lewis, Si, Singh, Hassan, Goyal, Torabi, Bashlykov, Bogoychev, Chatterji, Zhang, Duchenne, Çelebi, Alrassy, Zhang, Li, Vasic, Weng, Bhargava, Dubal, Krishnan, Koura, Xu, He, Dong, Srinivasan, Ganapathy, Calderer, Cabral, Stojnic, Raileanu, Maheswari, Girdhar, Patel, Sauvestre, Polidoro, Sumbaly, Taylor, Silva, Hou, Wang, Hosseini, Chennabasappa, Singh, Bell, Kim, Edunov, Nie, Narang, Raparthy, Shen, Wan, Bhosale, Zhang, Vandenhende, Batra, Whitman, Sootla, Collot, Gururangan, Borodinsky, Herman, Fowler, Sheasha, Georgiou, Scialom, Speckbacher, Mihaylov, Xiao, Karn, Goswami, Gupta, Ramanathan, Kerkez, Gonguet, Do, Vogeti, Albiero, Petrovic, Chu, Xiong, Fu, Meers, Martinet, Wang, Wang, Tan, Xia, Xie, Jia, Wang, Goldschlag, Gaur, Babaei, Wen, Song, Zhang, Li, Mao, Coudert, Yan, Chen, Papakipos, Singh, Srivastava, Jain, Kelsey, Shajnfeld, Gangidi, Victoria, Goldstand, Menon, Sharma, Boesenberg,
  Baevski, Feinstein, Kallet, Sangani, Teo, Yunus, Lupu, Alvarado, Caples, Gu, Ho, Poulton, Ryan, Ramchandani, Dong, Franco, Goyal, Saraf, Chowdhury, Gabriel, Bharambe, Eisenman, Yazdan, James, Maurer, Leonhardi, Huang, Loyd, Paola, Paranjape, Liu, Wu, Ni, Hancock, Wasti, Spence, Stojkovic, Gamido, Montalvo, Parker, Burton, Mejia, Liu, Wang, Kim, Zhou, Hu, Chu, Cai, Tindal, Feichtenhofer, Gao, Civin, Beaty, Kreymer, Li, Adkins, Xu, Testuggine, David, Parikh, Liskovich, Foss, Wang, Le, Holland, Dowling, Jamil, Montgomery, Presani, Hahn, Wood, Le, Brinkman, Arcaute, Dunbar, Smothers, Sun, Kreuk, Tian, Kokkinos, Ozgenel, Caggioni, Kanayet, Seide, Florez, Schwarz, Badeer, Swee, Halpern, Herman, Sizov, Guangyi, Zhang, Lakshminarayanan, Inan, Shojanazeri, Zou, Wang, Zha, Habeeb, Rudolph, Suk, Aspegren, Goldman, Zhan, Damlaj, Molybog, Tufanov, Leontiadis, Veliche, Gat, Weissman, Geboski, Kohli, Lam, Asher, Gaya, Marcus, Tang, Chan, Zhen, Reizenstein, Teboul, Zhong, Jin, Yang, Cummings, Carvill, Shepard, McPhie,
  Torres, Ginsburg, Wang, Wu, U, Saxena, Khandelwal, Zand, Matosich, Veeraraghavan, Michelena, Li, Jagadeesh, Huang, Chawla, Huang, Chen, Garg, A, Silva, Bell, Zhang, Guo, Yu, Moshkovich, Wehrstedt, Khabsa, Avalani, Bhatt, Mankus, Hasson, Lennie, Reso, Groshev, Naumov, Lathi, Keneally, Liu, Seltzer, Valko, Restrepo, Patel, Vyatskov, Samvelyan, Clark, Macey, Wang, Hermoso, Metanat, Rastegari, Bansal, Santhanam, Parks, White, Bawa, Singhal, Egebo, Usunier, Mehta, Laptev, Dong, Cheng, Chernoguz, Hart, Salpekar, Kalinli, Kent, Parekh, Saab, Balaji, Rittner, Bontrager, Roux, Dollar, Zvyagina, Ratanchandani, Yuvraj, Liang, Alao, Rodriguez, Ayub, Murthy, Nayani, Mitra, Parthasarathy, Li, Hogan, Battey, Wang, Howes, Rinott, Mehta, Siby, Bondu, Datta, Chugh, Hunt, Dhillon, Sidorov, Pan, Mahajan, Verma, Yamamoto, Ramaswamy, Lindsay, Lindsay, Feng, Lin, Zha, Patil, Shankar, Zhang, Zhang, Wang, Agarwal, Sajuyigbe, Chintala, Max, Chen, Kehoe, Satterfield, Govindaprasad, Gupta, Deng, Cho, Virk, Subramanian, Choudhury,
  Goldman, Remez, Glaser, Best, Koehler, Robinson, Li, Zhang, Matthews, Chou, Shaked, Vontimitta, Ajayi, Montanez, Mohan, Kumar, Mangla, Ionescu, Poenaru, Mihailescu, Ivanov, Li, Wang, Jiang, Bouaziz, Constable, Tang, Wu, Wang, Wu, Gao, Kleinman, Chen, Hu, Jia, Qi, Li, Zhang, Zhang, Adi, Nam, Yu, Wang, Zhao, Hao, Qian, Li, He, Rait, DeVito, Rosnbrick, Wen, Yang, Zhao, and Ma]{grattafiori2024llama3herdmodels}
A.~Grattafiori, A.~Dubey, A.~Jauhri, A.~Pandey, A.~Kadian, A.~Al-Dahle, A.~Letman, A.~Mathur, A.~Schelten, A.~Vaughan, A.~Yang, A.~Fan, A.~Goyal, A.~Hartshorn, A.~Yang, A.~Mitra, A.~Sravankumar, A.~Korenev, A.~Hinsvark, A.~Rao, A.~Zhang, A.~Rodriguez, A.~Gregerson, A.~Spataru, B.~Roziere, B.~Biron, B.~Tang, B.~Chern, C.~Caucheteux, C.~Nayak, C.~Bi, C.~Marra, C.~McConnell, C.~Keller, C.~Touret, C.~Wu, C.~Wong, C.~C. Ferrer, C.~Nikolaidis, D.~Allonsius, D.~Song, D.~Pintz, D.~Livshits, D.~Wyatt, D.~Esiobu, D.~Choudhary, D.~Mahajan, D.~Garcia-Olano, D.~Perino, D.~Hupkes, E.~Lakomkin, E.~AlBadawy, E.~Lobanova, E.~Dinan, E.~M. Smith, F.~Radenovic, F.~Guzmán, F.~Zhang, G.~Synnaeve, G.~Lee, G.~L. Anderson, G.~Thattai, G.~Nail, G.~Mialon, G.~Pang, G.~Cucurell, H.~Nguyen, H.~Korevaar, H.~Xu, H.~Touvron, I.~Zarov, I.~A. Ibarra, I.~Kloumann, I.~Misra, I.~Evtimov, J.~Zhang, J.~Copet, J.~Lee, J.~Geffert, J.~Vranes, J.~Park, J.~Mahadeokar, J.~Shah, J.~van~der Linde, J.~Billock, J.~Hong, J.~Lee, J.~Fu, J.~Chi, J.~Huang,
  J.~Liu, J.~Wang, J.~Yu, J.~Bitton, J.~Spisak, J.~Park, J.~Rocca, J.~Johnstun, J.~Saxe, J.~Jia, K.~V. Alwala, K.~Prasad, K.~Upasani, K.~Plawiak, K.~Li, K.~Heafield, K.~Stone, K.~El-Arini, K.~Iyer, K.~Malik, K.~Chiu, K.~Bhalla, K.~Lakhotia, L.~Rantala-Yeary, L.~van~der Maaten, L.~Chen, L.~Tan, L.~Jenkins, L.~Martin, L.~Madaan, L.~Malo, L.~Blecher, L.~Landzaat, L.~de~Oliveira, M.~Muzzi, M.~Pasupuleti, M.~Singh, M.~Paluri, M.~Kardas, M.~Tsimpoukelli, M.~Oldham, M.~Rita, M.~Pavlova, M.~Kambadur, M.~Lewis, M.~Si, M.~K. Singh, M.~Hassan, N.~Goyal, N.~Torabi, N.~Bashlykov, N.~Bogoychev, N.~Chatterji, N.~Zhang, O.~Duchenne, O.~Çelebi, P.~Alrassy, P.~Zhang, P.~Li, P.~Vasic, P.~Weng, P.~Bhargava, P.~Dubal, P.~Krishnan, P.~S. Koura, P.~Xu, Q.~He, Q.~Dong, R.~Srinivasan, R.~Ganapathy, R.~Calderer, R.~S. Cabral, R.~Stojnic, R.~Raileanu, R.~Maheswari, R.~Girdhar, R.~Patel, R.~Sauvestre, R.~Polidoro, R.~Sumbaly, R.~Taylor, R.~Silva, R.~Hou, R.~Wang, S.~Hosseini, S.~Chennabasappa, S.~Singh, S.~Bell, S.~S. Kim, S.~Edunov,
  S.~Nie, S.~Narang, S.~Raparthy, S.~Shen, S.~Wan, S.~Bhosale, S.~Zhang, S.~Vandenhende, S.~Batra, S.~Whitman, S.~Sootla, S.~Collot, S.~Gururangan, S.~Borodinsky, T.~Herman, T.~Fowler, T.~Sheasha, T.~Georgiou, T.~Scialom, T.~Speckbacher, T.~Mihaylov, T.~Xiao, U.~Karn, V.~Goswami, V.~Gupta, V.~Ramanathan, V.~Kerkez, V.~Gonguet, V.~Do, V.~Vogeti, V.~Albiero, V.~Petrovic, W.~Chu, W.~Xiong, W.~Fu, W.~Meers, X.~Martinet, X.~Wang, X.~Wang, X.~E. Tan, X.~Xia, X.~Xie, X.~Jia, X.~Wang, Y.~Goldschlag, Y.~Gaur, Y.~Babaei, Y.~Wen, Y.~Song, Y.~Zhang, Y.~Li, Y.~Mao, Z.~D. Coudert, Z.~Yan, Z.~Chen, Z.~Papakipos, A.~Singh, A.~Srivastava, A.~Jain, A.~Kelsey, A.~Shajnfeld, A.~Gangidi, A.~Victoria, A.~Goldstand, A.~Menon, A.~Sharma, A.~Boesenberg, A.~Baevski, A.~Feinstein, A.~Kallet, A.~Sangani, A.~Teo, A.~Yunus, A.~Lupu, A.~Alvarado, A.~Caples, A.~Gu, A.~Ho, A.~Poulton, A.~Ryan, A.~Ramchandani, A.~Dong, A.~Franco, A.~Goyal, A.~Saraf, A.~Chowdhury, A.~Gabriel, A.~Bharambe, A.~Eisenman, A.~Yazdan, B.~James, B.~Maurer,
  B.~Leonhardi, B.~Huang, B.~Loyd, B.~D. Paola, B.~Paranjape, B.~Liu, B.~Wu, B.~Ni, B.~Hancock, B.~Wasti, B.~Spence, B.~Stojkovic, B.~Gamido, B.~Montalvo, C.~Parker, C.~Burton, C.~Mejia, C.~Liu, C.~Wang, C.~Kim, C.~Zhou, C.~Hu, C.-H. Chu, C.~Cai, C.~Tindal, C.~Feichtenhofer, C.~Gao, D.~Civin, D.~Beaty, D.~Kreymer, D.~Li, D.~Adkins, D.~Xu, D.~Testuggine, D.~David, D.~Parikh, D.~Liskovich, D.~Foss, D.~Wang, D.~Le, D.~Holland, E.~Dowling, E.~Jamil, E.~Montgomery, E.~Presani, E.~Hahn, E.~Wood, E.-T. Le, E.~Brinkman, E.~Arcaute, E.~Dunbar, E.~Smothers, F.~Sun, F.~Kreuk, F.~Tian, F.~Kokkinos, F.~Ozgenel, F.~Caggioni, F.~Kanayet, F.~Seide, G.~M. Florez, G.~Schwarz, G.~Badeer, G.~Swee, G.~Halpern, G.~Herman, G.~Sizov, Guangyi, Zhang, G.~Lakshminarayanan, H.~Inan, H.~Shojanazeri, H.~Zou, H.~Wang, H.~Zha, H.~Habeeb, H.~Rudolph, H.~Suk, H.~Aspegren, H.~Goldman, H.~Zhan, I.~Damlaj, I.~Molybog, I.~Tufanov, I.~Leontiadis, I.-E. Veliche, I.~Gat, J.~Weissman, J.~Geboski, J.~Kohli, J.~Lam, J.~Asher, J.-B. Gaya, J.~Marcus,
  J.~Tang, J.~Chan, J.~Zhen, J.~Reizenstein, J.~Teboul, J.~Zhong, J.~Jin, J.~Yang, J.~Cummings, J.~Carvill, J.~Shepard, J.~McPhie, J.~Torres, J.~Ginsburg, J.~Wang, K.~Wu, K.~H. U, K.~Saxena, K.~Khandelwal, K.~Zand, K.~Matosich, K.~Veeraraghavan, K.~Michelena, K.~Li, K.~Jagadeesh, K.~Huang, K.~Chawla, K.~Huang, L.~Chen, L.~Garg, L.~A, L.~Silva, L.~Bell, L.~Zhang, L.~Guo, L.~Yu, L.~Moshkovich, L.~Wehrstedt, M.~Khabsa, M.~Avalani, M.~Bhatt, M.~Mankus, M.~Hasson, M.~Lennie, M.~Reso, M.~Groshev, M.~Naumov, M.~Lathi, M.~Keneally, M.~Liu, M.~L. Seltzer, M.~Valko, M.~Restrepo, M.~Patel, M.~Vyatskov, M.~Samvelyan, M.~Clark, M.~Macey, M.~Wang, M.~J. Hermoso, M.~Metanat, M.~Rastegari, M.~Bansal, N.~Santhanam, N.~Parks, N.~White, N.~Bawa, N.~Singhal, N.~Egebo, N.~Usunier, N.~Mehta, N.~P. Laptev, N.~Dong, N.~Cheng, O.~Chernoguz, O.~Hart, O.~Salpekar, O.~Kalinli, P.~Kent, P.~Parekh, P.~Saab, P.~Balaji, P.~Rittner, P.~Bontrager, P.~Roux, P.~Dollar, P.~Zvyagina, P.~Ratanchandani, P.~Yuvraj, Q.~Liang, R.~Alao, R.~Rodriguez,
  R.~Ayub, R.~Murthy, R.~Nayani, R.~Mitra, R.~Parthasarathy, R.~Li, R.~Hogan, R.~Battey, R.~Wang, R.~Howes, R.~Rinott, S.~Mehta, S.~Siby, S.~J. Bondu, S.~Datta, S.~Chugh, S.~Hunt, S.~Dhillon, S.~Sidorov, S.~Pan, S.~Mahajan, S.~Verma, S.~Yamamoto, S.~Ramaswamy, S.~Lindsay, S.~Lindsay, S.~Feng, S.~Lin, S.~C. Zha, S.~Patil, S.~Shankar, S.~Zhang, S.~Zhang, S.~Wang, S.~Agarwal, S.~Sajuyigbe, S.~Chintala, S.~Max, S.~Chen, S.~Kehoe, S.~Satterfield, S.~Govindaprasad, S.~Gupta, S.~Deng, S.~Cho, S.~Virk, S.~Subramanian, S.~Choudhury, S.~Goldman, T.~Remez, T.~Glaser, T.~Best, T.~Koehler, T.~Robinson, T.~Li, T.~Zhang, T.~Matthews, T.~Chou, T.~Shaked, V.~Vontimitta, V.~Ajayi, V.~Montanez, V.~Mohan, V.~S. Kumar, V.~Mangla, V.~Ionescu, V.~Poenaru, V.~T. Mihailescu, V.~Ivanov, W.~Li, W.~Wang, W.~Jiang, W.~Bouaziz, W.~Constable, X.~Tang, X.~Wu, X.~Wang, X.~Wu, X.~Gao, Y.~Kleinman, Y.~Chen, Y.~Hu, Y.~Jia, Y.~Qi, Y.~Li, Y.~Zhang, Y.~Zhang, Y.~Adi, Y.~Nam, Yu, Wang, Y.~Zhao, Y.~Hao, Y.~Qian, Y.~Li, Y.~He, Z.~Rait, Z.~DeVito,
  Z.~Rosnbrick, Z.~Wen, Z.~Yang, Z.~Zhao, and Z.~Ma.
\newblock The llama 3 herd of models, 2024.
\newblock URL \url{https://arxiv.org/abs/2407.21783}.

\bibitem[Grave et~al.(2016)Grave, Joulin, and Usunier]{grave2016improvingneurallanguagemodels}
E.~Grave, A.~Joulin, and N.~Usunier.
\newblock Improving neural language models with a continuous cache, 2016.
\newblock URL \url{https://arxiv.org/abs/1612.04426}.

\bibitem[Gu et~al.(2024)Gu, Dong, Wei, and Huang]{gu2024minillmknowledgedistillationlarge}
Y.~Gu, L.~Dong, F.~Wei, and M.~Huang.
\newblock Minillm: Knowledge distillation of large language models, 2024.
\newblock URL \url{https://arxiv.org/abs/2306.08543}.

\bibitem[Hastie et~al.(2020)Hastie, Montanari, Rosset, and Tibshirani]{hastie2020surpriseshighdimensionalridgelesssquares}
T.~Hastie, A.~Montanari, S.~Rosset, and R.~J. Tibshirani.
\newblock Surprises in high-dimensional ridgeless least squares interpolation, 2020.
\newblock URL \url{https://arxiv.org/abs/1903.08560}.

\bibitem[Hendrycks et~al.(2021)Hendrycks, Burns, Kadavath, Arora, Basart, Tang, Song, and Steinhardt]{hendrycks2021measuringmathematicalproblemsolving}
D.~Hendrycks, C.~Burns, S.~Kadavath, A.~Arora, S.~Basart, E.~Tang, D.~Song, and J.~Steinhardt.
\newblock Measuring mathematical problem solving with the math dataset, 2021.
\newblock URL \url{https://arxiv.org/abs/2103.03874}.

\bibitem[Henighan et~al.(2020)Henighan, Kaplan, Katz, Chen, Hesse, Jackson, Jun, Brown, Dhariwal, Gray, Hallacy, Mann, Radford, Ramesh, Ryder, Ziegler, Schulman, Amodei, and McCandlish]{henighan2020scalinglawsautoregressivegenerative}
T.~Henighan, J.~Kaplan, M.~Katz, M.~Chen, C.~Hesse, J.~Jackson, H.~Jun, T.~B. Brown, P.~Dhariwal, S.~Gray, C.~Hallacy, B.~Mann, A.~Radford, A.~Ramesh, N.~Ryder, D.~M. Ziegler, J.~Schulman, D.~Amodei, and S.~McCandlish.
\newblock Scaling laws for autoregressive generative modeling, 2020.
\newblock URL \url{https://arxiv.org/abs/2010.14701}.

\bibitem[Hestness et~al.(2017)Hestness, Narang, Ardalani, Diamos, Jun, Kianinejad, Patwary, Yang, and Zhou]{hestness2017deeplearningscalingpredictable}
J.~Hestness, S.~Narang, N.~Ardalani, G.~Diamos, H.~Jun, H.~Kianinejad, M.~M.~A. Patwary, Y.~Yang, and Y.~Zhou.
\newblock Deep learning scaling is predictable, empirically, 2017.
\newblock URL \url{https://arxiv.org/abs/1712.00409}.

\bibitem[Hestness et~al.(2019)Hestness, Ardalani, and Diamos]{hestness2019humanlevelaccuracycomputationalchallenges}
J.~Hestness, N.~Ardalani, and G.~Diamos.
\newblock Beyond human-level accuracy: Computational challenges in deep learning, 2019.
\newblock URL \url{https://arxiv.org/abs/1909.01736}.

\bibitem[Hinton et~al.(2015)Hinton, Vinyals, and Dean]{hinton2015distillingknowledgeneuralnetwork}
G.~Hinton, O.~Vinyals, and J.~Dean.
\newblock Distilling the knowledge in a neural network, 2015.
\newblock URL \url{https://arxiv.org/abs/1503.02531}.

\bibitem[Hoffmann et~al.(2022)Hoffmann, Borgeaud, Mensch, Buchatskaya, Cai, Rutherford, de~Las~Casas, Hendricks, Welbl, Clark, Hennigan, Noland, Millican, van~den Driessche, Damoc, Guy, Osindero, Simonyan, Elsen, Rae, Vinyals, and Sifre]{hoffmann2022trainingcomputeoptimallargelanguage}
J.~Hoffmann, S.~Borgeaud, A.~Mensch, E.~Buchatskaya, T.~Cai, E.~Rutherford, D.~de~Las~Casas, L.~A. Hendricks, J.~Welbl, A.~Clark, T.~Hennigan, E.~Noland, K.~Millican, G.~van~den Driessche, B.~Damoc, A.~Guy, S.~Osindero, K.~Simonyan, E.~Elsen, J.~W. Rae, O.~Vinyals, and L.~Sifre.
\newblock Training compute-optimal large language models, 2022.
\newblock URL \url{https://arxiv.org/abs/2203.15556}.

\bibitem[Huang et~al.(2017)Huang, Li, Pleiss, Liu, Hopcroft, and Weinberger]{huang2017snapshotensemblestrain1}
G.~Huang, Y.~Li, G.~Pleiss, Z.~Liu, J.~E. Hopcroft, and K.~Q. Weinberger.
\newblock Snapshot ensembles: Train 1, get m for free, 2017.
\newblock URL \url{https://arxiv.org/abs/1704.00109}.

\bibitem[Jordan(2024)]{jordan2024varianceneuralnetworktraining}
K.~Jordan.
\newblock On the variance of neural network training with respect to test sets and distributions, 2024.
\newblock URL \url{https://arxiv.org/abs/2304.01910}.

\bibitem[Juravsky et~al.(2025)Juravsky, Chakravarthy, Ehrlich, Eyuboglu, Brown, Shetaye, R{\'e}, and Mirhoseini]{juravsky2025tokasaurus}
J.~Juravsky, A.~Chakravarthy, R.~Ehrlich, S.~Eyuboglu, B.~Brown, J.~Shetaye, C.~R{\'e}, and A.~Mirhoseini.
\newblock Tokasaurus: An llm inference engine for high-throughput workloads.
\newblock \url{https://scalingintelligence.stanford.edu/blogs/tokasaurus/}, 2025.

\bibitem[Kaplan et~al.(2020)Kaplan, McCandlish, Henighan, Brown, Chess, Child, Gray, Radford, Wu, and Amodei]{kaplan2020scalinglawsneurallanguage}
J.~Kaplan, S.~McCandlish, T.~Henighan, T.~B. Brown, B.~Chess, R.~Child, S.~Gray, A.~Radford, J.~Wu, and D.~Amodei.
\newblock Scaling laws for neural language models, 2020.
\newblock URL \url{https://arxiv.org/abs/2001.08361}.

\bibitem[Keskar et~al.(2017)Keskar, Mudigere, Nocedal, Smelyanskiy, and Tang]{keskar2017largebatchtrainingdeeplearning}
N.~S. Keskar, D.~Mudigere, J.~Nocedal, M.~Smelyanskiy, and P.~T.~P. Tang.
\newblock On large-batch training for deep learning: Generalization gap and sharp minima, 2017.
\newblock URL \url{https://arxiv.org/abs/1609.04836}.

\bibitem[Kim and Rush(2016)]{kim2016sequencelevelknowledgedistillation}
Y.~Kim and A.~M. Rush.
\newblock Sequence-level knowledge distillation, 2016.
\newblock URL \url{https://arxiv.org/abs/1606.07947}.

\bibitem[Krause et~al.(2017)Krause, Kahembwe, Murray, and Renals]{krause2017dynamicevaluationneuralsequence}
B.~Krause, E.~Kahembwe, I.~Murray, and S.~Renals.
\newblock Dynamic evaluation of neural sequence models, 2017.
\newblock URL \url{https://arxiv.org/abs/1709.07432}.

\bibitem[Kumar et~al.(2024)Kumar, Ankner, Spector, Bordelon, Muennighoff, Paul, Pehlevan, Ré, and Raghunathan]{kumar2024scalinglawsprecision}
T.~Kumar, Z.~Ankner, B.~F. Spector, B.~Bordelon, N.~Muennighoff, M.~Paul, C.~Pehlevan, C.~Ré, and A.~Raghunathan.
\newblock Scaling laws for precision, 2024.
\newblock URL \url{https://arxiv.org/abs/2411.04330}.

\bibitem[Lakshminarayanan et~al.(2017)Lakshminarayanan, Pritzel, and Blundell]{lakshminarayanan2017simplescalablepredictiveuncertainty}
B.~Lakshminarayanan, A.~Pritzel, and C.~Blundell.
\newblock Simple and scalable predictive uncertainty estimation using deep ensembles, 2017.
\newblock URL \url{https://arxiv.org/abs/1612.01474}.

\bibitem[Lecun et~al.(1998)Lecun, Bottou, Bengio, and Haffner]{726791}
Y.~Lecun, L.~Bottou, Y.~Bengio, and P.~Haffner.
\newblock Gradient-based learning applied to document recognition.
\newblock \emph{Proceedings of the IEEE}, 86\penalty0 (11):\penalty0 2278--2324, 1998.
\newblock \doi{10.1109/5.726791}.

\bibitem[LeCun et~al.(1998)LeCun, Bottou, Orr, and M{\"u}ller]{LeCun1998}
Y.~LeCun, L.~Bottou, G.~B. Orr, and K.~R. M{\"u}ller.
\newblock \emph{Efficient BackProp}, pages 9--50.
\newblock Springer Berlin Heidelberg, Berlin, Heidelberg, 1998.
\newblock ISBN 978-3-540-49430-0.
\newblock \doi{10.1007/3-540-49430-8_2}.
\newblock URL \url{https://doi.org/10.1007/3-540-49430-8_2}.

\bibitem[Li et~al.(2025)Li, Fang, Smyrnis, Ivgi, Jordan, Gadre, Bansal, Guha, Keh, Arora, Garg, Xin, Muennighoff, Heckel, Mercat, Chen, Gururangan, Wortsman, Albalak, Bitton, Nezhurina, Abbas, Hsieh, Ghosh, Gardner, Kilian, Zhang, Shao, Pratt, Sanyal, Ilharco, Daras, Marathe, Gokaslan, Zhang, Chandu, Nguyen, Vasiljevic, Kakade, Song, Sanghavi, Faghri, Oh, Zettlemoyer, Lo, El-Nouby, Pouransari, Toshev, Wang, Groeneveld, Soldaini, Koh, Jitsev, Kollar, Dimakis, Carmon, Dave, Schmidt, and Shankar]{li2025datacomplmsearchgenerationtraining}
J.~Li, A.~Fang, G.~Smyrnis, M.~Ivgi, M.~Jordan, S.~Gadre, H.~Bansal, E.~Guha, S.~Keh, K.~Arora, S.~Garg, R.~Xin, N.~Muennighoff, R.~Heckel, J.~Mercat, M.~Chen, S.~Gururangan, M.~Wortsman, A.~Albalak, Y.~Bitton, M.~Nezhurina, A.~Abbas, C.-Y. Hsieh, D.~Ghosh, J.~Gardner, M.~Kilian, H.~Zhang, R.~Shao, S.~Pratt, S.~Sanyal, G.~Ilharco, G.~Daras, K.~Marathe, A.~Gokaslan, J.~Zhang, K.~Chandu, T.~Nguyen, I.~Vasiljevic, S.~Kakade, S.~Song, S.~Sanghavi, F.~Faghri, S.~Oh, L.~Zettlemoyer, K.~Lo, A.~El-Nouby, H.~Pouransari, A.~Toshev, S.~Wang, D.~Groeneveld, L.~Soldaini, P.~W. Koh, J.~Jitsev, T.~Kollar, A.~G. Dimakis, Y.~Carmon, A.~Dave, L.~Schmidt, and V.~Shankar.
\newblock Datacomp-lm: In search of the next generation of training sets for language models, 2025.
\newblock URL \url{https://arxiv.org/abs/2406.11794}.

\bibitem[Liu et~al.(2025)Liu, Su, Yao, Jiang, Lai, Du, Qin, Xu, Lu, Yan, Chen, Zheng, Liu, Liu, Yin, He, Zhu, Wang, Wang, Dong, Zhang, Kang, Zhang, Xu, Zhang, Wu, Zhou, and Yang]{liu2025muonscalablellmtraining}
J.~Liu, J.~Su, X.~Yao, Z.~Jiang, G.~Lai, Y.~Du, Y.~Qin, W.~Xu, E.~Lu, J.~Yan, Y.~Chen, H.~Zheng, Y.~Liu, S.~Liu, B.~Yin, W.~He, H.~Zhu, Y.~Wang, J.~Wang, M.~Dong, Z.~Zhang, Y.~Kang, H.~Zhang, X.~Xu, Y.~Zhang, Y.~Wu, X.~Zhou, and Z.~Yang.
\newblock Muon is scalable for llm training, 2025.
\newblock URL \url{https://arxiv.org/abs/2502.16982}.

\bibitem[Lobacheva et~al.(2021)Lobacheva, Chirkova, Kodryan, and Vetrov]{lobacheva2021powerlawsdeepensembles}
E.~Lobacheva, N.~Chirkova, M.~Kodryan, and D.~Vetrov.
\newblock On power laws in deep ensembles, 2021.
\newblock URL \url{https://arxiv.org/abs/2007.08483}.

\bibitem[Maini et~al.(2024)Maini, Seto, Bai, Grangier, Zhang, and Jaitly]{maini2024rephrasingwebrecipecompute}
P.~Maini, S.~Seto, H.~Bai, D.~Grangier, Y.~Zhang, and N.~Jaitly.
\newblock Rephrasing the web: A recipe for compute and data-efficient language modeling, 2024.
\newblock URL \url{https://arxiv.org/abs/2401.16380}.

\bibitem[Marcus et~al.(1993)Marcus, Santorini, and Marcinkiewicz]{marcus-etal-1993-building}
M.~P. Marcus, B.~Santorini, and M.~A. Marcinkiewicz.
\newblock Building a large annotated corpus of {E}nglish: The {P}enn {T}reebank.
\newblock \emph{Computational Linguistics}, 19\penalty0 (2):\penalty0 313--330, 1993.
\newblock URL \url{https://aclanthology.org/J93-2004/}.

\bibitem[Marek et~al.(2025)Marek, Lotfi, Somasundaram, Wilson, and Goldblum]{marek2025smallbatchsizetraining}
M.~Marek, S.~Lotfi, A.~Somasundaram, A.~G. Wilson, and M.~Goldblum.
\newblock Small batch size training for language models: When vanilla sgd works, and why gradient accumulation is wasteful, 2025.
\newblock URL \url{https://arxiv.org/abs/2507.07101}.

\bibitem[McCandlish et~al.(2018)McCandlish, Kaplan, Amodei, and Team]{mccandlish2018empiricalmodellargebatchtraining}
S.~McCandlish, J.~Kaplan, D.~Amodei, and O.~D. Team.
\newblock An empirical model of large-batch training, 2018.
\newblock URL \url{https://arxiv.org/abs/1812.06162}.

\bibitem[Merity et~al.(2017)Merity, Keskar, and Socher]{merity2017regularizingoptimizinglstmlanguage}
S.~Merity, N.~S. Keskar, and R.~Socher.
\newblock Regularizing and optimizing lstm language models, 2017.
\newblock URL \url{https://arxiv.org/abs/1708.02182}.

\bibitem[Mikolov et~al.(2010)Mikolov, Karafiát, Burget, Cernocký, and Khudanpur]{inproceedings}
T.~Mikolov, M.~Karafiát, L.~Burget, J.~Cernocký, and S.~Khudanpur.
\newblock Recurrent neural network based language model.
\newblock volume~2, pages 1045--1048, 09 2010.
\newblock \doi{10.21437/Interspeech.2010-343}.

\bibitem[Mobahi et~al.(2020)Mobahi, Farajtabar, and Bartlett]{mobahi2020selfdistillationamplifiesregularizationhilbert}
H.~Mobahi, M.~Farajtabar, and P.~L. Bartlett.
\newblock Self-distillation amplifies regularization in hilbert space, 2020.
\newblock URL \url{https://arxiv.org/abs/2002.05715}.

\bibitem[Muennighoff et~al.(2023)Muennighoff, Rush, Barak, Scao, Piktus, Tazi, Pyysalo, Wolf, and Raffel]{muennighoff2023scalingdataconstrainedlanguagemodels}
N.~Muennighoff, A.~M. Rush, B.~Barak, T.~L. Scao, A.~Piktus, N.~Tazi, S.~Pyysalo, T.~Wolf, and C.~Raffel.
\newblock Scaling data-constrained language models, 2023.
\newblock URL \url{https://arxiv.org/abs/2305.16264}.

\bibitem[Nakkiran et~al.(2019)Nakkiran, Kaplun, Bansal, Yang, Barak, and Sutskever]{nakkiran2019deepdoubledescentbigger}
P.~Nakkiran, G.~Kaplun, Y.~Bansal, T.~Yang, B.~Barak, and I.~Sutskever.
\newblock Deep double descent: Where bigger models and more data hurt, 2019.
\newblock URL \url{https://arxiv.org/abs/1912.02292}.

\bibitem[Nakkiran et~al.(2021)Nakkiran, Venkat, Kakade, and Ma]{nakkiran2021optimalregularizationmitigatedouble}
P.~Nakkiran, P.~Venkat, S.~Kakade, and T.~Ma.
\newblock Optimal regularization can mitigate double descent, 2021.
\newblock URL \url{https://arxiv.org/abs/2003.01897}.

\bibitem[Ni et~al.(2025)Ni, the, and team]{ni2025difflm}
J.~Ni, the, and team.
\newblock Diffusion language models are super data learners.
\newblock \url{https://jinjieni.notion.site/Diffusion-Language-Models-are-Super-Data-Learners-239d8f03a866800ab196e49928c019ac}, 2025.
\newblock Notion Blog.

\bibitem[Prabhudesai et~al.(2025)Prabhudesai, Wu, Zadeh, Fragkiadaki, and Pathak]{prabhudesai2025diffusionbeatsautoregressivedataconstrained}
M.~Prabhudesai, M.~Wu, A.~Zadeh, K.~Fragkiadaki, and D.~Pathak.
\newblock Diffusion beats autoregressive in data-constrained settings, 2025.
\newblock URL \url{https://arxiv.org/abs/2507.15857}.

\bibitem[Rosenfeld et~al.(2019)Rosenfeld, Rosenfeld, Belinkov, and Shavit]{rosenfeld2019constructivepredictiongeneralizationerror}
J.~S. Rosenfeld, A.~Rosenfeld, Y.~Belinkov, and N.~Shavit.
\newblock A constructive prediction of the generalization error across scales, 2019.
\newblock URL \url{https://arxiv.org/abs/1909.12673}.

\bibitem[Ruan et~al.(2024)Ruan, Maddison, and Hashimoto]{ruan2024observationalscalinglawspredictability}
Y.~Ruan, C.~J. Maddison, and T.~Hashimoto.
\newblock Observational scaling laws and the predictability of language model performance, 2024.
\newblock URL \url{https://arxiv.org/abs/2405.10938}.

\bibitem[Ruan et~al.(2025)Ruan, Band, Maddison, and Hashimoto]{ruan2025reasoninglearnlatentthoughts}
Y.~Ruan, N.~Band, C.~J. Maddison, and T.~Hashimoto.
\newblock Reasoning to learn from latent thoughts, 2025.
\newblock URL \url{https://arxiv.org/abs/2503.18866}.

\bibitem[Ruben et~al.(2024)Ruben, Tong, Chaudhry, and Pehlevan]{ruben2024freelunchrandomfeature}
B.~S. Ruben, W.~L. Tong, H.~T. Chaudhry, and C.~Pehlevan.
\newblock No free lunch from random feature ensembles, 2024.
\newblock URL \url{https://arxiv.org/abs/2412.05418}.

\bibitem[Sanh et~al.(2020)Sanh, Debut, Chaumond, and Wolf]{sanh2020distilbertdistilledversionbert}
V.~Sanh, L.~Debut, J.~Chaumond, and T.~Wolf.
\newblock Distilbert, a distilled version of bert: smaller, faster, cheaper and lighter, 2020.
\newblock URL \url{https://arxiv.org/abs/1910.01108}.

\bibitem[Sardana et~al.(2025)Sardana, Portes, Doubov, and Frankle]{sardana2025chinchillaoptimalaccountinginferencelanguage}
N.~Sardana, J.~Portes, S.~Doubov, and J.~Frankle.
\newblock Beyond chinchilla-optimal: Accounting for inference in language model scaling laws, 2025.
\newblock URL \url{https://arxiv.org/abs/2401.00448}.

\bibitem[Sevilla and Roldán(2024)]{epoch2024trainingcomputeoffrontieraimodelsgrowsby45xperyear}
J.~Sevilla and E.~Roldán.
\newblock Training compute of frontier ai models grows by 4-5x per year, 2024.
\newblock URL \url{https://epoch.ai/blog/training-compute-of-frontier-ai-models-grows-by-4-5x-per-year}.
\newblock Accessed: 2025-08-21.

\bibitem[Shannon(1951)]{shannon1951prediction}
C.~Shannon.
\newblock Prediction and entropy of printed english.
\newblock \emph{Bell system technical journal}, 30\penalty0 (1):\penalty0 50--64, 1951.

\bibitem[Shi et~al.(2021)Shi, Livescu, and Gimpel]{shi2021substructuresubstitutionstructureddata}
H.~Shi, K.~Livescu, and K.~Gimpel.
\newblock Substructure substitution: Structured data augmentation for nlp, 2021.
\newblock URL \url{https://arxiv.org/abs/2101.00411}.

\bibitem[Shumailov et~al.(2024)Shumailov, Shumaylov, Zhao, Papernot, Anderson, and Gal]{shumailov2024ai}
I.~Shumailov, Z.~Shumaylov, Y.~Zhao, N.~Papernot, R.~Anderson, and Y.~Gal.
\newblock Ai models collapse when trained on recursively generated data.
\newblock \emph{Nature}, 631\penalty0 (8022):\penalty0 755--759, 2024.

\bibitem[Simon et~al.(2024)Simon, Karkada, Ghosh, and Belkin]{simon2024bettermodernmachinelearning}
J.~B. Simon, D.~Karkada, N.~Ghosh, and M.~Belkin.
\newblock More is better in modern machine learning: when infinite overparameterization is optimal and overfitting is obligatory, 2024.
\newblock URL \url{https://arxiv.org/abs/2311.14646}.

\bibitem[Singh and Jaggi(2023)]{singh2023modelfusionoptimaltransport}
S.~P. Singh and M.~Jaggi.
\newblock Model fusion via optimal transport, 2023.
\newblock URL \url{https://arxiv.org/abs/1910.05653}.

\bibitem[Smith et~al.(2020)Smith, Elsen, and De]{smith2020generalizationbenefitnoisestochastic}
S.~L. Smith, E.~Elsen, and S.~De.
\newblock On the generalization benefit of noise in stochastic gradient descent, 2020.
\newblock URL \url{https://arxiv.org/abs/2006.15081}.

\bibitem[Snell et~al.(2024)Snell, Lee, Xu, and Kumar]{snell2024scalingllmtesttimecompute}
C.~Snell, J.~Lee, K.~Xu, and A.~Kumar.
\newblock Scaling llm test-time compute optimally can be more effective than scaling model parameters, 2024.
\newblock URL \url{https://arxiv.org/abs/2408.03314}.

\bibitem[Sorscher et~al.(2023)Sorscher, Geirhos, Shekhar, Ganguli, and Morcos]{sorscher2023neuralscalinglawsbeating}
B.~Sorscher, R.~Geirhos, S.~Shekhar, S.~Ganguli, and A.~S. Morcos.
\newblock Beyond neural scaling laws: beating power law scaling via data pruning, 2023.
\newblock URL \url{https://arxiv.org/abs/2206.14486}.

\bibitem[Springer et~al.(2025)Springer, Goyal, Wen, Kumar, Yue, Malladi, Neubig, and Raghunathan]{springer2025overtrainedlanguagemodelsharder}
J.~M. Springer, S.~Goyal, K.~Wen, T.~Kumar, X.~Yue, S.~Malladi, G.~Neubig, and A.~Raghunathan.
\newblock Overtrained language models are harder to fine-tune, 2025.
\newblock URL \url{https://arxiv.org/abs/2503.19206}.

\bibitem[Su et~al.(2025)Su, Kong, Lin, Jennings, Norick, Kliegl, Patwary, Shoeybi, and Catanzaro]{su2025nemotroncctransformingcommoncrawl}
D.~Su, K.~Kong, Y.~Lin, J.~Jennings, B.~Norick, M.~Kliegl, M.~Patwary, M.~Shoeybi, and B.~Catanzaro.
\newblock Nemotron-cc: Transforming common crawl into a refined long-horizon pretraining dataset, 2025.
\newblock URL \url{https://arxiv.org/abs/2412.02595}.

\bibitem[Summers and Dinneen(2021)]{summers2021nondeterminisminstabilityneuralnetwork}
C.~Summers and M.~J. Dinneen.
\newblock Nondeterminism and instability in neural network optimization, 2021.
\newblock URL \url{https://arxiv.org/abs/2103.04514}.

\bibitem[Sutton(2019)]{sutton2019bitterlesson}
R.~Sutton.
\newblock The bitter lesson.
\newblock \url{http://www.incompleteideas.net/IncIdeas/BitterLesson.html}, 2019.
\newblock Blog post.

\bibitem[Takase et~al.(2018)Takase, Suzuki, and Nagata]{takase2018directoutputconnectionhighrank}
S.~Takase, J.~Suzuki, and M.~Nagata.
\newblock Direct output connection for a high-rank language model, 2018.
\newblock URL \url{https://arxiv.org/abs/1808.10143}.

\bibitem[Taori and Hashimoto(2022)]{taori2022datafeedbackloopsmodeldriven}
R.~Taori and T.~B. Hashimoto.
\newblock Data feedback loops: Model-driven amplification of dataset biases, 2022.
\newblock URL \url{https://arxiv.org/abs/2209.03942}.

\bibitem[Team et~al.(2025{\natexlab{a}})Team, Kamath, Ferret, Pathak, Vieillard, Merhej, Perrin, Matejovicova, Ramé, Rivière, Rouillard, Mesnard, Cideron, bastien Grill, Ramos, Yvinec, Casbon, Pot, Penchev, Liu, Visin, Kenealy, Beyer, Zhai, Tsitsulin, Busa-Fekete, Feng, Sachdeva, Coleman, Gao, Mustafa, Barr, Parisotto, Tian, Eyal, Cherry, Peter, Sinopalnikov, Bhupatiraju, Agarwal, Kazemi, Malkin, Kumar, Vilar, Brusilovsky, Luo, Steiner, Friesen, Sharma, Sharma, Gilady, Goedeckemeyer, Saade, Feng, Kolesnikov, Bendebury, Abdagic, Vadi, György, Pinto, Das, Bapna, Miech, Yang, Paterson, Shenoy, Chakrabarti, Piot, Wu, Shahriari, Petrini, Chen, Lan, Choquette-Choo, Carey, Brick, Deutsch, Eisenbud, Cattle, Cheng, Paparas, Sreepathihalli, Reid, Tran, Zelle, Noland, Huizenga, Kharitonov, Liu, Amirkhanyan, Cameron, Hashemi, Klimczak-Plucińska, Singh, Mehta, Lehri, Hazimeh, Ballantyne, Szpektor, Nardini, Pouget-Abadie, Chan, Stanton, Wieting, Lai, Orbay, Fernandez, Newlan, yeong Ji, Singh, Black, Yu, Hui,
  Vodrahalli, Greff, Qiu, Valentine, Coelho, Ritter, Hoffman, Watson, Chaturvedi, Moynihan, Ma, Babar, Noy, Byrd, Roy, Momchev, Chauhan, Sachdeva, Bunyan, Botarda, Caron, Rubenstein, Culliton, Schmid, Sessa, Xu, Stanczyk, Tafti, Shivanna, Wu, Pan, Rokni, Willoughby, Vallu, Mullins, Jerome, Smoot, Girgin, Iqbal, Reddy, Sheth, Põder, Bhatnagar, Panyam, Eiger, Zhang, Liu, Yacovone, Liechty, Kalra, Evci, Misra, Roseberry, Feinberg, Kolesnikov, Han, Kwon, Chen, Chow, Zhu, Wei, Egyed, Cotruta, Giang, Kirk, Rao, Black, Babar, Lo, Moreira, Martins, Sanseviero, Gonzalez, Gleicher, Warkentin, Mirrokni, Senter, Collins, Barral, Ghahramani, Hadsell, Matias, Sculley, Petrov, Fiedel, Shazeer, Vinyals, Dean, Hassabis, Kavukcuoglu, Farabet, Buchatskaya, Alayrac, Anil, Dmitry, Lepikhin, Borgeaud, Bachem, Joulin, Andreev, Hardin, Dadashi, and Hussenot]{gemmateam2025gemma3technicalreport}
G.~Team, A.~Kamath, J.~Ferret, S.~Pathak, N.~Vieillard, R.~Merhej, S.~Perrin, T.~Matejovicova, A.~Ramé, M.~Rivière, L.~Rouillard, T.~Mesnard, G.~Cideron, J.~bastien Grill, S.~Ramos, E.~Yvinec, M.~Casbon, E.~Pot, I.~Penchev, G.~Liu, F.~Visin, K.~Kenealy, L.~Beyer, X.~Zhai, A.~Tsitsulin, R.~Busa-Fekete, A.~Feng, N.~Sachdeva, B.~Coleman, Y.~Gao, B.~Mustafa, I.~Barr, E.~Parisotto, D.~Tian, M.~Eyal, C.~Cherry, J.-T. Peter, D.~Sinopalnikov, S.~Bhupatiraju, R.~Agarwal, M.~Kazemi, D.~Malkin, R.~Kumar, D.~Vilar, I.~Brusilovsky, J.~Luo, A.~Steiner, A.~Friesen, A.~Sharma, A.~Sharma, A.~M. Gilady, A.~Goedeckemeyer, A.~Saade, A.~Feng, A.~Kolesnikov, A.~Bendebury, A.~Abdagic, A.~Vadi, A.~György, A.~S. Pinto, A.~Das, A.~Bapna, A.~Miech, A.~Yang, A.~Paterson, A.~Shenoy, A.~Chakrabarti, B.~Piot, B.~Wu, B.~Shahriari, B.~Petrini, C.~Chen, C.~L. Lan, C.~A. Choquette-Choo, C.~Carey, C.~Brick, D.~Deutsch, D.~Eisenbud, D.~Cattle, D.~Cheng, D.~Paparas, D.~S. Sreepathihalli, D.~Reid, D.~Tran, D.~Zelle, E.~Noland, E.~Huizenga,
  E.~Kharitonov, F.~Liu, G.~Amirkhanyan, G.~Cameron, H.~Hashemi, H.~Klimczak-Plucińska, H.~Singh, H.~Mehta, H.~T. Lehri, H.~Hazimeh, I.~Ballantyne, I.~Szpektor, I.~Nardini, J.~Pouget-Abadie, J.~Chan, J.~Stanton, J.~Wieting, J.~Lai, J.~Orbay, J.~Fernandez, J.~Newlan, J.~yeong Ji, J.~Singh, K.~Black, K.~Yu, K.~Hui, K.~Vodrahalli, K.~Greff, L.~Qiu, M.~Valentine, M.~Coelho, M.~Ritter, M.~Hoffman, M.~Watson, M.~Chaturvedi, M.~Moynihan, M.~Ma, N.~Babar, N.~Noy, N.~Byrd, N.~Roy, N.~Momchev, N.~Chauhan, N.~Sachdeva, O.~Bunyan, P.~Botarda, P.~Caron, P.~K. Rubenstein, P.~Culliton, P.~Schmid, P.~G. Sessa, P.~Xu, P.~Stanczyk, P.~Tafti, R.~Shivanna, R.~Wu, R.~Pan, R.~Rokni, R.~Willoughby, R.~Vallu, R.~Mullins, S.~Jerome, S.~Smoot, S.~Girgin, S.~Iqbal, S.~Reddy, S.~Sheth, S.~Põder, S.~Bhatnagar, S.~R. Panyam, S.~Eiger, S.~Zhang, T.~Liu, T.~Yacovone, T.~Liechty, U.~Kalra, U.~Evci, V.~Misra, V.~Roseberry, V.~Feinberg, V.~Kolesnikov, W.~Han, W.~Kwon, X.~Chen, Y.~Chow, Y.~Zhu, Z.~Wei, Z.~Egyed, V.~Cotruta, M.~Giang, P.~Kirk,
  A.~Rao, K.~Black, N.~Babar, J.~Lo, E.~Moreira, L.~G. Martins, O.~Sanseviero, L.~Gonzalez, Z.~Gleicher, T.~Warkentin, V.~Mirrokni, E.~Senter, E.~Collins, J.~Barral, Z.~Ghahramani, R.~Hadsell, Y.~Matias, D.~Sculley, S.~Petrov, N.~Fiedel, N.~Shazeer, O.~Vinyals, J.~Dean, D.~Hassabis, K.~Kavukcuoglu, C.~Farabet, E.~Buchatskaya, J.-B. Alayrac, R.~Anil, Dmitry, Lepikhin, S.~Borgeaud, O.~Bachem, A.~Joulin, A.~Andreev, C.~Hardin, R.~Dadashi, and L.~Hussenot.
\newblock Gemma 3 technical report, 2025{\natexlab{a}}.
\newblock URL \url{https://arxiv.org/abs/2503.19786}.

\bibitem[Team et~al.(2025{\natexlab{b}})Team, Bai, Bao, Chen, Chen, Chen, Chen, Chen, Chen, Chen, Chen, Cui, Ding, Dong, Du, Du, Du, Du, Fan, Feng, Fu, Gao, Gao, Gao, Gao, Gu, Guan, Guo, Guo, Hu, Hao, He, He, He, Hong, Hu, Hu, Huang, Huang, Huang, Jiang, Jiang, Jin, Kang, Lai, Li, Li, Li, Li, Li, Li, Li, Li, Li, Lin, Lin, Lin, Liu, Liu, Liu, Liu, Liu, Liu, Liu, Liu, Liu, Liu, Liu, Liu, Liu, Liu, Liu, Lu, Lu, Ma, Ma, Ma, Mao, Mei, Men, Miao, Pan, Peng, Qin, Qu, Shang, Shi, Shi, Song, Su, Su, Sun, Sung, Tang, Tao, Teng, Wang, Wang, Wang, Wang, Wang, Wang, Wang, Wang, Wang, Wang, Wang, Wang, Wang, Wang, Wang, Wang, Wang, Wei, Wei, Wu, Wu, Wu, Xiao, Xie, Xiong, Xu, Xu, Xu, Xu, Xu, Xu, Xu, Xu, Xu, Xu, Yan, Yan, Yang, Yang, Yang, Yang, Yang, Yao, Yao, Ye, Ye, Yin, Yu, Yuan, Yuan, Yuan, Zhan, Zhang, Zhang, Zhang, Zhang, Zhang, Zhang, Zhang, Zhang, Zhang, Zhang, Zhang, Zhao, Zhao, Zheng, Zheng, Zhou, Zhou, Zhou, Zhu, Zhuang, and Zu]{kimiteam2025kimik2openagentic}
K.~Team, Y.~Bai, Y.~Bao, G.~Chen, J.~Chen, N.~Chen, R.~Chen, Y.~Chen, Y.~Chen, Y.~Chen, Z.~Chen, J.~Cui, H.~Ding, M.~Dong, A.~Du, C.~Du, D.~Du, Y.~Du, Y.~Fan, Y.~Feng, K.~Fu, B.~Gao, H.~Gao, P.~Gao, T.~Gao, X.~Gu, L.~Guan, H.~Guo, J.~Guo, H.~Hu, X.~Hao, T.~He, W.~He, W.~He, C.~Hong, Y.~Hu, Z.~Hu, W.~Huang, Z.~Huang, Z.~Huang, T.~Jiang, Z.~Jiang, X.~Jin, Y.~Kang, G.~Lai, C.~Li, F.~Li, H.~Li, M.~Li, W.~Li, Y.~Li, Y.~Li, Z.~Li, Z.~Li, H.~Lin, X.~Lin, Z.~Lin, C.~Liu, C.~Liu, H.~Liu, J.~Liu, J.~Liu, L.~Liu, S.~Liu, T.~Y. Liu, T.~Liu, W.~Liu, Y.~Liu, Y.~Liu, Y.~Liu, Y.~Liu, Z.~Liu, E.~Lu, L.~Lu, S.~Ma, X.~Ma, Y.~Ma, S.~Mao, J.~Mei, X.~Men, Y.~Miao, S.~Pan, Y.~Peng, R.~Qin, B.~Qu, Z.~Shang, L.~Shi, S.~Shi, F.~Song, J.~Su, Z.~Su, X.~Sun, F.~Sung, H.~Tang, J.~Tao, Q.~Teng, C.~Wang, D.~Wang, F.~Wang, H.~Wang, J.~Wang, J.~Wang, J.~Wang, S.~Wang, S.~Wang, Y.~Wang, Y.~Wang, Y.~Wang, Y.~Wang, Y.~Wang, Z.~Wang, Z.~Wang, Z.~Wang, C.~Wei, Q.~Wei, W.~Wu, X.~Wu, Y.~Wu, C.~Xiao, X.~Xie, W.~Xiong, B.~Xu, J.~Xu, J.~Xu, L.~H. Xu,
  L.~Xu, S.~Xu, W.~Xu, X.~Xu, Y.~Xu, Z.~Xu, J.~Yan, Y.~Yan, X.~Yang, Y.~Yang, Z.~Yang, Z.~Yang, Z.~Yang, H.~Yao, X.~Yao, W.~Ye, Z.~Ye, B.~Yin, L.~Yu, E.~Yuan, H.~Yuan, M.~Yuan, H.~Zhan, D.~Zhang, H.~Zhang, W.~Zhang, X.~Zhang, Y.~Zhang, Y.~Zhang, Y.~Zhang, Y.~Zhang, Y.~Zhang, Y.~Zhang, Z.~Zhang, H.~Zhao, Y.~Zhao, H.~Zheng, S.~Zheng, J.~Zhou, X.~Zhou, Z.~Zhou, Z.~Zhu, W.~Zhuang, and X.~Zu.
\newblock Kimi k2: Open agentic intelligence, 2025{\natexlab{b}}.
\newblock URL \url{https://arxiv.org/abs/2507.20534}.

\bibitem[Thrush et~al.(2025)Thrush, Potts, and Hashimoto]{thrush2025improvingpretrainingdatausing}
T.~Thrush, C.~Potts, and T.~Hashimoto.
\newblock Improving pretraining data using perplexity correlations, 2025.
\newblock URL \url{https://arxiv.org/abs/2409.05816}.

\bibitem[Van~der Vaart(2000)]{van2000asymptotic}
A.~W. Van~der Vaart.
\newblock \emph{Asymptotic statistics}, volume~3.
\newblock Cambridge university press, 2000.

\bibitem[van Hasselt et~al.(2015)van Hasselt, Guez, and Silver]{vanhasselt2015deepreinforcementlearningdouble}
H.~van Hasselt, A.~Guez, and D.~Silver.
\newblock Deep reinforcement learning with double q-learning, 2015.
\newblock URL \url{https://arxiv.org/abs/1509.06461}.

\bibitem[Veit et~al.(2016)Veit, Wilber, and Belongie]{veit2016residualnetworksbehavelike}
A.~Veit, M.~Wilber, and S.~Belongie.
\newblock Residual networks behave like ensembles of relatively shallow networks, 2016.
\newblock URL \url{https://arxiv.org/abs/1605.06431}.

\bibitem[Villalobos et~al.(2024)Villalobos, Ho, Sevilla, Besiroglu, Heim, and Hobbhahn]{villalobos2024rundatalimitsllm}
P.~Villalobos, A.~Ho, J.~Sevilla, T.~Besiroglu, L.~Heim, and M.~Hobbhahn.
\newblock Will we run out of data? limits of llm scaling based on human-generated data, 2024.
\newblock URL \url{https://arxiv.org/abs/2211.04325}.

\bibitem[Vyas et~al.(2023)Vyas, Atanasov, Bordelon, Morwani, Sainathan, and Pehlevan]{vyas2023featurelearningnetworksconsistentwidths}
N.~Vyas, A.~Atanasov, B.~Bordelon, D.~Morwani, S.~Sainathan, and C.~Pehlevan.
\newblock Feature-learning networks are consistent across widths at realistic scales, 2023.
\newblock URL \url{https://arxiv.org/abs/2305.18411}.

\bibitem[Wang et~al.(2025)Wang, Zhou, Li, and Liu]{wang2025octothinkermidtrainingincentivizesreinforcement}
Z.~Wang, F.~Zhou, X.~Li, and P.~Liu.
\newblock Octothinker: Mid-training incentivizes reinforcement learning scaling, 2025.
\newblock URL \url{https://arxiv.org/abs/2506.20512}.

\bibitem[Warstadt et~al.(2023)Warstadt, Choshen, Mueller, Williams, Wilcox, and Zhuang]{warstadt2023papersbabylmchallenge}
A.~Warstadt, L.~Choshen, A.~Mueller, A.~Williams, E.~Wilcox, and C.~Zhuang.
\newblock Call for papers -- the babylm challenge: Sample-efficient pretraining on a developmentally plausible corpus, 2023.
\newblock URL \url{https://arxiv.org/abs/2301.11796}.

\bibitem[Welbl et~al.(2017)Welbl, Liu, and Gardner]{welbl2017crowdsourcingmultiplechoicescience}
J.~Welbl, N.~F. Liu, and M.~Gardner.
\newblock Crowdsourcing multiple choice science questions, 2017.
\newblock URL \url{https://arxiv.org/abs/1707.06209}.

\bibitem[Wen et~al.(2025)Wen, Hall, Ma, and Liang]{wen2025fantasticpretrainingoptimizers}
K.~Wen, D.~Hall, T.~Ma, and P.~Liang.
\newblock Fantastic pretraining optimizers and where to find them, 2025.
\newblock URL \url{https://arxiv.org/abs/2509.02046}.

\bibitem[Wortsman et~al.(2022)Wortsman, Ilharco, Gadre, Roelofs, Gontijo-Lopes, Morcos, Namkoong, Farhadi, Carmon, Kornblith, and Schmidt]{wortsman2022modelsoupsaveragingweights}
M.~Wortsman, G.~Ilharco, S.~Y. Gadre, R.~Roelofs, R.~Gontijo-Lopes, A.~S. Morcos, H.~Namkoong, A.~Farhadi, Y.~Carmon, S.~Kornblith, and L.~Schmidt.
\newblock Model soups: averaging weights of multiple fine-tuned models improves accuracy without increasing inference time, 2022.
\newblock URL \url{https://arxiv.org/abs/2203.05482}.

\bibitem[Xie et~al.(2017)Xie, Wang, Li, Lévy, Nie, Jurafsky, and Ng]{xie2017datanoisingsmoothingneural}
Z.~Xie, S.~I. Wang, J.~Li, D.~Lévy, A.~Nie, D.~Jurafsky, and A.~Y. Ng.
\newblock Data noising as smoothing in neural network language models, 2017.
\newblock URL \url{https://arxiv.org/abs/1703.02573}.

\bibitem[Yang et~al.(2025)Yang, Li, Yang, Zhang, Hui, Zheng, Yu, Gao, Huang, Lv, Zheng, Liu, Zhou, Huang, Hu, Ge, Wei, Lin, Tang, Yang, Tu, Zhang, Yang, Yang, Zhou, Zhou, Lin, Dang, Bao, Yang, Yu, Deng, Li, Xue, Li, Zhang, Wang, Zhu, Men, Gao, Liu, Luo, Li, Tang, Yin, Ren, Wang, Zhang, Ren, Fan, Su, Zhang, Zhang, Wan, Liu, Wang, Cui, Zhang, Zhou, and Qiu]{yang2025qwen3technicalreport}
A.~Yang, A.~Li, B.~Yang, B.~Zhang, B.~Hui, B.~Zheng, B.~Yu, C.~Gao, C.~Huang, C.~Lv, C.~Zheng, D.~Liu, F.~Zhou, F.~Huang, F.~Hu, H.~Ge, H.~Wei, H.~Lin, J.~Tang, J.~Yang, J.~Tu, J.~Zhang, J.~Yang, J.~Yang, J.~Zhou, J.~Zhou, J.~Lin, K.~Dang, K.~Bao, K.~Yang, L.~Yu, L.~Deng, M.~Li, M.~Xue, M.~Li, P.~Zhang, P.~Wang, Q.~Zhu, R.~Men, R.~Gao, S.~Liu, S.~Luo, T.~Li, T.~Tang, W.~Yin, X.~Ren, X.~Wang, X.~Zhang, X.~Ren, Y.~Fan, Y.~Su, Y.~Zhang, Y.~Zhang, Y.~Wan, Y.~Liu, Z.~Wang, Z.~Cui, Z.~Zhang, Z.~Zhou, and Z.~Qiu.
\newblock Qwen3 technical report, 2025.
\newblock URL \url{https://arxiv.org/abs/2505.09388}.

\bibitem[Yang et~al.(2022)Yang, Hu, Babuschkin, Sidor, Liu, Farhi, Ryder, Pachocki, Chen, and Gao]{yang2022tensorprogramsvtuning}
G.~Yang, E.~J. Hu, I.~Babuschkin, S.~Sidor, X.~Liu, D.~Farhi, N.~Ryder, J.~Pachocki, W.~Chen, and J.~Gao.
\newblock Tensor programs v: Tuning large neural networks via zero-shot hyperparameter transfer, 2022.
\newblock URL \url{https://arxiv.org/abs/2203.03466}.

\bibitem[Yang et~al.(2018)Yang, Dai, Salakhutdinov, and Cohen]{yang2018breakingsoftmaxbottleneckhighrank}
Z.~Yang, Z.~Dai, R.~Salakhutdinov, and W.~W. Cohen.
\newblock Breaking the softmax bottleneck: A high-rank rnn language model, 2018.
\newblock URL \url{https://arxiv.org/abs/1711.03953}.

\bibitem[Yang et~al.(2024)Yang, Band, Li, Candès, and Hashimoto]{yang2024syntheticcontinuedpretraining}
Z.~Yang, N.~Band, S.~Li, E.~Candès, and T.~Hashimoto.
\newblock Synthetic continued pretraining, 2024.
\newblock URL \url{https://arxiv.org/abs/2409.07431}.

\bibitem[Zaremba et~al.(2015)Zaremba, Sutskever, and Vinyals]{zaremba2015recurrentneuralnetworkregularization}
W.~Zaremba, I.~Sutskever, and O.~Vinyals.
\newblock Recurrent neural network regularization, 2015.
\newblock URL \url{https://arxiv.org/abs/1409.2329}.

\bibitem[Zhang et~al.(2019)Zhang, Song, Gao, Chen, Bao, and Ma]{zhang2019teacherimproveperformanceconvolutional}
L.~Zhang, J.~Song, A.~Gao, J.~Chen, C.~Bao, and K.~Ma.
\newblock Be your own teacher: Improve the performance of convolutional neural networks via self distillation, 2019.
\newblock URL \url{https://arxiv.org/abs/1905.08094}.

\bibitem[Zilly et~al.(2017)Zilly, Srivastava, Koutník, and Schmidhuber]{zilly2017recurrenthighwaynetworks}
J.~G. Zilly, R.~K. Srivastava, J.~Koutník, and J.~Schmidhuber.
\newblock Recurrent highway networks, 2017.
\newblock URL \url{https://arxiv.org/abs/1607.03474}.

\end{thebibliography}


\newpage
\appendix
\tableofcontents

\section{Problem setting}\label{app:problem-setting}

\paragraph{Pre-training algorithm.} We instantiate the pre-training algorithm $\trainalg$ using a standard pre-training recipe developed through the Marin project (\url{https://marin.community}) by following the best practices shared publicly and found internally.

\begin{itemize}
    \item \textbf{Optimizer.} We train with AdamW, either with a default of $0.1$ or a tuned weight decay. We set other hyperparameters to standard defaults ($\beta_1 = 0.9, \beta_2 = 0.95, \epsilon = 10^{-8}$). We clip the norm of the gradient at $1$.
    \item \textbf{Learning rate.} We use a cosine learning rate schedule with a warmup for the first $1\%$ of training, decaying to 0 by the end of training. We always tune learning rate for all of our experiments. Every run has its own learning rate schedule and we never report the loss before the learning rate anneals to zero in the main body .
    \item \textbf{Architecture.} We train Llama-style auto-regressive language models. We specify the main architectural choices in Table~\ref{tab:model_configs}. When scaling models, we change the initialization scheme to have variance inversely proportional to the hidden dimension (this is known to outperform $\mu$P~\citep{yang2022tensorprogramsvtuning} within our framework: \url{https://github.com/marin-community/marin/issues/621}. For other architectural choices, we default to SiLU activations, untied word embeddings, and rotary position embeddings. We note that the non-standard scaling for the 1.4B model is a consequence of using presets in our pre-training framework.
    \item \textbf{Systems.} We train in mixed-precision with parameters in fp32 and compute + output in bf16. Most of our jobs were run on v4-64 or v4-128 TPUs, with bitwise-determinism for handling preemption and promoting reproducibility. 
    \item \textbf{Data.} We pre-train using DCLM  data~\citep{li2025datacomplmsearchgenerationtraining}. We keep a fixed validation set of 1024 sequences (4 million tokens) across all experiments for clear comparison. When increasing the size of the train pool, we ensure that smaller pools are a subset of larger pools. 
    \item \textbf{Data order.} We generate a random permutation of the windows after tokenization and use this same permutation across epochs. We note that performance might have been better if we used a unique permutation every epoch but keep this fixed across models which further reduces randomness of training.
\end{itemize}

\begin{table}[h]
\centering
\begin{tabular}{lrrrr}
\toprule
\textbf{Parameter} & \textbf{150M} & \textbf{300M} & \textbf{600M} & \textbf{1.4B} \\
\midrule
Context Length & 4096 & 4096 & 4096 & 4096 \\
Hidden Dimension & 512 & 768 & 1024 & 2048 \\
Intermediate Dimension & 1792 & 2688 & 7168 & 7168 \\
Attention Heads & 8 & 12 & 16 & 16 \\
KV Heads & 8 & 12 & 8 & 8 \\
Layers & 6 & 12 & 24 & 16 \\
\bottomrule
\end{tabular}
\vspace{1em}
\caption{Model architecture configurations for different model sizes. We default to the 300M model if not specified.}
\label{tab:model_configs}
\end{table}

\section{Standard pre-training details}

\subsection{Locally optimal hyperparameters}\label{app:convex-certificates}

We are interested in finding the best setting of hyperparameters (e.g. learning rate, epoch count, and weight decay) for a fixed parameter count and token count in the data-efficient pre-training setting. To make this search problem tractable, we first discretize the space of hyperparameters (e.g. only try epoch counts that are powers of 2). Under this discretization, it is prohibitively expensive to try every possible hyperparameter selection within our grid. Therefore, we search for \textbf{locally-optimal hyperparameters} as defined below.

\begin{definition}[Locally-optimal hyperparameters]
We define the neighborhood $\ball(H)$ of hyperparameter tuple $H$ containing $m$ variables as the $2m$ neighbors from incrementing/decrementing exactly one of the variables. We say $H$ is locally optimal if and only if
$$\forall H' \in \ball(H),\;\loss\p{\trainalg\p{D, N, H}} \leq \loss\p{\trainalg\p{D, N, H'}}$$
\end{definition}

Under certain assumptions, the locally-optimal hyperparameters are also globally optimal (for example, if the loss was convex in each dimension when the other variables are fixed). Though this may seem like a big assumption, we did not observe counter-examples to this in early experiments. One can verify this property for the single model losses presented in Figure~\ref{fig:ensemble-heuristic-justification}. 

Assuming this property, we use the following search procedure
\begin{enumerate}
    \item Seed the search with initial runs around a best guess for optimal hyperparameters (heuristically set by us, using all runs until this point in time)
    \item Take the best run so far and run its neighbors. 
    \item If any of its neighbors is better than the current candidate, the current candidate is sub-optimal. Repeat step 2 with the new best run.
    \item If this run is better than all of its neighbors, we consider it ``certified'' and terminate the search procedure.
\end{enumerate}

Though this procedure is quite expensive, we found it better than other natural heuristics which don't rigorously follow this procedure separately for each parameter and token count (Appendix~\ref{app:ablate-convex-tuning}). Furthermore, it seems to give clean scaling in both loss and hyperparameters, suggesting that the hyperparameter optimization landscape is nice enough for this coordinate descent algorithm to work. We note that this is a simplified version of the assumptions used in~\cite{wen2025fantasticpretrainingoptimizers}.

Our specific discretization was forcing learning rate to be $1$ or $3$ times a power of 10, epoch count to be an integer power of 2, and weight decay to be either $0.0$ or an integer power of $2$ times $0.1$. To further restrict the search space, we set bounds based on initial experiments tuning these hyperparameters, with a maximum learning rate of 3e-3, maximum weight decay of $6.4$, and a maximum epoch count of $64$ (these bounds only triggered for three of our searches).

\subsection{Ablating on coordinate descent}\label{app:ablate-convex-tuning}

\begin{figure}
  \centering
  \begin{minipage}[c]{0.4\textwidth}
    \centering
    \includegraphics[width=\linewidth]{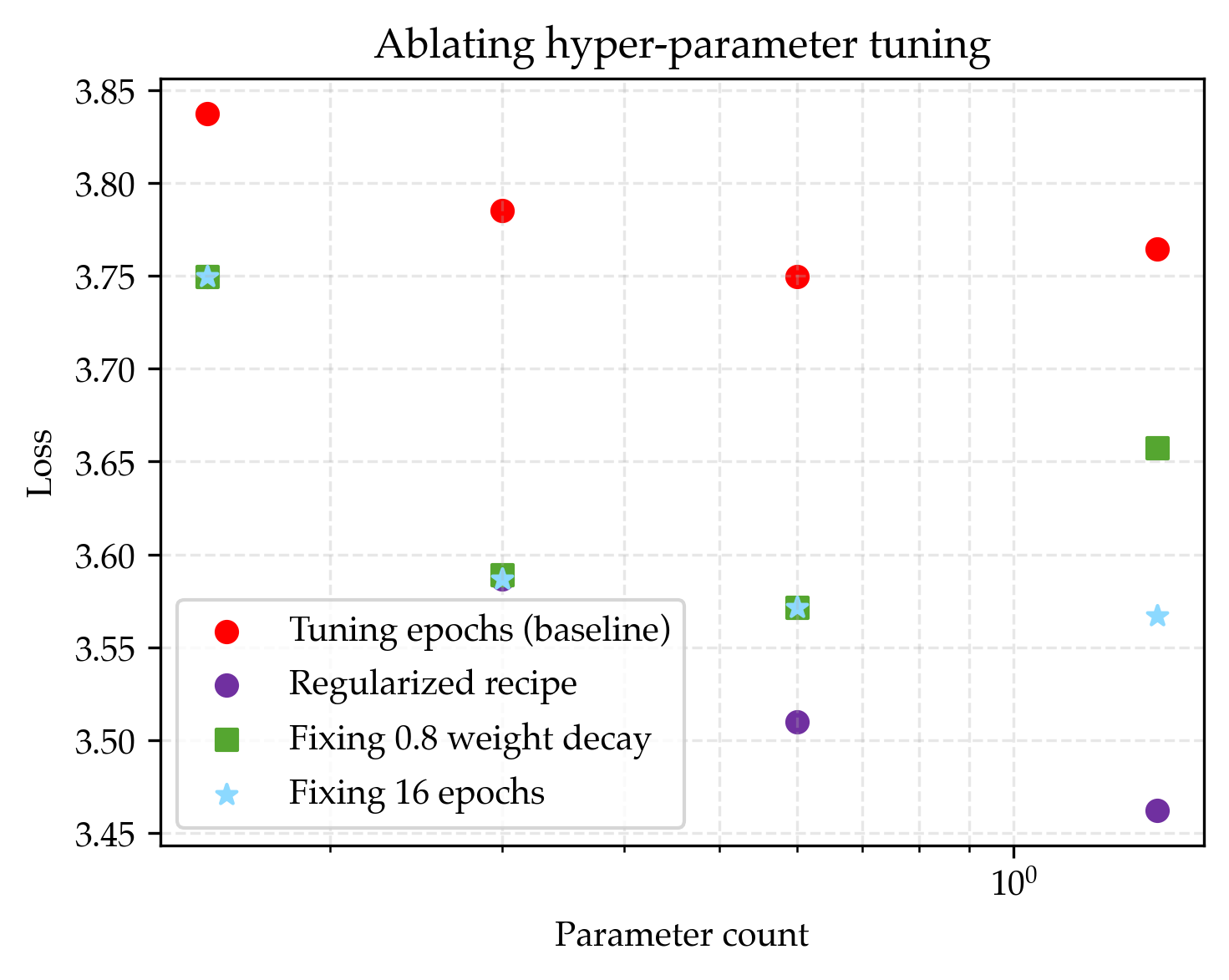}
  \end{minipage}
  \hfill
  \begin{minipage}[c]{0.55\textwidth}
    \captionof{figure}{\textbf{Ablating joint tuning procedure.} We generally jointly tune learning rate, epoch count, and weight decay separately for each parameter count (purple). We show that trying to naively transfer hyperparameters across scales is a bad idea. Red: fixing weight decay to $0.1$ (default regularization). Green: assuming that the optimal weight decay for 150M models ($0.8$) is optimal across $N$. Blue: assuming that the optimal epoch count for 150M models ($16$) is optimal across $N$.}
    \label{fig:ablating-convex-tuning}
  \end{minipage}
\end{figure}

For the parameter scaling experiments, we jointly tune learning rate, epoch count, and weight decay. We find this joint tuning necessary for clean scaling. To demonstrate this, we consider three alternatives and show their scaling in Figure~\ref{fig:ablating-convex-tuning}.

\begin{enumerate}
    \item \textbf{Fixing weight decay $0.1$ and tuning learning rate and epoch count (red).} This baseline has already been shown to fail in Section~\ref{sec:chinchilla-fails}.
    \item \textbf{Jointly tuning weight decay only for 150M (green).} Here, we jointly tune the weight decay for the 150M model, finding the optimal value is $0.8$. We then assume this is the optimal value for higher parameter counts and correspondingly tune learning rate and epoch count. We find that this scaling is not even monotonic.
    \item \textbf{Jointly tuning epoch count only for 150M (blue).} Here, we jointly tune the epoch count for the 150M model, finding the optimal value is $16$. We then assume this is the optimal value for higher parameter counts and correspondingly tune learning rate and weight decay. We find that this scaling plateaus much faster than the regularized recipe.
\end{enumerate}

This shows the importance of jointly tuning both weight decay and epoch count at each model scale instead of blindly hoping for transfer.

\subsection{Tuned hyperparameters}

\begin{figure}
    \centering
    \includegraphics[width=0.99\textwidth]{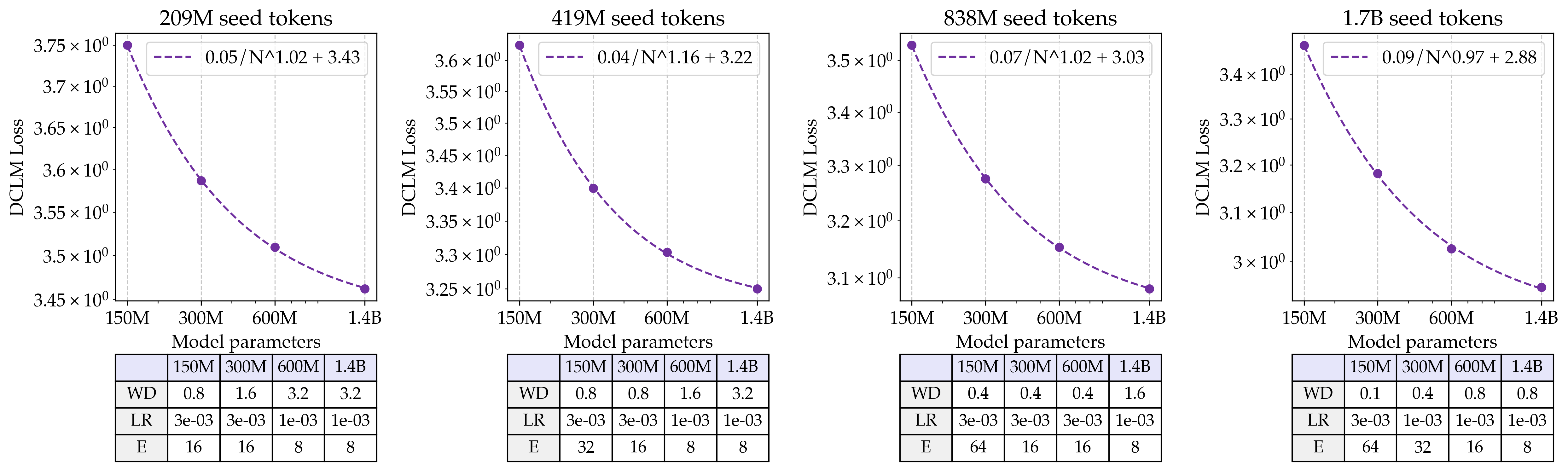}
    \caption{\textbf{Tuned hyperparameters for regularized scaling.} We show the optimal hyperparameters tuned seperately for each parameter and token count. We find that as parameter count increases, optimal weight decay goes up, optimal epoch count goes down, and optimal learning rate goes down. We find the trends for weight decay and epoch count hold when token count decreases.}
    \label{fig:dump-regularized-hypers}
\end{figure}

In Figure~\ref{fig:dump-regularized-hypers}, we share the locally optimal hyperparameters we found for 4 different token counts and 4 different parameter counts. We notice a few trends when looking at the resulting hyperparameters and power laws.

\begin{itemize}
    \item The optimal learning rate decreases for larger models, noted by prior work~\citep{yang2022tensorprogramsvtuning,everett2024scalingexponentsparameterizationsoptimizers}. The optimal learning rate does not strongly depend on the number of tokens.
    \item The optimal weight decay increases for larger models. Similarly, the optimal weight decay decreases for larger token counts. When fixing the parameter-to-token ratio, the weight decay stays around $0.8$.
    \item The optimal epoch count decreases for larger models. Similarly, the optimal epoch count increases for larger token counts. When fixing the parameter-to-token ratio, the epoch count stays around $16$.
    \item The power laws fit across all token counts share similar scaling exponents close to $1$. This holds even though almost every model is over-parameterized at 200M tokens and under-parameterized at 1.6B tokens.
\end{itemize}

\subsection{Hyperparameter ablations}\label{app:hyper-ablations}

We perform additional ablations on hyperparameters to understand their role beyond their optimal values. We start from a recipe of single epoch pre-training with $0.1$ weight decay and tuned learning rate, and build our way up to tuning all hyperparameters.

\begin{figure}
    \centering
    \begin{subfigure}[b]{0.45\textwidth}
        \centering
        \includegraphics[width=\textwidth]{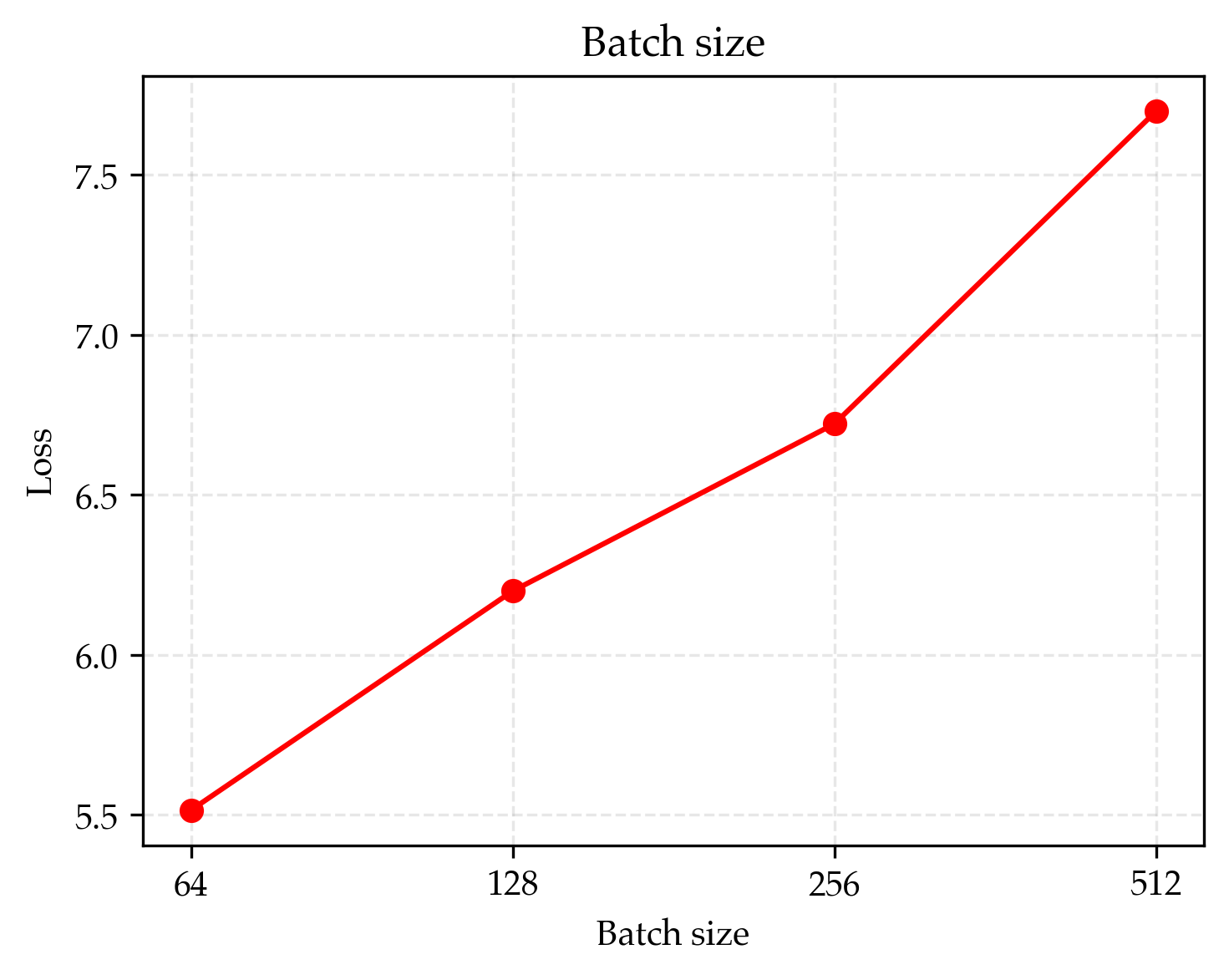}
    \end{subfigure}
    \hfill
    \begin{subfigure}[b]{0.45\textwidth}
        \centering
        \includegraphics[width=\textwidth]{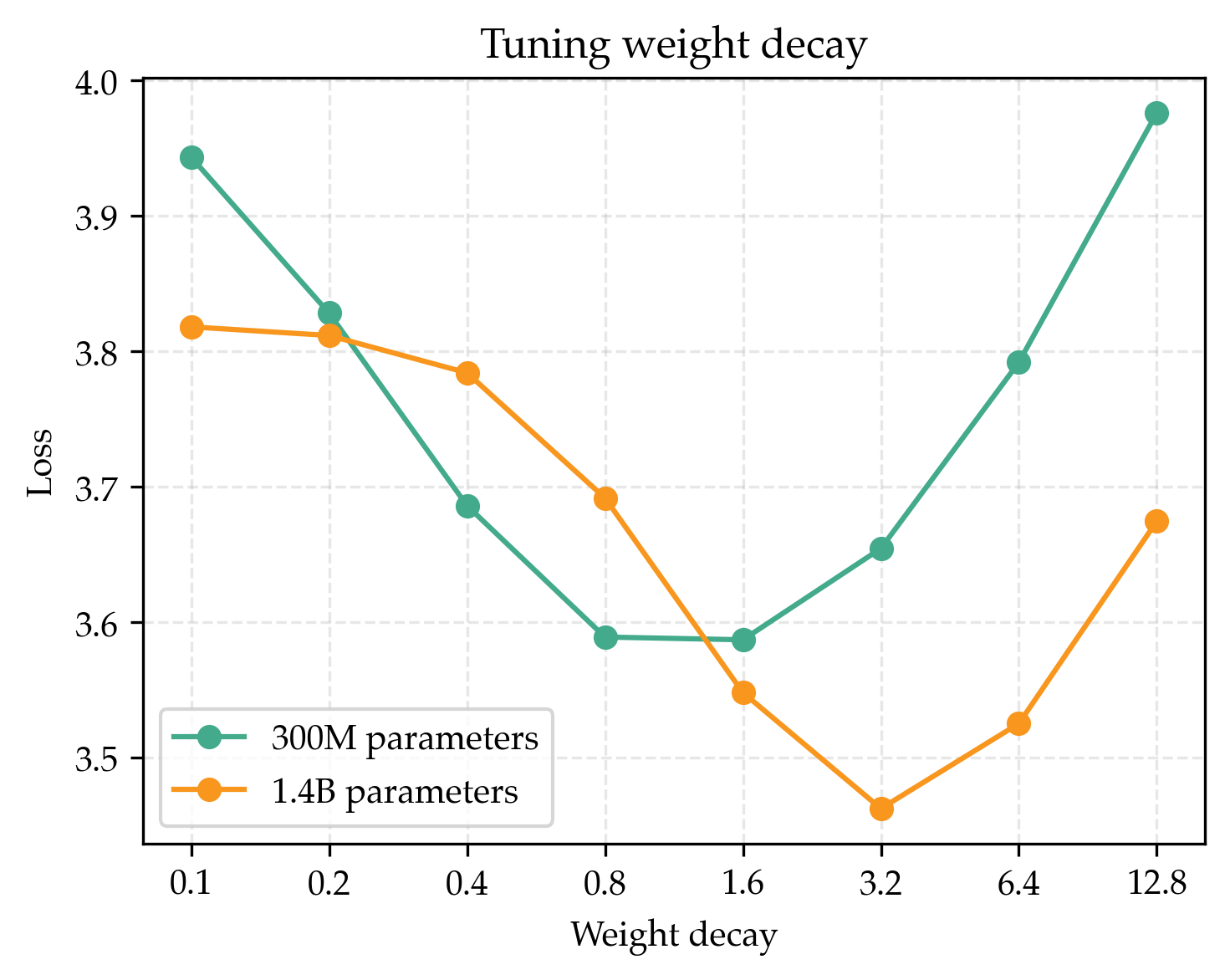}
    \end{subfigure}
    \caption{\textbf{Re-evaluating hyperparameters for standard pre-training.} Left: smaller batch sizes are better, we stop at 64 since this is the smallest our hardware practically supports (shown for 1 epoch training with 0.1 weight decay and a fixed learning rate of 3e-3). Right: weight decay helps, and the optimal weight decay is higher for larger models (300M is 16 epochs 3e-3 learning rate, 1.4B is 8 epochs 1e-3 learning rate).}
    \label{fig:sketch-pushing-hparams}
\end{figure}

\paragraph{Batch size.} It is known that when optimizing for throughput, it is best to train at the ``critical batch size'' to best utilizes hardware~\citep{mccandlish2018empiricalmodellargebatchtraining}. However if we drop this constraint and instead measure performance for a fixed number of data points, we find that it is best to use smaller batch sizes, as shown in Figure~\ref{fig:sketch-pushing-hparams}, left, corroborating prior work in optimization~\citep{smith2020generalizationbenefitnoisestochastic,keskar2017largebatchtrainingdeeplearning,LeCun1998,marek2025smallbatchsizetraining}. We use a batch size of $64$, which is the smallest size that is practical for our hardware.

\paragraph{Weight decay.} 
It is known that regularization can further improve generalization when repeating data~\citep{dangelo2024needweightdecaymodern}. Figure~\ref{fig:sketch-pushing-hparams}, right shows how varying the weight decay impacts epoched models (1.4B parameters for 8 epochs vs 300M parameters for 16 epochs). More over-parametrized models  require a larger amount of regularization ($3.2$ vs $1.6$). Without optimally tuning weight decay, one may draw the incorrect conclusion that larger models are worse than smaller models in the data-constrained setting. We also reproduce findings that higher weight decay enables a higher optimal learning rate and epoch count, shown by contrasing Figure~\ref{fig:sketch-chinchilla-fails} and Figure~\ref{fig:chinchilla-comparison}.

\begin{figure}
    \centering
    \includegraphics[width=0.9\textwidth]{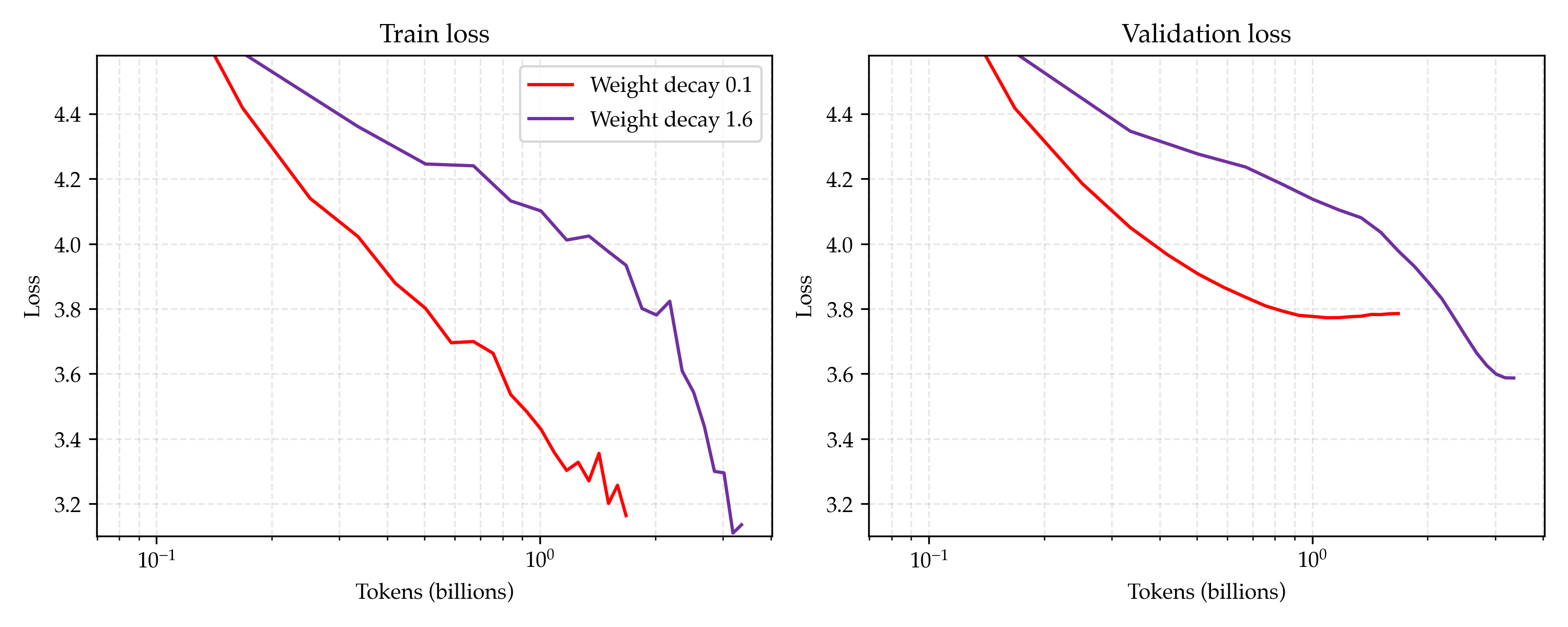}
    \caption{\textbf{Loss trajectories for different weight decays.} We compare the best run with default weight decay (8 epochs, 1e-3 learning rate, $0.1$ weight decay) and the best run with tuned weight decay (16 epochs, 3e-3 learning rate, $1.6$ weight decay) for 200M tokens and 300M parameters. We find that loss for runs with high weight decay decreases much more slowly at the start of training, but quickly decreases near the end of training.}
    \label{fig:weight-decay-trajectories}
\end{figure}

We find that increasing weight decay strongly changes the training dynamics. Though the train and validation losses decrease much slower at the start of training with high weight decay, they decrease rapidly by the end of training, eventually beating the run with low weight decay. In Figure~\ref{fig:weight-decay-trajectories}, we visualize this for the best run with weight decay $0.1$ and the best run tuning weight decay. This phenomenon also holds when using a higher weight decay on top of the best $0.1$ weight decay hyperparameters. This is in line with previous findings in optimization research~\citep{wen2025fantasticpretrainingoptimizers,liu2025muonscalablellmtraining,dangelo2024needweightdecaymodern} and suggests that we should only look at the performance at the end of training.

\subsection{Overfitting analysis}\label{app:overfitting}

\begin{figure}
    \centering
    \begin{subfigure}[b]{0.45\textwidth}
        \centering
        \includegraphics[width=\textwidth]{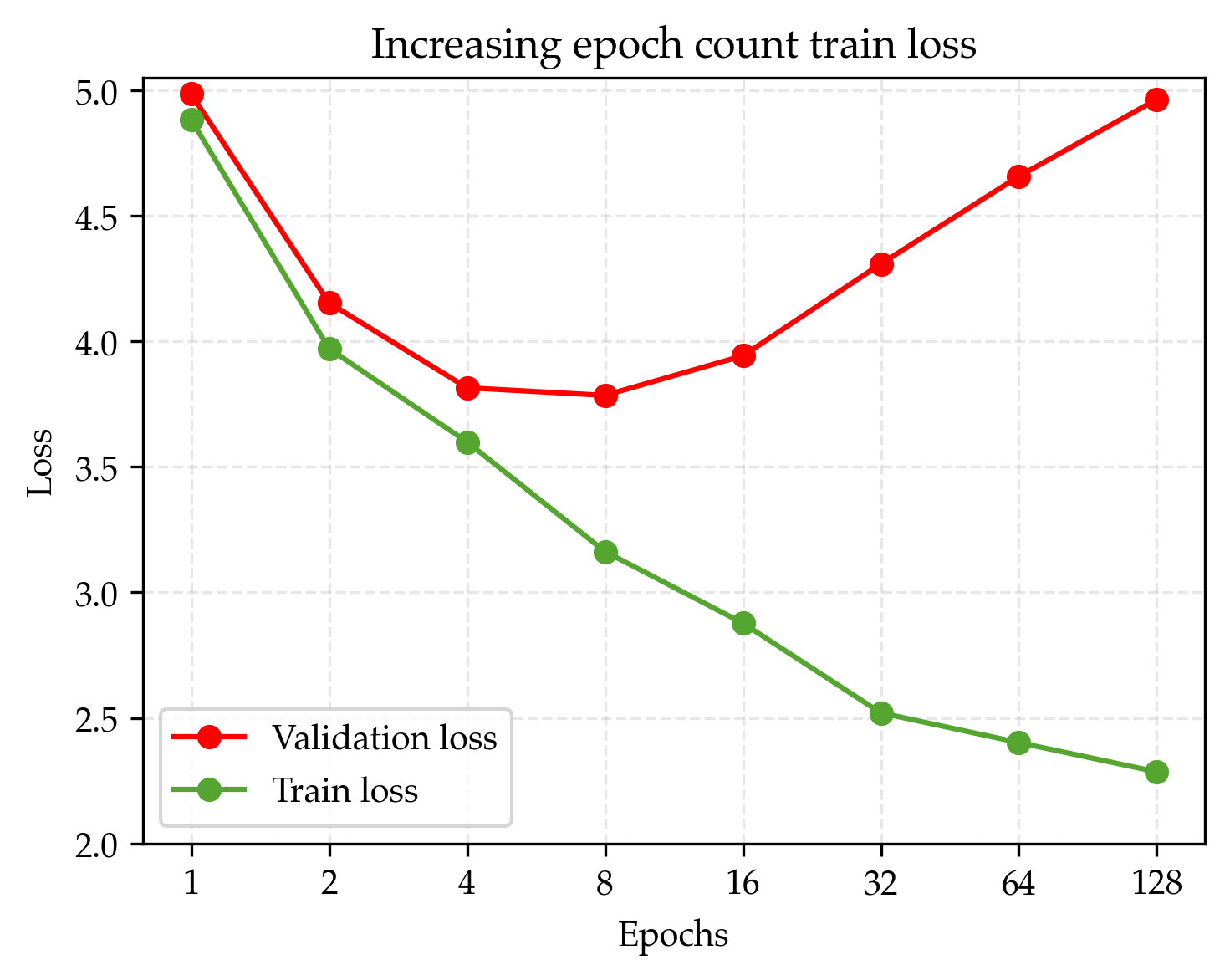}
    \end{subfigure}
    \begin{subfigure}[b]{0.45\textwidth}
        \centering
        \includegraphics[width=\textwidth]{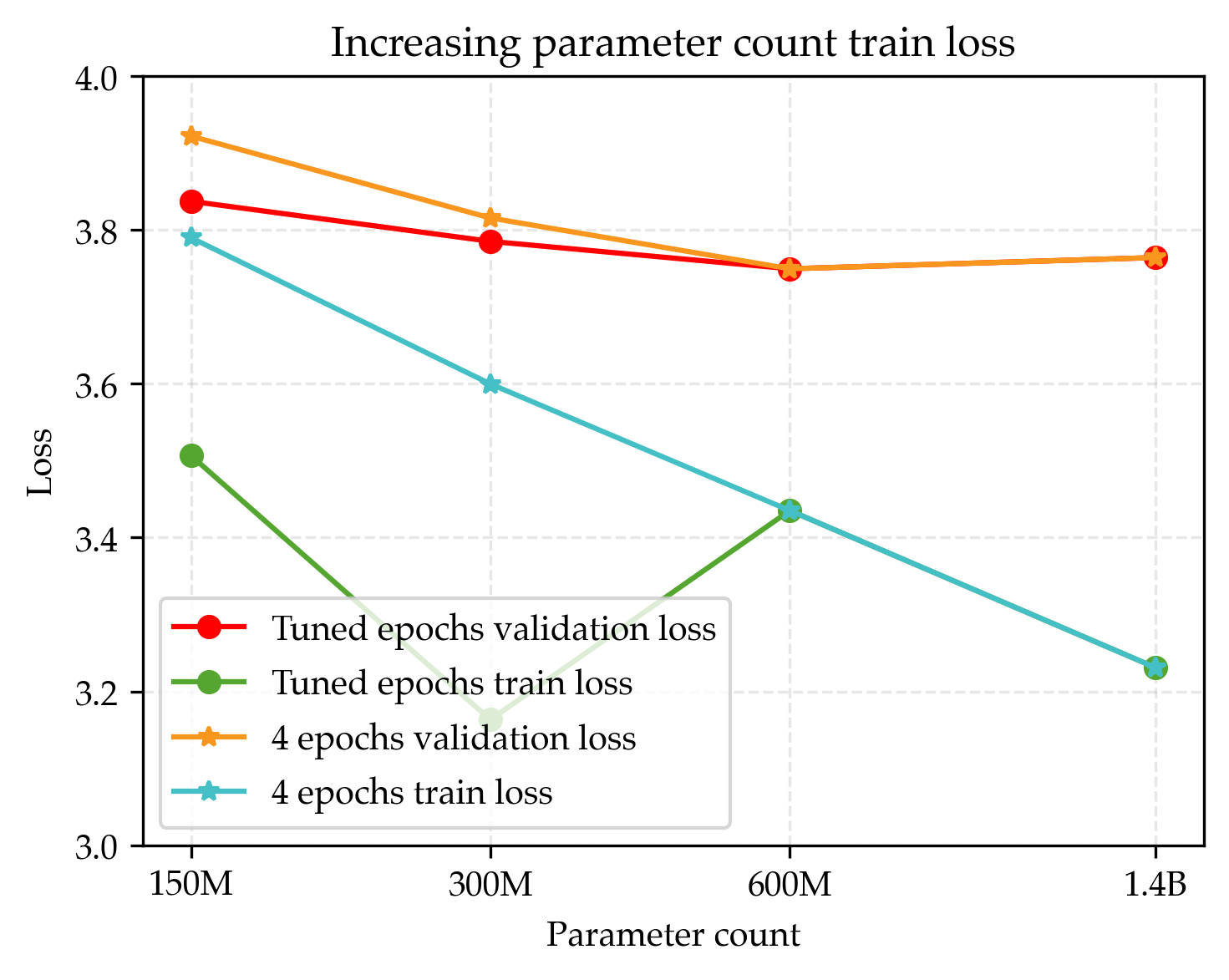}
    \end{subfigure}
    \caption{\textbf{Train losses for epoching and parameter scaling.} Left: Increasing epoch count results in train loss decreasing while validation loss starts increasing. Right: Increasing parameter count does not always decrease loss, potentially due to optimal epoch count changing (8, 8, 4, 4). Train loss monotonically goes down when restricted to 4 epochs.}
    \label{fig:train-loss}
\end{figure}

In Section~\ref{sec:chinchilla-fails}, we discuss how introducing too many epochs or parameters results in validation loss going up which we believe is due to over-fitting. In Figure~\ref{fig:train-loss}, we track train loss for the interventions of increasing epoch count and parameter count. On the left, we show how increasing epoch count monotonically decreases train loss but eventually results in validation loss going up. On the right, we show how increasing parameter count results in erratic changes in train loss. We hypothesize this is because the optimal epoch count changes from 8 for the first two models to 4 for the last two models. We find that when we restrict all models to only use 4 epochs, train loss decreases monotonically and validation loss still goes up, suggesting over-fitting.

We note that another reason over-fitting may be happening is because our 1.4B model trades depth for width. We did not recognize our model scaling was non-standard until the majority of our experiments had finished because these were the default settings in our pre-training framework. We do not think this is a severe issue since correctly tuning weight decay seems to correct for the fact that this architecture has less layers. Moreover, the large improvement from weight decay is also suggestive of the fact that larger models are over-fitting.

\section{Ensembling details}

\subsection{Seed science}\label{app:seed-science}
For our training runs, the randomness only comes from the sampled initialization (train seed) and shuffled data order (data seed). We first characterize the run-to-run variance when varying both seeds, only train seed, or only data seed. We train 5 models for each of the three randomness options. When using the optimal hyperparameters for a 300M model with 200M tokens, the standard deviation is estimated to be $0.008207$ for both seeds, $0.007605$ for only train seed, and $0.007213$ for only data seed. This is in line with prior work that shows how only a small amount of instability is needed to induce the majority of the variance over a final model's loss~\citep{summers2021nondeterminisminstabilityneuralnetwork,jordan2024varianceneuralnetworktraining}.

\begin{figure}
    \centering
    \includegraphics[width=0.6\textwidth]{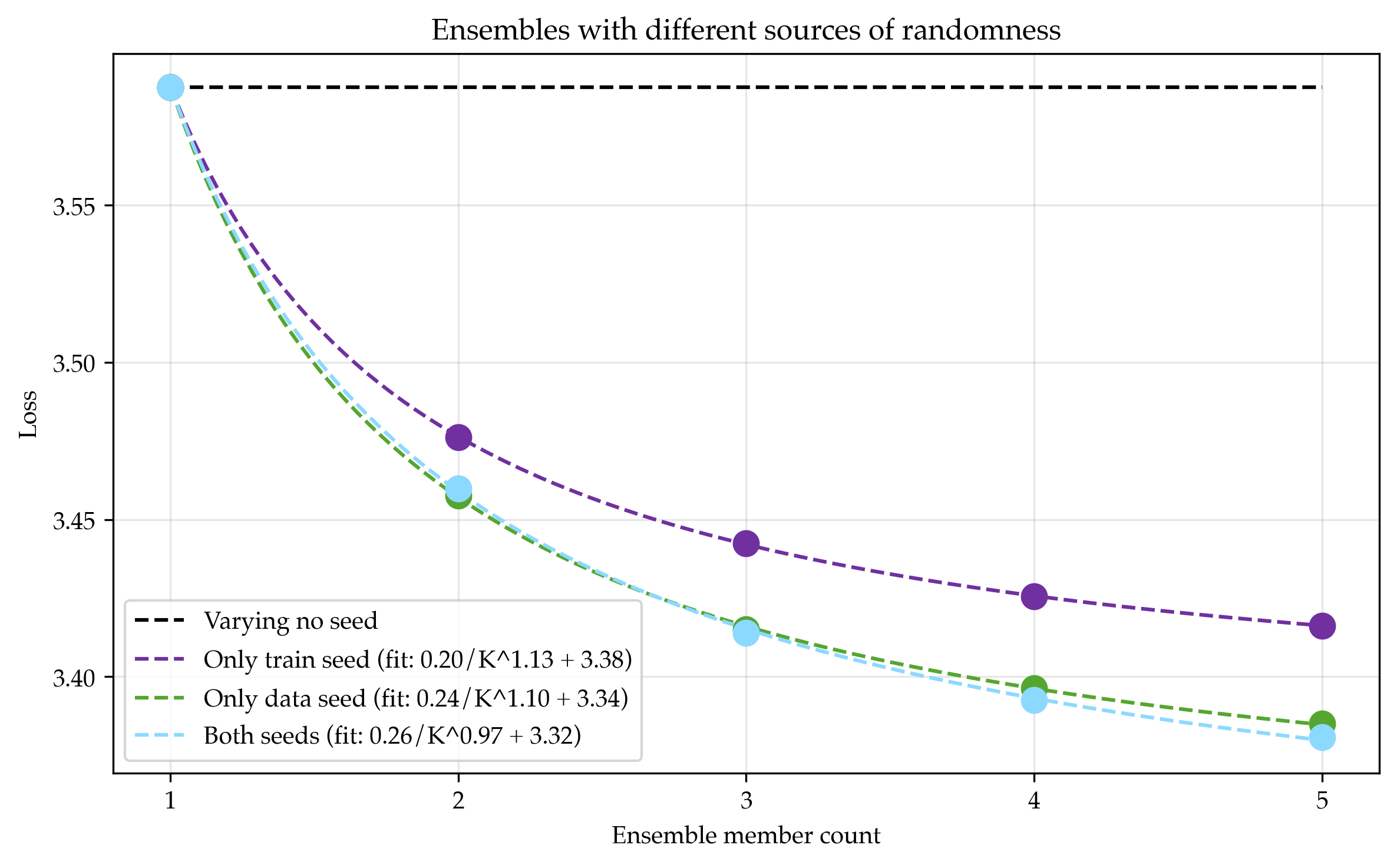}
    \caption{\textbf{Sources of randomness for ensembling.} Only varying train or data seed is enough to induce the benefits of ensembling. Varying data seed (i.e. order) is better between these two.}
    \label{fig:sketch-seed-science}
\end{figure}

We now measure the loss of ensembles with these sources of variance in Figure~\ref{fig:sketch-seed-science}. Either of these sources delivers most of the benefit of ensembling, with data order helping more. Considering the marginal benefit of introducing additional sources of randomness, we didn't strongly explore adding more sources of randomness during training.

\subsection{Hyperparameter tuning for ensembles}\label{app:heuristic-hparams}

Similar to parameter scaling, we hope to find the locally optimal hyperparameters for a given parameter count and token count. However, we care about the hyperparameters as $K\to\infty$, not at $K=1$ (as discussed in Section~\ref{sec:ens-vs-parameter}). Since it is too experimentally prohibitive to search for locally optimal hyperparameters for the asymptote, we study how the hyperparameters change relative to the optimal $H$ for single models.

\begin{figure}
    \centering
    \includegraphics[width=0.55\textwidth]{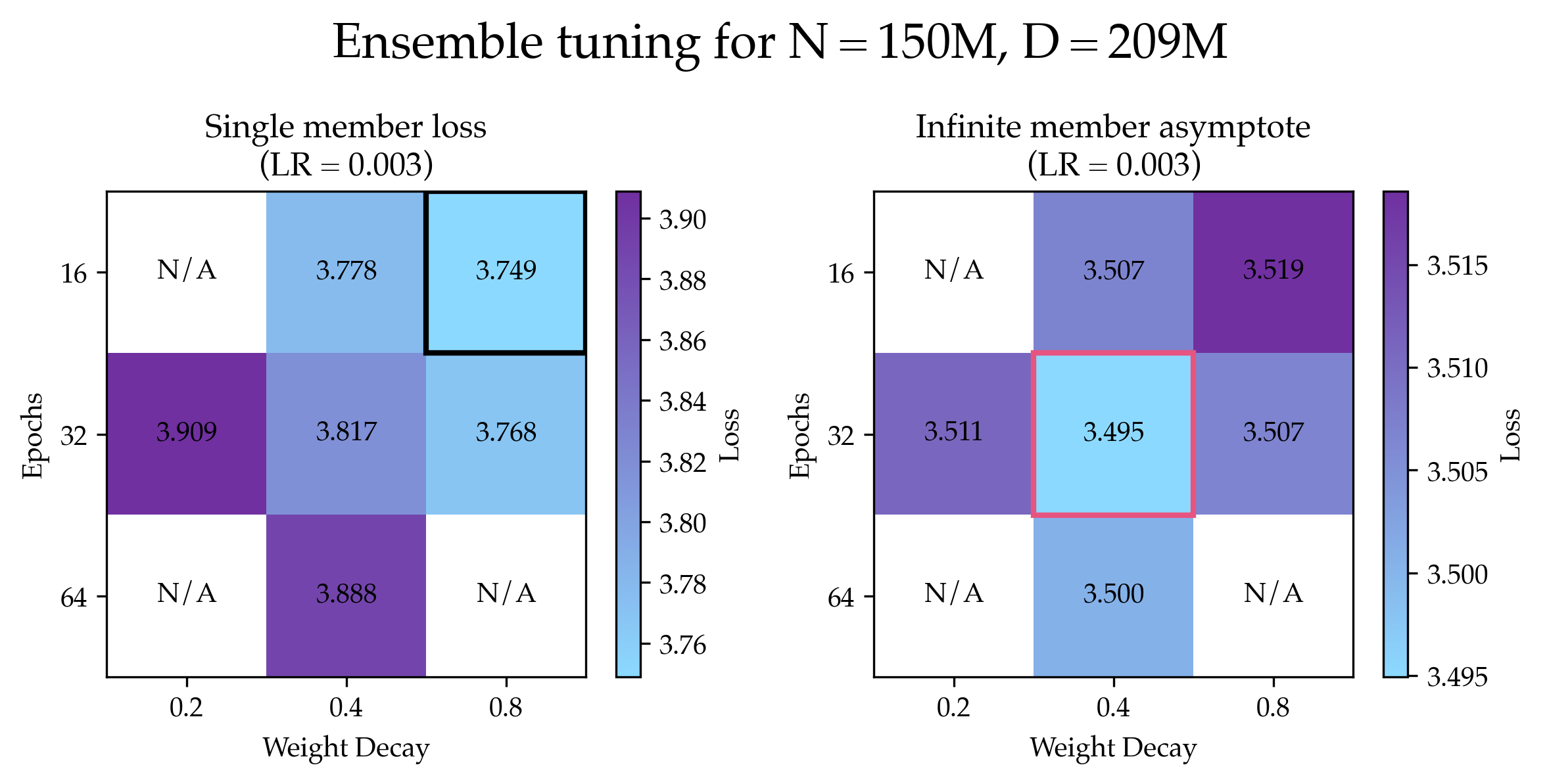}
    
    \includegraphics[width=0.55\textwidth]{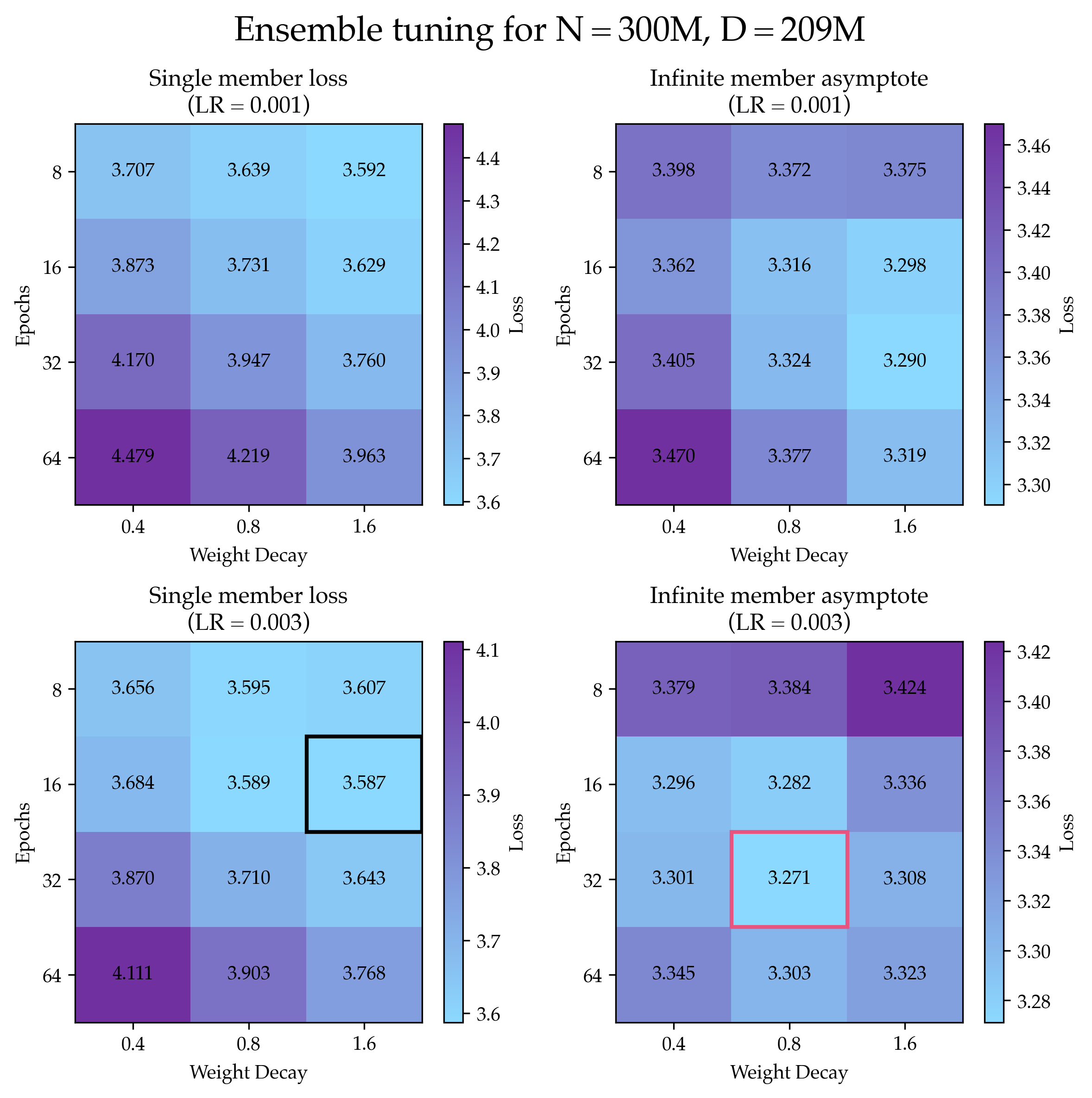}

    \includegraphics[width=0.55\textwidth]{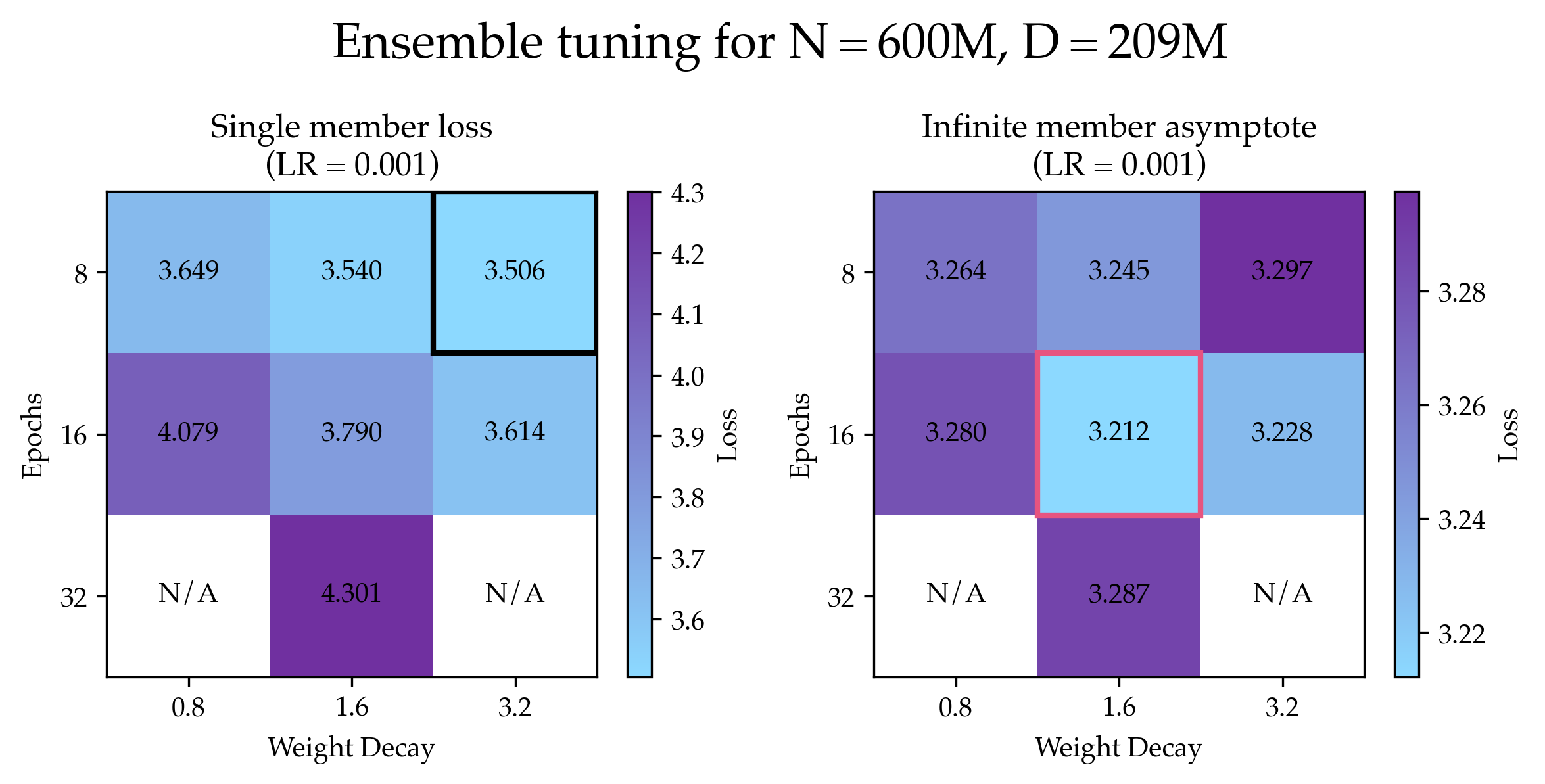}
    \caption{\textbf{Single model and asymptote loss when varying epoch count and weight decay for different model sizes, token counts, and learning rates.} We display an extended version of Figure~\ref{fig:sketch-ensemble-hparams} for 200M tokens with 150M, 300M (sub-optimal and optimal LR), and 600M parameter models. For these settings, the ``double epoch, half weight decay'' heuristic correctly predicts the best ensemble. This heuristic is consistent with all of our parameter and token counts except for our most over-parameterized setting.}
    \label{fig:ensemble-heuristic-justification}
\end{figure}

We repeat the analysis in Figure~\ref{fig:sketch-ensemble-hparams} for 3 different parameter counts as well as 2 different learning rates at 300M parameters, shown in Figure~\ref{fig:ensemble-heuristic-justification}. We find that for all of the displayed settings, our best run comes from an ensemble with the same learning rate, half weight decay, and double epoch count relative to the optimal single model hyperparameters. For the three parameter counts we display, we verify that these are locally optimal hyperparameters. In fact, across all of the ensembles we trained across our scales (including many that are not directly visualized), we find only one counter-example to this heuristic. This occurs for 1.4B models trained on 200M tokens, where the hyperparameters that minimize the asymptote do not halve the weight decay. We suspect this change occurs because this is our most over-parameterized setting, and our scaling laws use this best run over the heuristic for this single setting.

\subsection{Alternatives to ensembling}\label{app:alternatives}

The success of ensembling suggests alternative parameterizations that might also boost data efficiency. We discuss commonly considered ones here, which we were not able to get to outperform ensembling.

\subsubsection{Mixture-of-Experts}

Ensembling may qualitatively seem similar to training with Mixture-of-Experts (MoE). However, we find an important distinction: when training an ensemble, the learning trajectory for each model is completely independent of each other, whereas for a MoE, it is still a single learning trajectory. Unfortunately, the intuition from~\cite{allenzhu2023understandingensembleknowledgedistillation} suggests that the sparsity of the MoE architecture is not guaranteed to benefit from ``multi-view'' data in the way ensembles do. In their paper, they consider a simplified analogue where they construct a model that runs the ten models in parallel and takes the gradient step through the ensemble jointly. They find that this barely improves performance over a single model. In early experiments, we were able to reproduce this phenomenon, with a jointly trained 10 ensemble of models outperforming a single model by only $0.02$ loss. Therefore, we hypothesize that if MoE's were to help, their benefits would come from the drop-out aspect instead of the sparsity aspect, which does not require MoE's (note that we did not tune or consider drop-out in this work, though we expect it to further help performance). We hope future experiments can settle this intuition more concretely.

\subsubsection{Model soups}

Prior works have shown that averaging the weights of independent training runs can result in better models~\citep{wortsman2022modelsoupsaveragingweights}. However, we note that most success from averaging weights comes at fine-tuning, not pre-training. We replicate these results in our own settings, with model soups achieving close to random performance on downstream benchmarks (Table~\ref{tab:appendix_benchmarks_indented}) but slightly outperforming ensembles in continued pre-training (Table~\ref{tab:cpt-soups}).

One intuition for this discrepancy is that models need to be in the same ``loss basin'' for averaging to help final performance, and pre-trained models enter different loss basins~\citep{singh2023modelfusionoptimaltransport,ainsworth2023gitrebasinmergingmodels}. Past studies also design compute-efficient algorithms for merging models trained from scratch, but they find that the more expensive procedure of distillation outperforms their method~\citep{singh2023modelfusionoptimaltransport}. Since we are in the infinite compute regime, we opt to use distillation over model merging for the best performance.

\subsection{Order of limits}\label{app:order-of-limits}

In Section~\ref{sec:infinite-ensembles}, we are interested in computing the best possible performance of ensembles as $N$ and $K$ both go to $\infty$. There are a couple of different ways to compute this, some of which are enumerated below 

\paragraph{Double Limit 1 (Our Approach).} We first solve for the limit as $K\to\infty$ by tuning asymptotes and then solve for the outer limit via a second power law over the inner asymptotes. $$\lim_{N\to\infty} \lim_{K\to\infty} \min_{\substack{H}} \loss\p{\ensalg_{\trainalg}\p{D, N, K, H}}$$

\paragraph{Hypothetical Double Limit 2.} We can flip the above order and instead take $N\to\infty$ before $K\to\infty$, corresponding to $$\lim_{K\to\infty} \lim_{N\to\infty} \min_{\substack{H}} \loss\p{\ensalg_{\trainalg}\p{D, N, K, H}}$$

\paragraph{Hypothetical Double Limit 3.} Following literature in compute-optimal scaling, we can find the best possible performance for a given compute budget $C$ and take $C\to\infty$.
$$\lim_{C\to\infty} \min_{\substack{H, N, K\\\text{s.t. FLOPs}(D,N,K,H)=C}} \loss\p{\ensalg_{\trainalg}\p{D, N, K, H}}$$

We believe that our approach is experimentally much more convenient than the hypothetical approaches even though they are equivalent in output under assumptions. We share our reasoning by answering the following questions comparing the approaches.

The core assumption that we will make is that $f(N, K) = \min_{\substack{H}} \loss\p{\ensalg_{\trainalg}\p{D, N, K, H}}$ is monotone in $N$ and $K$ when the other is fixed. Across all of our experiments, we do not observe any contradictions to this assumption as long as we tune regularization (for examples, refer to Figure~\ref{fig:seed-token-scaling-standard} and Figure~\ref{fig:seed-token-scaling-ensembling}).

\begin{itemize}
    \item \textbf{Are Double Limit 1 and 2 mathematically equivalent?} Mathematically, both limits are equivalent. For a quick proof, define $k_i := \lim_{K\to\infty} f(D, N)$ and $n_i := \lim_{N\to\infty} f(D, N)$. By monotonicity, both $k_i$ and $n_i$ exist and are non-increasing sequences. Define $k = \lim_{N\to\infty} k_i$ and $n = \lim_{K\to\infty} n_i$ which exist for the same reason. Note that $f(N, K) \leq n_i$, and since limits preserve inequality, it follows that $k_i \leq n$, and further follows that $k \leq n$. By repeating this argument in the reverse direction, it follows that $k = n$.
    \item \textbf{Are Double Limits 1 and 2 mathematically equivalent to 3?} If we make the same monotonicity assumption as earlier, then we have that the minimization problem is monotonic in $C$. With this, we can apply a similar argument as above to show that the limit converges and is equivalent to $k$ and $n$.
    \item \textbf{How do we solve for the inner limit in Double Limit 1?} It is computationally prohibitive to tune all the hyperparameters for each choice of $N$ and $K$. Therefore, we would prefer searching for optimal hyperparameters once per choice of $N$ to fit the outer limit. Since we are only interested in the performance as $K \to\infty$ increases, we can reasonably approximate this via our hyperparameter heuristic  from Appendix~\ref{app:heuristic-hparams} without ever determining the optimal hyperparameters at lower values of $K$.
    \item \textbf{Why do we prefer Double Limit 1 to Double Limit 2?} We first make the observation that tuning hyperparameters depends on $N$, but if we follow the previous bullet's asymptote tuning heuristic, tuning hyperparameters does not depend on any finite value of $K$. Therefore, Approach 2 would still have to fit hyperparameters for each $N, K$, whereas Approach 1 avoids this by fitting it once per $N$.
    \item \textbf{Why do we not use Double Limit 3?} In Double Limit 1, we keep the hyperparameter as the scaling axis, instead of Double Limit 3 which sets compute as the scaling axis. When we choose to scale the hyperparameter, we can use our locally optimal hyperparameter search algorithm to find the best possible performance for that hyperparameter. This is difficult when scaling compute, since our hyperparameters such as model size and epoch count influence the compute spent during pre-training. Our preference reflects how practitioners typically use Chinchilla Approach 3 (fitting loss for best run given $N, D$, closer to Double Limit 1) over Chinchilla Approach 1 (fitting the envelope of the runs, closer to Double Limit 3)~\citep{besiroglu2024chinchillascalingreplicationattempt}.
\end{itemize}

\section{Data scaling}

\subsection{Epoch tuned baseline}\label{app:epoch-data-scaling}

Unlike regularized parameter scaling and ensemble scaling where we estimate the best possible loss via asymptotes (Section~\ref{sec:scaling-seed-token}), epoch tuning eventually over-fits. To estimate the best possible loss under epoch tuning for each token count, we use the following procedure.

\begin{enumerate}
    \item For a fixed token count $D$ and fixed parameter count $N$, search for locally optimal hyperparameters while fixing weight decay to be $0.1$
    \item Perform this for all parameter counts $N$ and token counts $D$
\end{enumerate}

\begin{figure}
  \centering
  \begin{minipage}[c]{0.32\textwidth}
    \centering
    \includegraphics[width=\linewidth]{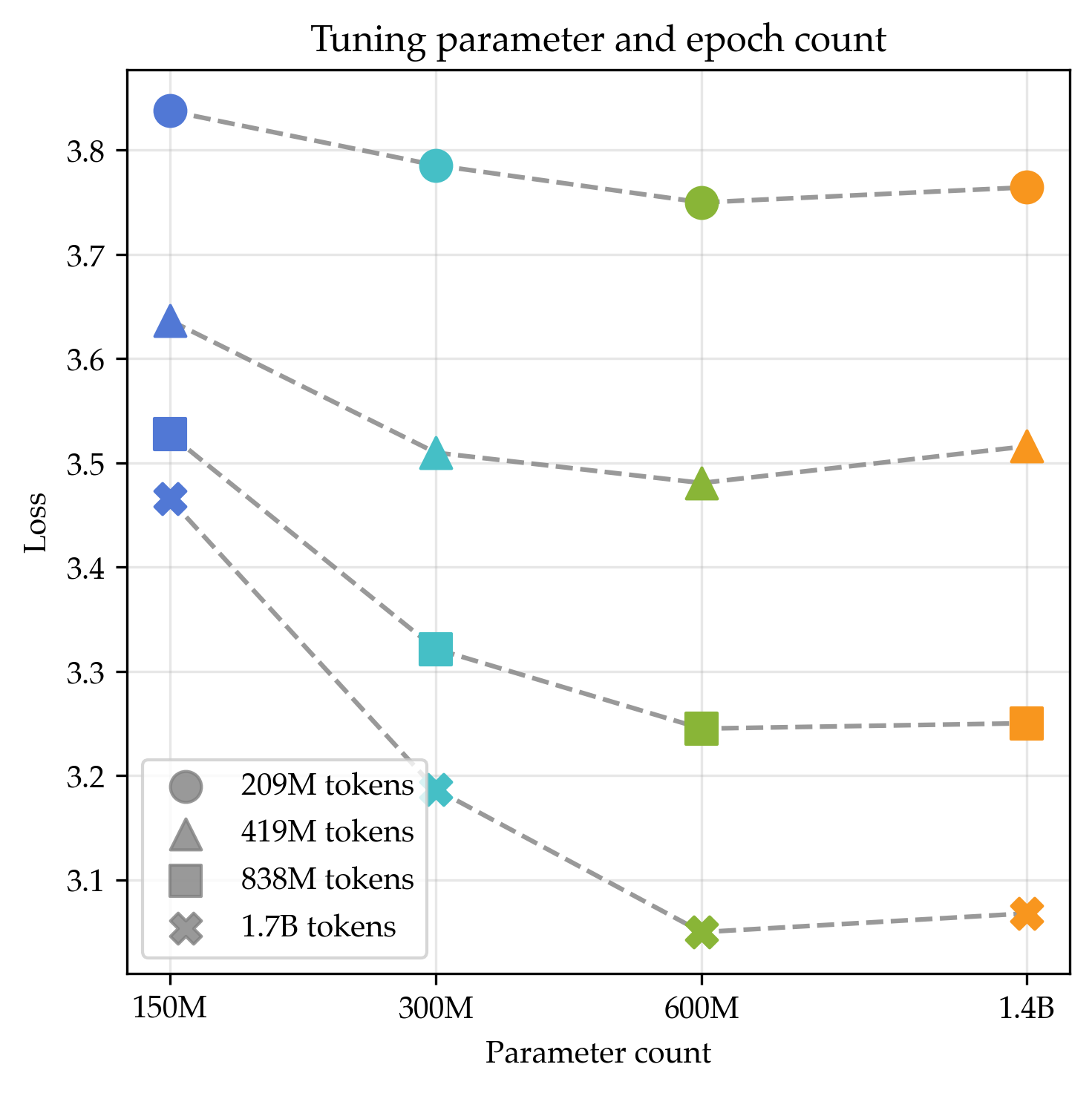}
  \end{minipage}
  \hfill
  \begin{minipage}[c]{0.6\textwidth}
    \captionof{figure}{\textbf{Scaling seed token count for epoching and parameter scaling.} For each seed token count, we train models of varying $N$ with jointly tuned learning rate and epoch count with a weight decay of $0.1$. We take the best possible model at each token count as an estimate of the best performance under the standard recipe.}
    \label{fig:baseline-seed-scaling}
  \end{minipage}
\end{figure}

After following this procedure, we found that 600M models and 1.4B models were within $\approx 0.02$ loss of each other and were much better than the other models, as shown in Figure~\ref{fig:baseline-seed-scaling}. Across all our token scales, 600M models slightly outperformed 1.4B models, which we discuss in~\ref{app:overfitting}. Therefore, we take the 600M performance as an estimate of the best possible loss under regularized parameter scaling. We believe the performance of this algorithm would be better under better width/depth/architectures, though we expect this benefit to translate to our other recipes as well.

\section{Distillation details}\label{app:distill}

\subsection{Data generation}
For all distillation experiments, we generate teacher data from the same model family: $K$-ensembles of 300M models with optimal hyperparameters for asymptotic performance. We choose to perform self-distillation with a $1$-ensemble from this family rather than the 300M model from the regularized recipe to cleanly isolate the effect of distilling from a stronger teacher. In practice, we don't observe a significant difference: self-distillation from the $1$-ensemble (blue point, Figure~\ref{fig:sketch-distill}) gives a loss of 3.43 while self-distillation from the 300M regularized recipe (purple point, Figure~\ref{fig:sketch-distill}) gives a loss of 3.44. 

Because we are unconstrained by train compute, optimal distillation should never epoch on teacher data and instead generate more. We pre-generate a large pool of teacher distillation data by sampling unconditionally with temperature 1 using a high-throughput inference engine designed for batched workloads~\citep{juravsky2025tokasaurus}. For generating ensemble teacher distillation data, we experiment with inferencing both the logit averaged ensemble as well as the individual members. We observe better student performance using the individual members. 

\subsection{Hyperparameters}\label{app:distill-hp}
For both ensemble distillation and self-distillation, we search for optimal hyperparameters using a procedure similar to Appendix~\ref{app:convex-certificates}. Our distillation recipe also introduces a new hyperparameter which we refer to as the mixing ratio: the ratio of batches of real data to synthetic data. A mixing ratio of $1:1$ indicates that we take the same number of gradient steps on real data as teacher-generated data. For example, if we have 209M tokens of real data that we wish to epoch on $16$ times, a $1:1$ mixing ratio would require $209 \cdot 10^{6} \times 16 = 3.3$B tokens of teacher-generated distillation data. We find tuning the mixing ratio to be important for performance. 

We detail the exact values of hyperparameters in Table~\ref{tab:distill_hps}. Interestingly, we observe that optimal weight decay for distillation is lower than that of our regularized recipes, in line with standard practice. In addition, we find that ensemble distillation admits a higher optimal mixing ratio, likely due to the greater diversity from the teacher's synthetic data. Our ensemble distillation run trains on a total of $16 \times 209 \cdot 10^{6} \times (1 + 9) = 33.4$B tokens, while our self-distillation run trains on a total of  $16 \times 209 \cdot 10^{6} \times (1 + 3) = 13.4$B tokens. Due to limitations in inferencing, we only generate 10B tokens each of ensemble distillation and self-distillation data, so our ensemble distillation may epoch up to 3 times on the teacher data. 
\begin{table}[h]
\centering
\begin{tabular}{lcc}
\toprule
\textbf{Parameter} & \textbf{Ensemble Distill} & \textbf{Self-Distill}\\
\midrule
Learning Rate & 3e-3 & 3e-3 \\
Weight Decay & 0.1 & 0.1 \\
Mixing Ratio & $1:9$
 & $1:3$ \\
Epochs & 16 & 16 \\
\bottomrule
\end{tabular}
\vspace{1em}
\caption{Optimal hyperparameters for ensemble and self-distillation.}
\label{tab:distill_hps}
\end{table}

\subsection{Mixing data ablation}
We provide a token-matched ablation for the effect of mixing in the real pre-training data when doing self-distillation. As in Appendix~\ref{app:distill-hp}, we start with the same pool of $10$B pre-generated tokens from a $300$M $1$-ensemble. Perfect distillation into a student model of the same size (with an infinite amount of teacher data) would achieve the same loss as the teacher. 

We compare self-distillation with and without mixing in real data. For mixing in real data, we epoch the real data $16$ times and use a $1 : 1$ mixing ratio so that the total number of tokens we train on is less than $10$B. For no mixing, we simply train on a subset of the pre-generated pool. Both methods use the same learning rate and batch size, train on a total of $16 \times 209 \cdot 10^{6} \times (1 + 1) = 6.688$B tokens, and never repeat the synthetic teacher data. For no mixing, we additionally search over weight decay. 

Table~\ref{tab:mixing_data_ablation} shows that without mixing in real pre-training data, self-distillation is substantially worse than the teacher model (as one might expect). Mixing data allows for self-distillation to exceed the teacher model.
\begin{table}[h]
\centering
\begin{tabular}{lccc}
\toprule
& Teacher Model & Self-Distill ($1:1$ mixing) &  Self-Distill (No mixing)\\
\midrule
Val Loss  & 3.7103 & 3.4373 & 4.0693 \\
\bottomrule
\end{tabular}
\vspace{1em}
\caption{Effect of mixing real pre-training data for self-distillation.}
\label{tab:mixing_data_ablation}
\end{table}

\section{Downstream task details}\label{app:downstream}

\subsection{Downstream tasks}
We provide a full breakdown of downstream benchmark scores per model type in Table~\ref{tab:appendix_benchmarks_indented}. We use \texttt{lm-evaluation-harness}~\citep{eval-harness} for our evaluations. Our evaluation code is available at \href{https://github.com/konwook/lm-eval-ensemble}{\texttt{https://github.com/konwook/lm-eval-ensemble}}.

\begin{table*}[t]
\centering
\setlength{\tabcolsep}{6pt}
\begin{tabular}{l l c c c c}
\toprule
\multicolumn{2}{l}{\textbf{Model type}} & \textbf{ARC-Easy (\%)} & \textbf{PIQA (\%)} & \textbf{SciQ (\%)} & \textbf{Avg (\%)} \\
\midrule
\multirow{4}{*}{Unregularized model scaling} & \hspace{0.8em}150M & $40.95_{\pm 1.01}$ & $59.68_{\pm 1.14}$ & $62.40_{\pm 1.53}$ & $54.35_{\pm 0.72}$ \\
& \hspace{0.8em}300M & $41.96_{\pm 1.01}$ & $61.15_{\pm 1.14}$ & $62.90_{\pm 1.53}$ & $55.34_{\pm 0.72}$ \\
& \hspace{0.8em}600M & $39.86_{\pm 1.00}$ & $59.90_{\pm 1.14}$ & $60.50_{\pm 1.55}$ & $53.42_{\pm 0.72}$ \\
& \hspace{0.8em}1.4B  & $40.61_{\pm 1.01}$ & $60.39_{\pm 1.14}$ & $61.40_{\pm 1.54}$ & $54.14_{\pm 0.72}$ \\
\midrule
\multirow{4}{*}{Model scaling} & \hspace{0.8em}150M & $41.29_{\pm 1.01}$ & $60.17_{\pm 1.14}$ & $63.90_{\pm 1.52}$ & $55.12_{\pm 0.72}$ \\
 & \hspace{0.8em}300M & $44.28_{\pm 1.02}$ & $61.81_{\pm 1.13}$ & $69.10_{\pm 1.46}$ & $58.39_{\pm 0.70}$ \\
 & \hspace{0.8em}600M & $47.10_{\pm 1.02}$ & $63.06_{\pm 1.13}$ & $69.70_{\pm 1.45}$ & $59.95_{\pm 0.70}$ \\
 & \hspace{0.8em}1.4B & $45.66_{\pm 1.02}$ & $63.82_{\pm 1.12}$ & $72.70_{\pm 1.41}$ & $60.73_{\pm 0.69}$ \\
\midrule
\multirow{5}{*}{150M ensembles} & \hspace{0.8em}$K=1$ & $42.85_{\pm 1.02}$ & $60.88_{\pm 1.14}$ & $64.90_{\pm 1.51}$ & $56.21_{\pm 0.72}$ \\
 & \hspace{0.8em}$K=2$ & $44.61_{\pm 1.02}$ & $62.02_{\pm 1.13}$ & $65.80_{\pm 1.50}$ & $57.48_{\pm 0.71}$ \\
 & \hspace{0.8em}$K=3$ & $44.91_{\pm 1.02}$ & $62.08_{\pm 1.13}$ & $68.30_{\pm 1.47}$ & $58.43_{\pm 0.71}$ \\
 & \hspace{0.8em}$K=4$ & $45.83_{\pm 1.02}$ & $62.19_{\pm 1.13}$ & $69.20_{\pm 1.46}$ & $59.07_{\pm 0.70}$ \\
 & \hspace{0.8em}$K=5$ & $45.50_{\pm 1.02}$ & $62.30_{\pm 1.13}$ & $70.40_{\pm 1.44}$ & $59.40_{\pm 0.70}$ \\
\midrule
\multirow{5}{*}{300M ensembles} & \hspace{0.8em}$K=1$ & $44.36_{\pm 1.02}$ & $62.95_{\pm 1.13}$ & $68.10_{\pm 1.47}$ & $58.47_{\pm 0.71}$ \\
 & \hspace{0.8em}$K=2$ & $46.72_{\pm 1.02}$ & $63.87_{\pm 1.12}$ & $70.70_{\pm 1.44}$ & $60.43_{\pm 0.70}$ \\
 & \hspace{0.8em}$K=3$ & $47.77_{\pm 1.02}$ & $64.74_{\pm 1.11}$ & $72.90_{\pm 1.41}$ & $61.80_{\pm 0.69}$ \\
 & \hspace{0.8em}$K=4$ & $48.36_{\pm 1.03}$ & $65.89_{\pm 1.11}$ & $73.10_{\pm 1.40}$ & $62.45_{\pm 0.69}$ \\
 & \hspace{0.8em}$K=5$ & $49.33_{\pm 1.03}$ & $65.67_{\pm 1.11}$ & $74.00_{\pm 1.39}$ & $63.00_{\pm 0.68}$ \\
\midrule
\multirow{5}{*}{600M ensembles} & \hspace{0.8em}$K=1$ & $45.92_{\pm 1.02}$ & $62.84_{\pm 1.13}$ & $68.50_{\pm 1.47}$ & $59.09_{\pm 0.71}$ \\
 & \hspace{0.8em}$K=2$ & $47.56_{\pm 1.02}$ & $64.04_{\pm 1.12}$ & $71.80_{\pm 1.42}$ & $61.13_{\pm 0.69}$ \\
 & \hspace{0.8em}$K=3$ & $48.44_{\pm 1.03}$ & $64.25_{\pm 1.12}$ & $73.30_{\pm 1.40}$ & $62.00_{\pm 0.69}$ \\
 & \hspace{0.8em}$K=4$ & $49.03_{\pm 1.03}$ & $64.80_{\pm 1.11}$ & $73.70_{\pm 1.39}$ & $62.51_{\pm 0.69}$ \\
 & \hspace{0.8em}$K=5$ & $50.34_{\pm 1.03}$ & $64.80_{\pm 1.11}$ & $75.30_{\pm 1.36}$ & $63.48_{\pm 0.68}$ \\
\midrule
\multirow{5}{*}{1.4B ensembles} & \hspace{0.8em}$K=1$ & $43.56_{\pm 1.02}$ & $64.20_{\pm 1.12}$ & $68.80_{\pm 1.47}$ & $58.85_{\pm 0.70}$ \\
 & \hspace{0.8em}$K=2$ & $47.26_{\pm 1.02}$ & $65.13_{\pm 1.11}$ & $75.30_{\pm 1.36}$ & $62.56_{\pm 0.68}$ \\
 & \hspace{0.8em}$K=3$ & $49.33_{\pm 1.03}$ & $65.40_{\pm 1.11}$ & $76.50_{\pm 1.34}$ & $63.74_{\pm 0.67}$ \\
 & \hspace{0.8em}$K=4$ & $48.86_{\pm 1.03}$ & $66.38_{\pm 1.10}$ & $77.80_{\pm 1.31}$ & $64.35_{\pm 0.67}$ \\
 & \hspace{0.8em}$K=5$ & $49.71_{\pm 1.03}$ & $66.38_{\pm 1.10}$ & $77.10_{\pm 1.33}$ & $64.39_{\pm 0.67}$ \\
\midrule
\multirow{2}{*}{Distillation (300M)} & \hspace{0.8em}Self & $46.68_{\pm 1.02}$ & $62.35_{\pm 1.13}$ & $72.60_{\pm 1.41}$ & $60.54_{\pm 0.69}$ \\
& \hspace{0.8em}Ensemble & $48.44_{\pm 1.03}$ & $62.84_{\pm 1.13}$ & $75.30_{\pm 1.36}$ & $62.19_{\pm 0.68}$ \\
\midrule 
\multirow{2}{*}{Model soups} & \hspace{0.8em}$K=2$ & $26.56_{\pm 0.91}$ & $54.84_{\pm 1.16}$ & $24.70_{\pm 1.36}$ & $35.37_{\pm 0.67}$ \\
 & \hspace{0.8em}$K=4$ & $24.96_{\pm 0.89}$ & $55.28_{\pm 1.16}$ & $23.90_{\pm 1.35}$ & $34.71_{\pm 0.66}$\\
\bottomrule
\end{tabular}
\caption{Benchmark accuracies of all methods using 200M tokens on ARC-Easy, PIQA, and SciQ with averages. Entries are value$_{\pm\,\mathrm{SE}}$ in percentage points.}
\label{tab:appendix_benchmarks_indented}
\end{table*}

\subsection{Hyperparameter tuning}
\begin{figure}
    \centering
    \begin{subfigure}[b]{0.45\textwidth}
        \centering
        \includegraphics[width=\textwidth]{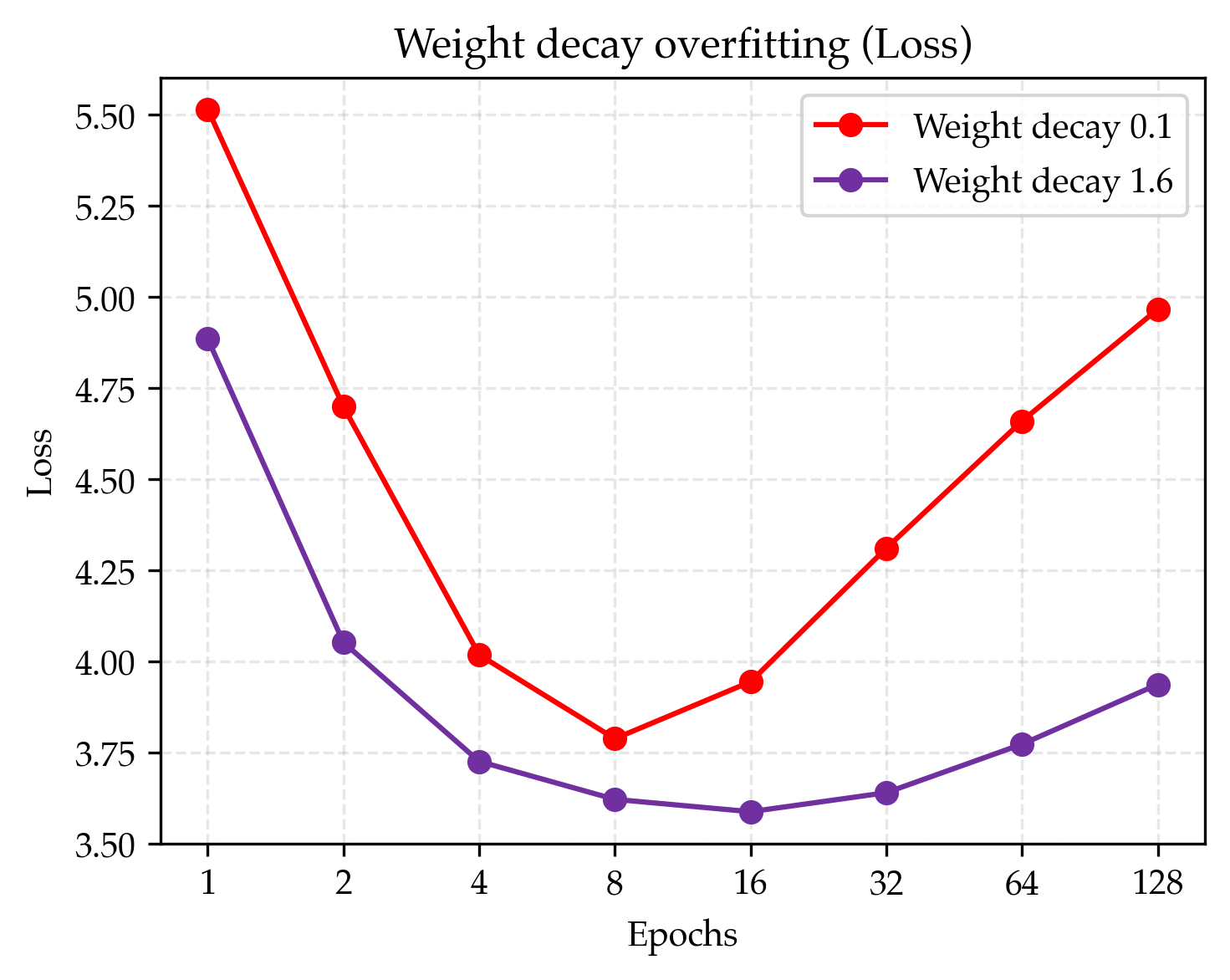}
    \end{subfigure}
    \begin{subfigure}[b]{0.45\textwidth}
        \centering
        \includegraphics[width=\textwidth]{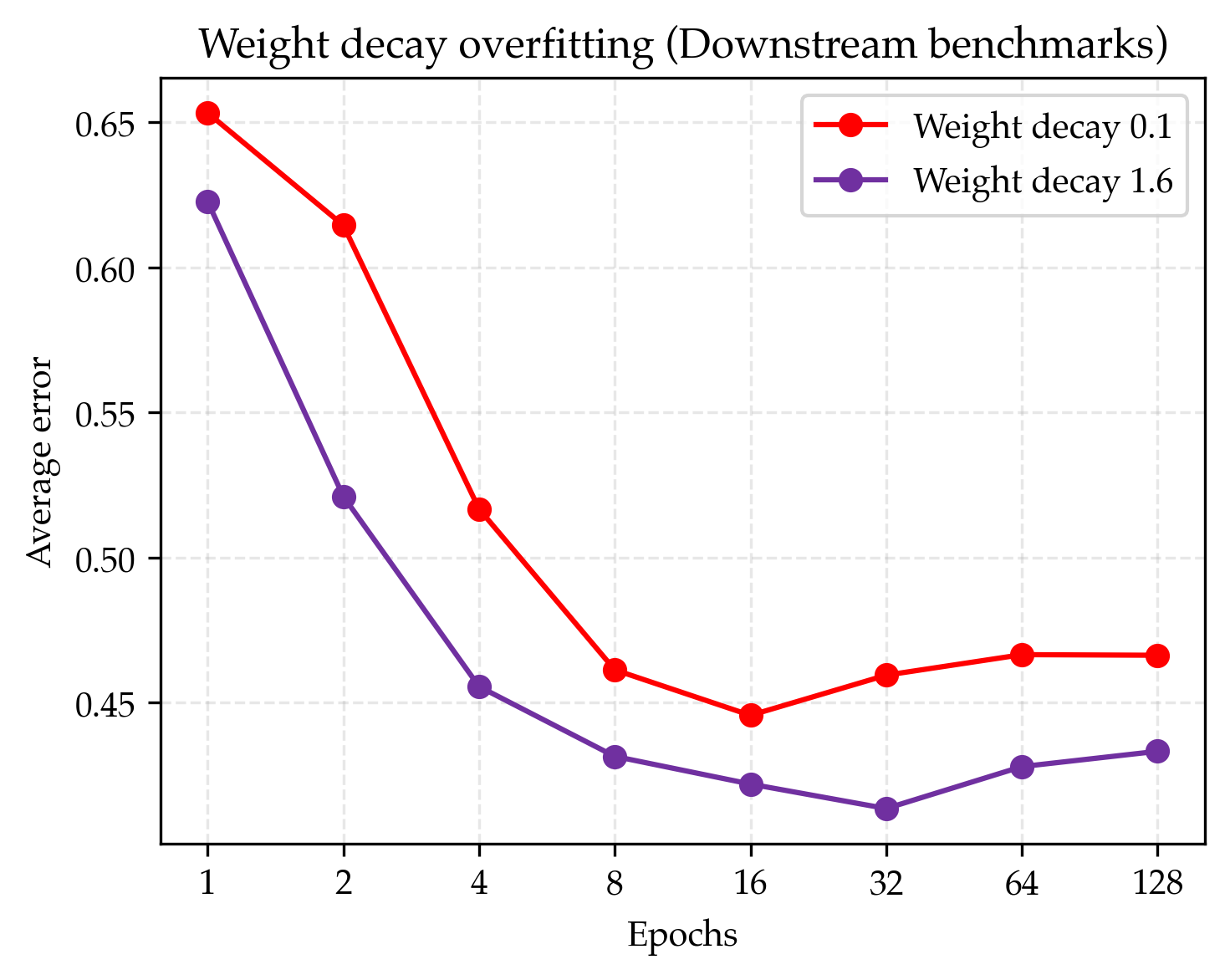}
    \end{subfigure}
    \caption{\textbf{Effect of regularization on overfitting for downstream benchmarks.} Downstream benchmarks also reflect the benefit of heavy regularization on performance. The effect of overfitting on downstream benchmarks (right) appears at twice the epoch count compared to validation loss (left).}
    \label{fig:wd-overfit-benchmark}
\end{figure}

We find that hyperparameter tuning from validation loss transfers to downstream benchmarks as well. Figure~\ref{fig:wd-overfit-benchmark} (left) shows how adding heavy regularization with weight decay (with a fixed learning rate of 3e-3) shifts the overfitting point based on validation loss to the right and down. We observe a similar effect in Figure~\ref{fig:wd-overfit-benchmark} (right), although the overfitting threshold (in epochs) is twice the threshold observed for validation loss.  

\section{Continued pre-training}\label{app:cpt}

\subsection{Hyperparameters}

Hyperparameters for continued pre-training baselines are shown in Table~\ref{tab:cpt_hps}. The $73$B CPT run uses the default hyperparameters from~\citep{wang2025octothinkermidtrainingincentivizesreinforcement}, except for learning rate which we tuned ourselves. The individual members of the $K$-ensembles use the same hyperparameters as the standard recipe. 
\begin{table}[h!]
\centering
\begin{tabular}{lrrr}
\toprule
\textbf{Parameter} & \textbf{Default} & \textbf{Lower BS} & \textbf{Epoching}\\
\midrule
Learning Rate & 3e-5 & 3e-5 & 3e-5\\
Weight Decay & 0.1 & 0.1  & 0.1\\
Batch Size & 512 & 64 & 64\\
Epochs & 1 & 1  & 4 \\
\bottomrule
\end{tabular}
\vspace{1em}
\caption{Hyperparameters for continued pre-training.}
\label{tab:cpt_hps}
\end{table}

\subsection{CPT soups}
We ablate the performance of model soups compared to ensembling in our continued pre-training setting by averaging the weights of the members instead of ensembling them. Unlike standard pre-training, CPT soups perform strongly and slightly outperform ensembles as we increase the number of averaged models (Table~\ref{tab:cpt-soups}).
\renewcommand{\arraystretch}{1.2}
\begin{table}[h!]
\centering
\caption{Continually pre-trained ensembles vs. soups}
\vspace{1.5mm}
\begin{adjustbox}{max width=\linewidth}
\begin{tabular}{l|c|ccc|ccc}
\hline
\multirow{2}{*}{\textbf{Benchmarks}}
& \multirow{2}{*}{\textbf{Llama 3B}}
  & \multicolumn{3}{c|}{\textbf{$K$-ensembles}}
  & \multicolumn{3}{c}{\textbf{$K$-soups}} \\
\cline{3-5}\cline{6-8}
&& $K=2$ & $K=4$ & $K=8$ & $K=2$ & $K=4$ & $K=8$ \\
\hline
$\text{GSM8K}_{\text{(8-shot)}}$  & 28.23 &  49.28 & 51.80 & 52.99  & 49.73 & 53.83 & \textbf{54.96} \\
$\text{MATH}_{\text{(4-shot)}}$    & 6.90 &  21.84 & 23.04 & 23.50  & 22.40 & 23.02 & \textbf{23.72} \\
$\text{MATHQA}_{\text{(8-shot)}}$  & 35.07 & 45.12 & 46.06 & 45.26  & 44.59 & \textbf{46.10} & 45.33 \\
\hline
Average                 & 24.25 & 38.79 &  40.35  & 40.58 & 38.91 & 40.98 & \textbf{41.34}  \\
\hline
\end{tabular}
\end{adjustbox}
\label{tab:cpt-soups}
\end{table}
\renewcommand{\arraystretch}{1.0}

\section{Power laws}

\subsection{Sensitivity analysis}\label{app:sensitivity}

\begin{figure}
    \centering
    \includegraphics[width=0.45\textwidth]{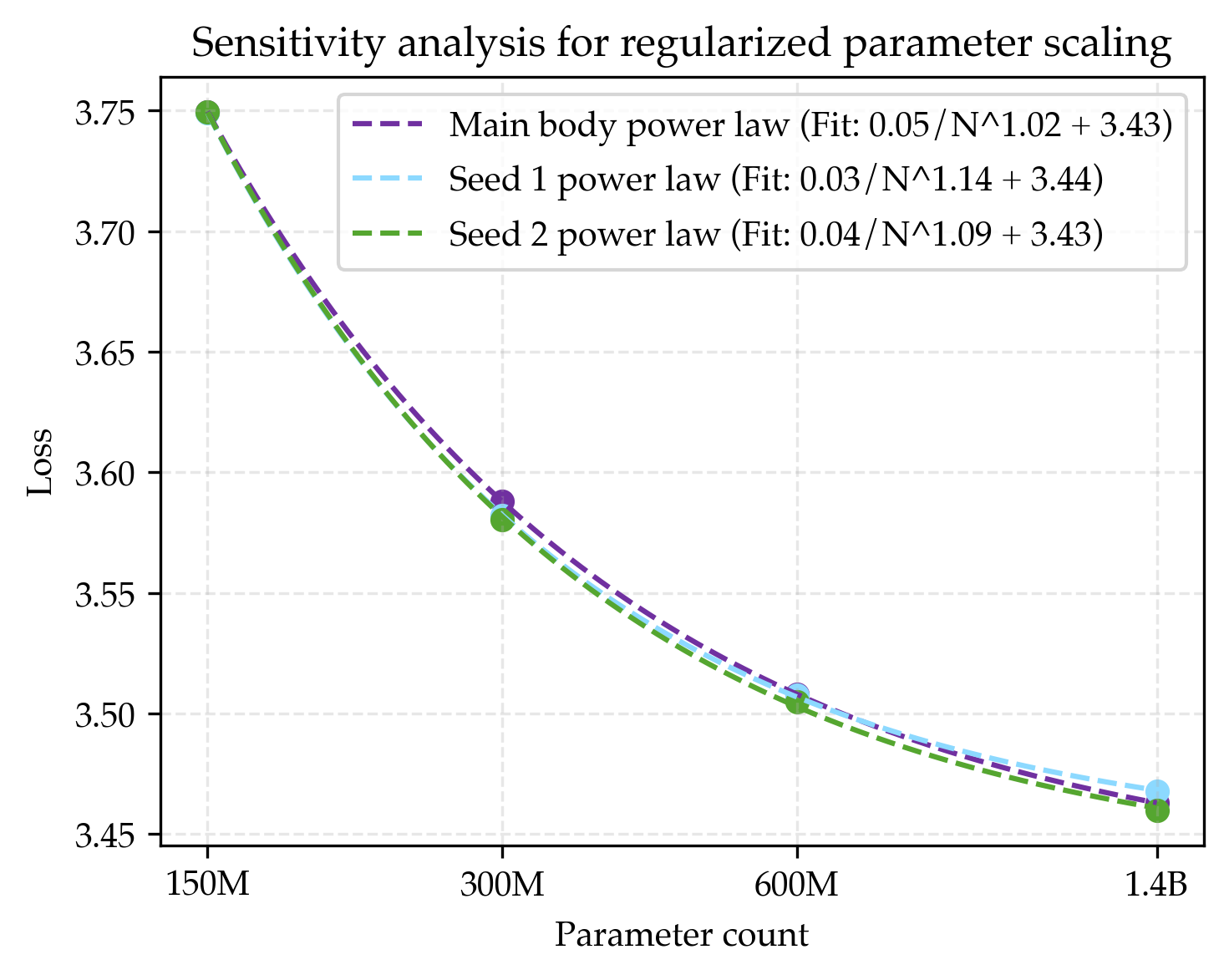}
    \includegraphics[width=0.45\textwidth]{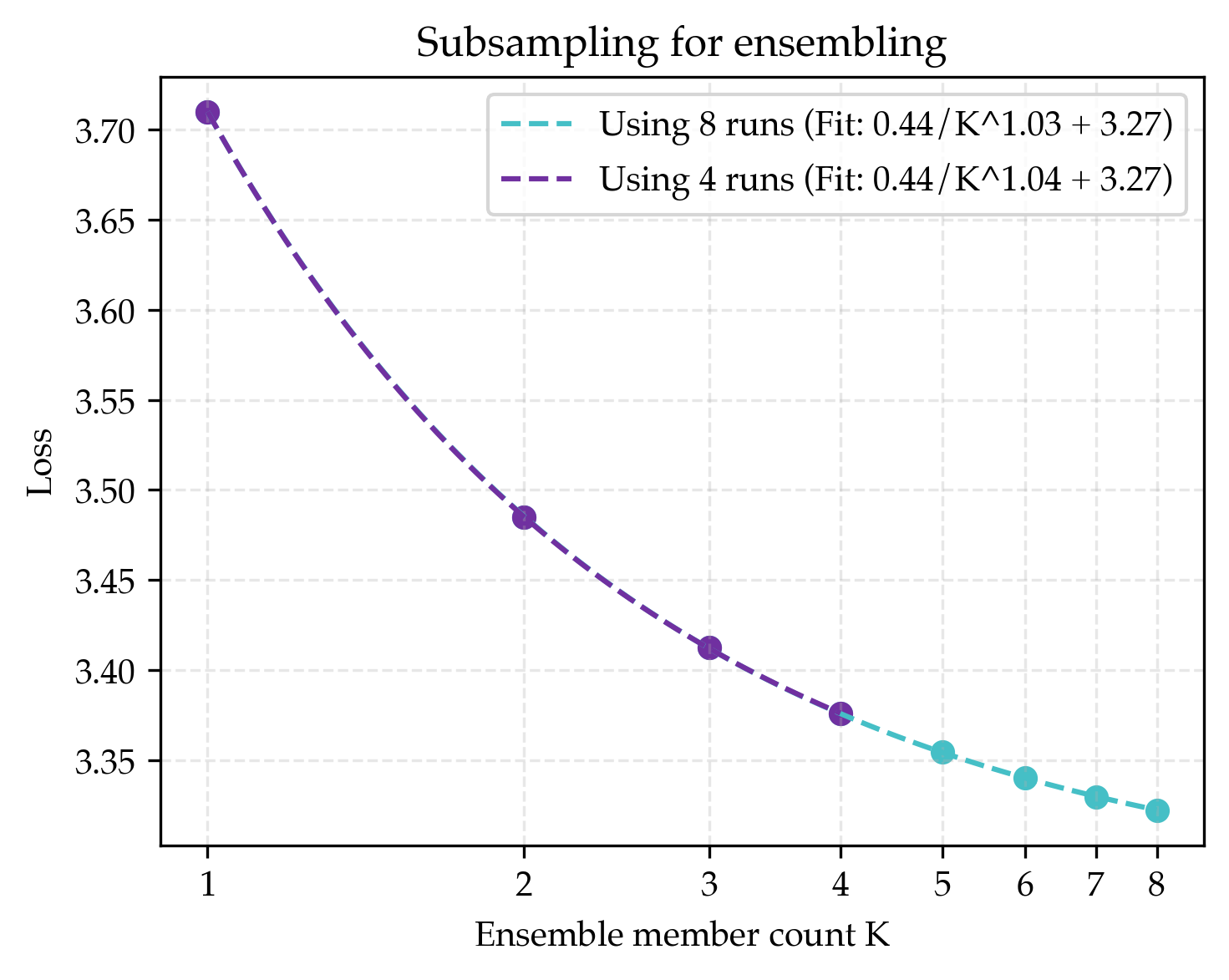}
    \caption{\textbf{Sensitity analysis.} Left: When re-fitting the regularized power law across two additional seeds, we find that the asymptote stays relatively stable. Right: When subsampling the number of points for the ensemble scaling law, we find that the power law barely changes.}
    \label{fig:sensitivity}
\end{figure}

To test whether our asymptote estimation is reliable due to run-to-run variance, we conduct a sensitivity analysis for regularized parameter scaling and ensembling, shown in Figure~\ref{fig:sensitivity}.

To test parameter scaling, we fit three power laws to all the models trained where each power law uses a different seed (governing data order and model initialization). Though the scaling laws change per seed, they remain relatively consistent, with the asymptotes staying close together. This is encouraging, as the standard deviation in asymptotes is close to the run-to-run standard deviation for 300M models (Appendix~\ref{app:seed-science}).

Since we have more runs for ensembling, we test the reliability of ensembling by subsampling the number of members. When fitting a power law using up to four ensemble members, we find an extremely similar law to using up to eight ensemble members. Qualitatively, over the course of our experiments, we found that the scaling law for ensembling is a lot more stable than the scaling law for parameter scaling.

We note that this is a limited stress-test and that it is likely our asymptote estimation procedure is quite noisy. Furthermore, we note that this does not test our two-tier and three-tier power laws for joint scaling of parameters, members, and data, nor does it test our settings where our best run is further from the asymptote. We advise taking these asymptotes with a grain of salt and interpreting them as rough estimates.

\subsection{Fitting laws}

To fit our power laws, we use \texttt{scipy.optimize.curve\_fit}, either with no initial conditions and bounds or with \texttt{p0=[1.0, 0.5, 2.0]} and \texttt{bounds=([0, 0, 0], [np.inf, np.inf, np.inf])`}. We note that unlike prior work where such parameters have been found to be important~\citep{besiroglu2024chinchillascalingreplicationattempt,hoffmann2022trainingcomputeoptimallargelanguage}, we did not find them to be critical considering how our fits are simple and over 1 dimension.

\end{document}